\documentclass[sigconf,authorversion]{acmart}
\usepackage{float, xcolor, multirow, subcaption, tikz}
\usetikzlibrary{shapes,fit,positioning,calc}

\newcommand{\lb}{\mathbin{\text{\scalebox{1.5}{$\bullet$}}}}
\newcommand{\lc}{\mathbin{\text{\scalebox{1.5}{$\circ$}}}}
\newcommand{\chref}[2]{%
  \href{#1}{\textcolor[rgb]{0,0.518,0.871}{#2}}
}

\AtBeginDocument{%
  \providecommand\BibTeX{{%
    \normalfont B\kern-0.5em{\scshape i\kern-0.25em b}\kern-0.8em\TeX}}}

\copyrightyear{2023}
\acmYear{2023}
\setcopyright{acmlicensed}\acmConference[Web3D '23]{The 28th International ACM Conference on 3D Web Technology}{October 9--11, 2023}{San Sebastian, Spain}
\acmBooktitle{The 28th International ACM Conference on 3D Web Technology (Web3D '23), October 9--11, 2023, San Sebastian, Spain}
\acmPrice{15.00}
\acmDOI{10.1145/3611314.3615904}
\acmISBN{979-8-4007-0324-9/23/10}

\begin{document}

\title[Tirtha]{Tirtha - An Automated Platform to Crowdsource Images and Create 3D Models of Heritage Sites}

\author{Jyotirmaya Shivottam}
\email{jyotirmaya.shivottam@niser.ac.in}
\orcid{0000-0002-2688-9421}
\author{Subhankar Mishra}
\email{smishra@niser.ac.in}
\orcid{0000-0002-9910-7291}
\affiliation{%
  \institution{National Institute of Science Education and Research\\An OCC of Homi Bhabha National Institute}
  \streetaddress{Jatni}
  \city{Bhubaneswar}
  \state{Odisha}
  \country{India}
  \postcode{752050}
}

\begin{abstract}
  Digital preservation of Cultural Heritage (CH) sites is crucial to protect them against damage from natural disasters or human activities. Creating 3D models of CH sites has become a popular method of digital preservation thanks to advancements in computer vision and photogrammetry. However, the process is time-consuming, expensive, and typically requires specialized equipment and expertise, posing challenges in resource-limited developing countries. Additionally, the lack of an open repository for 3D models hinders research and public engagement with their heritage. To address these issues, we propose Tirtha, a web platform for crowdsourcing images of CH sites and creating their 3D models. Tirtha utilizes state-of-the-art Structure from Motion (SfM) and Multi-View Stereo (MVS) techniques. It is modular, extensible and cost-effective, allowing for the incorporation of new techniques as photogrammetry advances. Tirtha is accessible through a web interface at \url{https://tirtha.niser.ac.in} and can be deployed on-premise or in a cloud environment. In our case studies, we demonstrate the pipeline's effectiveness by creating 3D models of temples in Odisha, India, using crowdsourced images. These models are available for viewing, interaction, and download on the Tirtha website. Our work aims to provide a dataset of crowdsourced images and 3D reconstructions for research in computer vision, heritage conservation, and related domains. Overall, Tirtha is a step towards democratizing digital preservation, primarily in resource-limited developing countries.
\end{abstract}

\begin{CCSXML}
  <ccs2012>
    <concept>
        <concept_id>10002951.10003260.10003282.10003296</concept_id>
        <concept_desc>Information systems~Crowdsourcing</concept_desc>
        <concept_significance>500</concept_significance>
    </concept>
    <concept>
        <concept_id>10010405.10010476.10003392</concept_id>
        <concept_desc>Applied computing~Digital libraries and archives</concept_desc>
        <concept_significance>500</concept_significance>
    </concept>
    <concept>
        <concept_id>10010147.10010371.10010382.10010385</concept_id>
        <concept_desc>Computing methodologies~Image-based rendering</concept_desc>
        <concept_significance>300</concept_significance>
    </concept>
  </ccs2012>
\end{CCSXML}

\ccsdesc[500]{Information systems~Crowdsourcing}
\ccsdesc[500]{Applied computing~Digital libraries and archives}
\ccsdesc[300]{Computing methodologies~Image-based rendering}

\keywords{digital heritage, crowdsourcing, photogrammetry, 3D dataset, open source}

\maketitle

\section{Introduction}
\label{sec:intro}
\begin{figure*}
  \centering
  \begin{subfigure}{.6\textwidth}
      \centering
      \includegraphics[width=\linewidth]{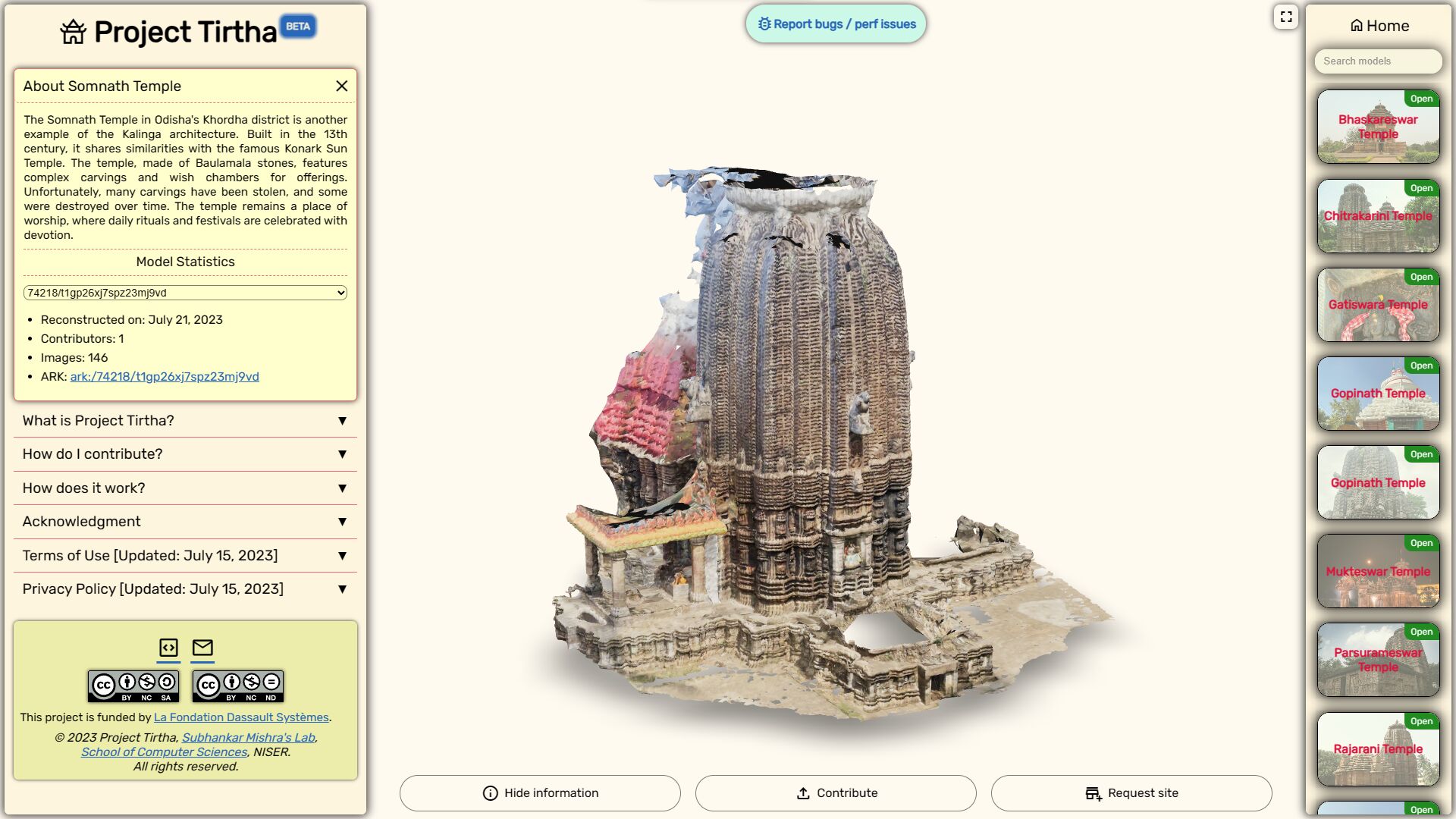}
      \caption{On desktop}
  \end{subfigure}%
  \hspace*{0.5cm}
  \begin{subfigure}{.15\textwidth}
      \centering
      \includegraphics[width=\linewidth]{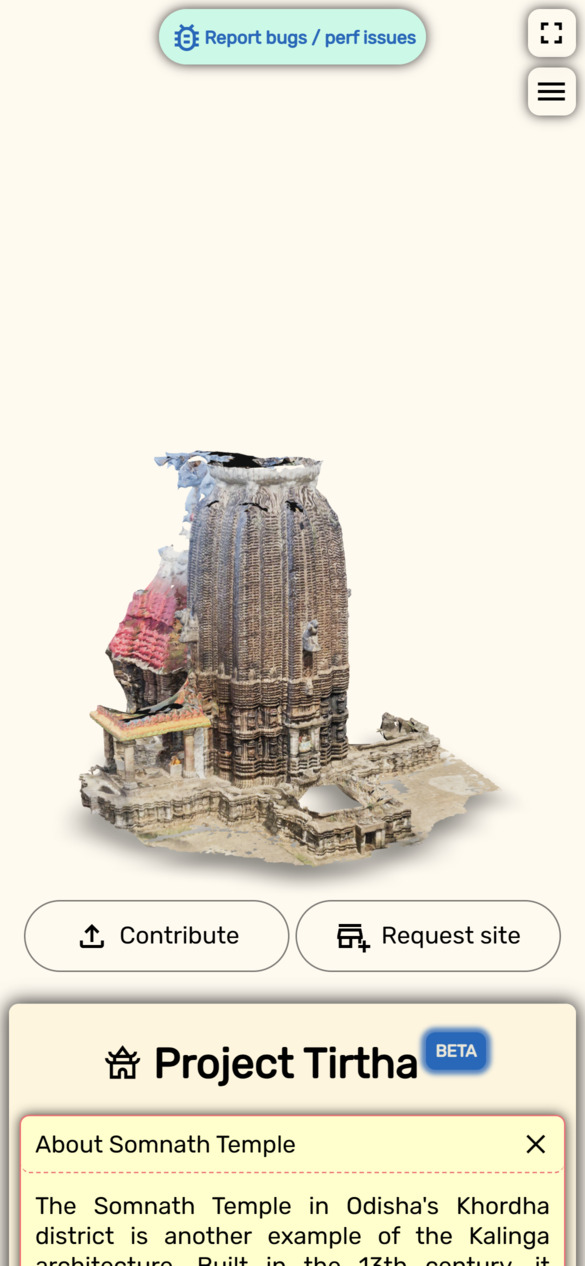}
      \caption{On mobile}
  \end{subfigure}
  \caption{Screenshots of the Tirtha website \textsuperscript{(\chref{https://tirtha.niser.ac.in/ark:/74218/t1gp26xj7spz23mj9vd}{ARK})}}
  \Description{Screenshots of the responsive Tirtha website, showing the three-column layout on desktop and the single-column layout on mobile.}
  \label{fig:landing}
\end{figure*}
According to the United Nations Educational, Scientific and Cultural Organization (UNESCO)\footnote{\url{https://uis.unesco.org/en/glossary-term/cultural-heritage}}, \textit{Cultural Heritage} (CH) includes ``artefacts, monuments, a group of buildings and sites, museums that have a diversity of values including symbolic, historic, artistic, aesthetic, ethnological or anthropological, scientific and social significance''. This encompasses \textit{tangible cultural heritage} (TCH), such as buildings, paintings, sculptures, books and manuscripts, and \textit{intangible cultural heritage} (ICH), such as the practices, expressions, knowledge, skills and instruments embedded into cultural and natural heritage artefacts, sites or monuments. TCH or CH sites usually house ICH, often being inseparable. These sites form a vital part of the world's history and identity, acting as a looking glass into our ancestors' beliefs and way of life. As the world follows a trajectory of rapid socioeconomic growth, these sites are increasingly coming under threat, be it from natural sources such as earthquakes, floods, and landslides \cite{sochproto, soch}, or from anthropogenic sources such as urbanization, vandalism and unmitigated pollution, making it exigent to document and preserve them. UNESCO recognizes this, as seen in their 2015 policy \cite{unesco}, adopted by the World Heritage Committee, which aims to integrate sustainable development with the preservation of CH sites by aiding member states and institutions in harnessing the potential of these sites to contribute to sustainable development. Recently, concerns regarding the destruction of sites in geopolitically unstable regions have also been raised, and the need to document and preserve these sites has been recognized \cite{rekrei}.

However, resources for UNESCO and national organizations are often limited \cite{niti}, making it challenging to capture the vast number of sites. For example, as of May 2023, there are 1157 UNESCO-recognized World Heritage sites, with 40 in India\footnote{\url{https://whc.unesco.org/en/list/stat/}}. However, the Archaeological Survey of India (ASI) has recently reported that India alone has about 3695 sites recognized as Monuments of National Importance \cite{moni}. When considering State Protected Monuments, the number increases by about 4506, excluding several sites, yet to be recognized. The disparity between the number of sites formalized by UNESCO and ASI is due to limited resources available for documentation and nomination of sites \cite{niti, moni}, and is a prevalent issue in many developing countries \cite{developing}. To address this challenge, we explore the fields of \textit{digital heritage} (DH) and \textit{crowdsourcing} as potential sources of cost-effective solutions. DH uses digital technologies to record, preserve, and disseminate knowledge from cultural heritage (CH) sites. It can range from the creation of 3D models of monuments \cite{earlydh, chng1} and monitoring of historical sites \cite{remotemon1, remotemon2}, to using extended reality (XR) to create immersive experiences. UNESCO had adopted\footnote{\url{https://www.unesco.org/en/legal-affairs/charter-preservation-digital-heritage}} a charter for DH in 2003, identifying its potential. On the other hand, \textit{crowdsourcing}, a portmanteau of `crowd' and `outsourcing', involves obtaining information and services from large groups of people \cite{cs}. Notable examples of crowdsourcing include Wikipedia \cite{wiki}, the Zooniverse project \cite{zoo}, and OpenStreetMap \cite{osm}. Inspired by these concepts and related works, we present \textit{Tirtha}\footnote{\url{https://tirtha.niser.ac.in}}: a crowdsourcing platform for digital documentation of CH sites. The word `Tirtha' is Sanskrit for ``a place of pilgrimage'', commonly used to refer to the sacred sites of Hinduism \& Buddhism. Tirtha includes a web-based frontend for collecting images and showcasing 3D models, along with an automated pipeline for 3D reconstructions. This work addresses the lack of a cost-effective method to document CH sites, in culturally rich but resource-limited countries like India. By leveraging the proliferation of smartphones and advancements in smartphone photography, web technologies, and mobile internet, Tirtha aims to overcome the cost and accessibility challenges in documenting CH sites. The project aims to democratize the process by involving the general public along with experts through open-source contributions and an accessible interface.

The paper is organized as follows: Section~\ref{sec:rel} reviews related work in digital heritage and crowdsourcing, section~\ref{sec:appr} describes our proposed approach, and section~\ref{sec:res} presents results from two pilot studies. Finally, section~\ref{sec:disc} discusses Tirtha's limitations and explores possible solutions, while section~\ref{sec:conc} concludes the paper.

\section{Related Work}
\label{sec:rel}
Tirtha broadly falls under crowdsourced digital heritage \cite{developing}. One of the earliest works in this field is the now inactive PhotoCity \cite{photocity}, which used a gamified approach for volunteer-based image collection and 3D reconstruction of urban sites. While it had an end-to-end pipeline for data collection and 3D reconstruction, it lacked a web viewer for the models and a proper citation mechanism. WikiLovesMonuments \cite{wlm} is a popular modern-day example of crowdsourced digital heritage, focused on collecting images of cultural heritage (CH) sites for use in Wikipedia articles and research purposes. However, it does not explicitly target 3D reconstruction of CH sites.

Other works concentrate on the technical aspects of digital heritage, for example, pipelines for 3D reconstruction, without exploring the social aspects of crowdsourcing, such as \cite{earlydh, chng1, pipeline, arduino}. Some prominent works in this category include 3D-ICONS~\cite{3dicons1} and Ai\"oli~\cite{aioli}. Generally, these works utilize data from platforms like Europeana\footnote{\url{https://www.europeana.eu/portal/en}}, Google Arts \& Culture\footnote{\url{https://artsandculture.google.com/}}, and Open Heritage 3D\footnote{\url{https://openheritage3d.org/}}, and the resulting 3D models are archived and exhibited on those platforms. However, these pipelines may involve proprietary components or manual steps, limiting their reusability and extensibility. Moreover, the released models may have different licenses, complicating sharing and reuse, and the lack of web-based viewers makes it challenging to discover and cite the generated models.

Some works scrape pictures from social media and image-sharing platforms to reconstruct CH sites, in some cases to remotely monitor them \cite{diginvNzo, remotemon1, cs1scrape, cs2scrape, cs3scrape}. This approach eliminates the need for specialized equipment or dedicated data collection campaigns but suffers from an overrepresentation of popular sites and a lack of reliable metadata. But for sites that are at risk, or have been damaged due to natural disasters or human conflict, these sources are the only way to reconstruct the sites \cite{bridge, bel}.

Recent works incorporate crowdsourcing as an interactive activity between volunteers and heritage objects, particularly museum assets \cite{diginva, invasifaux, virlivmus1, virlivmus2, virlivmus3umm}. These projects involve virtual tours and storytelling experiences using XR tools to enhance volunteer engagement and museum performance \cite{pipeline, xrreview}. These have shown good public engagement, and result in better 3D models. However, they do not scale well for large-scale 3D reconstruction of heritage sites spread over vast areas, such as temples, forts, and palaces.
\begin{table*}
  \centering
  \caption{Comparison of Tirtha with related works. Note that the `Large-scale' column tracks the scale of the 3D reconstructions in these projects.}
  \begin{tabular}{lcccc|c|c}
    \hline
    \textbf{Platforms} & \textbf{Crowdsourced} & \textbf{CrossRef} & \textbf{FOSS} & \textbf{E2E Automated} & \textbf{Large-scale} & \textbf{Active} \\
    \hline
    \chref{http://www.photocitygame.com/}{PhotoCity} \cite{photocity} & $\lb$ & $\lc$ & $\lc$ & $\lb$ & $\lb$ & $\lc$ \\
    \chref{http://3dicons-project.eu/}{3D-ICONS} \cite{3dicons1} & $\lc$ & $\lb$ & $\lc$ & $\lc$ & $\lb$ & $\lc$ \\
    \chref{http://heritagetogether.org/}{HeritageTogether} \cite{heritoge1} & $\lb$ & $\lc$ & $\lc$ & $\lb$ & $\lb$ & $\lc$ \\
    \chref{https://crowdsourced.micropasts.org/}{MicroPasts} \cite{microp} & $\lb$ & $\lc$ & $\lb$ & $\lc$ & $\lc$ & $\lb$ \\
    \chref{https://rekrei.org/}{Rekrei} \cite{rekrei} & $\lb$ & $\lc$ & $\lb$ & $\lc$ & $\lb$ & $\lb$ \\
    SOCH \cite{soch} & $\lb$ & $\lc$ & $\lc$ & $\lc$ & $\lb$ & $\lc$ \\
    \chref{http://www.aioli.cloud/}{Aioli} \cite{aioli} & $\lc$ & $\lc$ & $\lc$ & $\lb$ & $\lb$ & $\lc$ \\
    \chref{https://crowdheritage.eu/en}{CrowdHeritage} \cite{ch} & $\lb$ & $\lb$ & $\lc$ & $\lc$ & $\lc$ & $\lb$ \\
    \hline
    Tirtha & $\lb$ & $\lb$ & $\lb$ & $\lb$ & $\lb$ & $\lb$ \\
    \hline
  \end{tabular}
  \label{tab:chart}
\end{table*}

Most of the aforementioned works rely on \textit{nichesourcing}, which targets specific groups like experts or volunteers for small-scale reconstructions. In contrast, we aim to democratize digital documentation for CH sites by expanding the audience to non-experts such as tourists or locals and involve them in contributing images of CH sites, which can be used to generate 3D models through low-cost photogrammetry pipelines. This approach aligns with \textit{participatory digital heritage}, empowering local communities and allowing them to share their heritage globally, as opposed to being limited by the resources available to their cultural institutions. Several past attempts have been made in this regard. HeritageTogether \cite{heritoge1, heritoge2} offers a web platform for crowdsourcing images and generating \& displaying 3D models. SOCH \cite{soch, sochproto} and Rekrei \cite{rekrei, rekreiLong} were created to crowdsource the reconstruction of CH sites destroyed in natural disasters or due to iconoclasm. CrowdHeritage \cite{ch} and MicroPasts \cite{microp} provide platforms for crowdsourcing and citizen science. However, these platforms are often not fully open-source and lack easy referenceability for data products. Additionally, while platforms like Rekrei connect crowdsourced data to volunteers willing to process the data, the absence of an end-to-end pipeline creates barriers for non-experts and those in resource-constrained settings. To address these issues, we outline the following requirements for a crowdsourced image-collection platform for 3D reconstruction, that can serve to democratize the digital documentation of CH sites:
\begin{enumerate}
  \item \textbf{Crowdsourcing / Participatory DH}: The platform should have an intuitive web interface for crowdsourcing images from the general public and experts alike, while also displaying 3D models, aiding in site discoverability.
  \item \textbf{Cross-Referenceability}: The platform should archive individual reconstructions, assign persistent identifiers for easy referencing, and offer an API for research use.
  \item \textbf{Free and Open-Source Software (FOSS)}: The platform should be FOSS, built entirely on FOSS technologies, fostering a community and ensuring sustainability.
  \item \textbf{End-to-end (E2E) Automation}: The platform should require minimal input, making it user-friendly and adaptable to resource-constrained settings.
\end{enumerate}
Tirtha, which we describe in detail in the next section, is designed to meet all these specifications. Table~\ref{tab:chart} compares Tirtha with existing platforms that fulfill some of these requirements. Currently, Tirtha is the only active platform of its kind for 3D digital reconstruction of cultural heritage sites.

\section{Tirtha}
\label{sec:appr}
Broadly, Tirtha consists of three components: crowdsourced image collection of CH sites (frontend or presentation layer), 3D reconstruction of the sites using photogrammetry (backend or application layer), and storage of the images, 3D models \& associated information (data layer). All of these components use FOSS libraries and Tirtha is itself FOSS, with the code available on GitHub\footnote{https://github.com/smlab-niser/tirtha-public/}. In the following subsections, we discuss each of these components in detail.

\subsection{Presentation Layer}
Tirtha is made available through a single-page application (SPA), where users can upload images of CH sites and view or download the 3D models. The website uses \chref{https://www.djangoproject.com/}{Django}, a Python-based web framework and \chref{https://www.postgresql.org/}{PostgreSQL} for the backend and database, respectively. Django offers a general site administration site, facilitating database management, including adding CH site entries and metadata, as well as content and user moderation. The website is served using \chref{https://gunicorn.org/}{gunicorn} and \chref{https://www.nginx.com/}{Nginx}. The UI design features a straightforward 3-column layout with project \& model details and other relevant information in the left column; main content (i.e., 3D models, contribution form \& site request form) in the middle column using Google's \chref{https://github.com/google/model-viewer}{\texttt{<model-viewer>}} web component; and a navigation pane in the right column listing available 3D models, along with a search bar. The \texttt{<model-viewer>} component supports all major browsers and integrates with AR platforms like Android's SceneViewer, Apple's QuickLook, and WebXR, allowing users to view the 3D models in augmented reality (AR), if supported by their device. Given our focus on crowdsourcing and Tirtha being mobile-first, the website must be accessible from smartphones and so, this layout is fully responsive and adapts to the screen size of the device. Screenshots of the website are shown in Fig.~\ref{fig:landing}.

To streamline the contributor experience, we have implemented a Single Sign-On (SSO) system. SSO allows users to log in to the website using their existing accounts from external identity providers. Currently, we support Google accounts, but any identity provider implementing the OpenID Connect (OIDC)\footnote{\url{https://auth0.com/docs/authenticate/protocols/openid-connect-protocol}} protocol can be used with Tirtha. This approach minimizes the personal information we store, complying with various privacy laws, while allowing us to identify users for moderation and attribution purposes. We only store email addresses and names provided by the identity provider. Moreover, logging in is not required to view or download low-resolution 3D models. It is required only to contribute images or to access high-resolution versions of the models. External identity providers also offer security features like two-factor authentication and User Managed Access (UMA), ensuring secure user accounts with user-controlled identity data. Additionally, relying on external providers avoids the need for us to build and maintain our own authentication system, reducing security vulnerabilities and overhead, especially for a small team like ours.

\begin{figure}
  \centering
  \resizebox{0.75\columnwidth}{!}{
    \begin{tikzpicture}[
      node distance=1.5cm,
      startstop/.style={rectangle, rounded corners, minimum width=3cm, minimum height=1cm, text centered, draw=black, fill=red!30},
      process/.style={rectangle, rounded corners, minimum width=3cm, minimum height=1cm, text centered, draw=black, fill=orange!30},
      arrow/.style={->, >=stealth, thick}
    ]
    \node (start) [startstop] {Start};
    \node (login) [process, below of=start] {Login with SSO};
    \node (selmod) [process, below of=login] {Select CH site};
    \node (selimg) [process, below of=selmod] {Select images for upload};
    \node (check) [process, right of=selimg, yshift=1cm, xshift=2.5cm, fill=blue!30] {Validate images};
    \node (compress) [process, below of=check, yshift=-0.25cm, fill=blue!30] {Compress images};
    \node (tandc) [process, below of=selimg] {Accept terms and privacy policy};
    \node (upload) [process, below of=tandc] {Upload images};
    \node (end) [startstop, below of=upload] {End};
    \draw [arrow] (start) -- (login);
    \draw [arrow] (login) -- (selmod);
    \draw [arrow] (selmod) -- (selimg);
    \draw [arrow] (selimg) -- (check);
    \draw [arrow] (check) -- (compress);
    \draw [arrow] (compress) -- (selimg);
    \draw [arrow] (selimg) -- (tandc);
    \draw [arrow] (tandc) -- (upload);
    \draw [arrow] (upload) -- (end);
    \end{tikzpicture}
  }
  \caption{Contribution flow. Here, the steps in the lavender boxes are completely client-side.}
  \Description{A flowchart showing the contribution flow for Tirtha, with the steps in the lavender boxes being completely client-side. The flow starts with the user logging in with their Google account, followed by selecting the CH site, selecting images, which are then validated and compressed, accepting the terms and privacy policy, and finally, uploading the images.}
  \label{fig:cont_flow}
\end{figure}
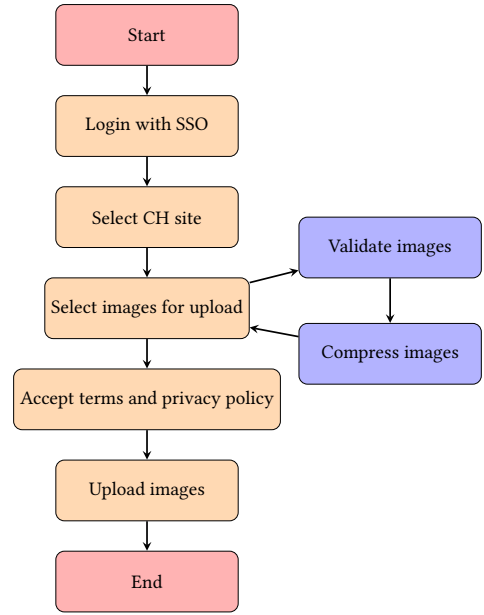

Our upload form enhances the contributor experience with features like single-input CH site selection, drag \& drop image upload, image preview with an option to unselect individual images, and a progress bar with clear status messages. We perform client-side checks on images, such as EXIF data validation, file type validation, and conversion to JPEG for consistency. We also validate image resolution and apply EXIF-preserving compression to reduce bandwidth usage by up to 85\% and improve upload time, benefiting users with slow internet connections and limited data plans. The contribution flow is depicted in Fig.~\ref{fig:cont_flow}. In the future, we plan to assist contributors in focusing on missing or poorly covered parts of sites. One approach is to overlay the relative poses on the 3D model, which would highlight the areas with insufficient coverage. Augmenting this with the cardinal directions can then guide the contributors to take pictures from the correct angles. Another approach is to use GIS to mark those areas, although GIS data resolution may be a limiting factor.

An essential aspect of crowdsourcing is incentivizing contributors. Crowdsourcing in Tirtha incentivizes contributors by involving them in documenting CH sites, that hold cultural and historical significance to them but might have been neglected due to a dearth of resources. This process democratizes digital documentation and raises awareness for site preservation. All platform-collected images are licensed under Creative Commons Attribution-NonCommercial-ShareAlike 4.0 International (CC BY-NC-SA 4.0)\footnote{\url{https://creativecommons.org/licenses/by-nc-sa/4.0/}}, while data products are available under Creative Commons Attribution-NonCommercial-NoDerivatives 4.0 International (CC BY-NC-ND 4.0)\footnote{\url{https://creativecommons.org/licenses/by-nc-nd/4.0/}}. Anyone can freely view and download 3D models for research, education, heritage conservation, or other non-commercial purposes. All contributions are credited on the Tirtha website as well as in the metadata for each reconstruction. The dataset of 3D models benefits computer vision researchers in training and evaluating algorithms for 3D reconstruction, semantic segmentation, and other tasks. Integration with social media platforms like Facebook and Instagram is also considered to expand awareness and serve as an additional incentive for contributors. Through this effort, we aim to build a community of experts and non-experts.

\subsection{Application Layer}
Tirtha's pipeline has three main components: image preprocessing, creating 3D models through photogrammetry, and mesh conversion \& compression for web delivery. In the image preprocessing stage, received images are passed through a content safety filter to identify and filter out inappropriate content using a \chref{https://github.com/GantMan/nsfw_model}{pre-trained model}. Currently, the server-side implementation is used, but there are plans to explore running the filter on the client side using TensorFlowJS for improved data privacy and reduced server load. If the local check is inconclusive, the images are sent to the \chref{https://cloud.google.com/vision}{Google Cloud Vision API} for further validation. Based on the content filter results, images are either put in a manual moderation queue or passed to the next stage for image quality assessment (IQA).

The IQA component assesses image brightness \& contrast by calculating the Dynamic Range (DR) and Contrast-to-Noise (CNR) ratio using the \chref{https://github.com/opencv/opencv-python}{OpenCV} Python package. Each image is also passed through a No Reference IQA model, namely \chref{https://github.com/IIGROUP/MANIQA}{MANIQA} \cite{maniqa}, to provide a general evaluation of the image quality. Based on the results, the images are labeled as ``good'' or ``bad''. ``Good'' images proceed to photogrammetry, while ``bad'' images are discarded but kept in the database for research. Contributions are processed on the server immediately after upload to minimize processing time. In the future, we will explore ways to integrate the results from IQA into an image enhancement component, which could adjust image brightness \& contrast and perform denoising, deblurring, and color correction to ensure good image quality, even in challenging conditions. This is particularly important for crowdsourced images, as the quality of the images can vary significantly, depending on the camera used, the lighting conditions, and the skill of the photographer.

The second component, i.e., the photogrammetry pipeline, yields 3D models of the CH sites. Photogrammetry is the process of creating dense 3D models from 2D images. The current state-of-the-art encompasses chaining techniques such as Structure from Motion (SfM) \cite{sfm} and Multi-View Stereo (MVS) \cite{mvs}. There are several steps involved here, such as feature extraction, image matching, feature matching, incremental SfM, dense reconstruction, meshing and texturing. Each of these steps can be performed using a variety of methods and are complex research areas in their own right. So, for this project, we have chosen to abstract away the complexities, and instead, leverage the comprehensive implementations provided by AliceVision Meshroom, which is a free and open-source photogrammetry suite, developed by the AliceVision project \cite{aV}. It uses state-of-the-art techniques and is sufficient for most use cases, including ours, which includes large-scale scene reconstruction from forward-facing views \cite{maincomp, avcomp, comp2, 23Dornot, rebar}. It is also one of the most popular FOSS photogrammetry software and is widely used by the general research community \cite{uav, insect}. 

\begin{figure*}
  \centering
  \includegraphics[width=\linewidth]{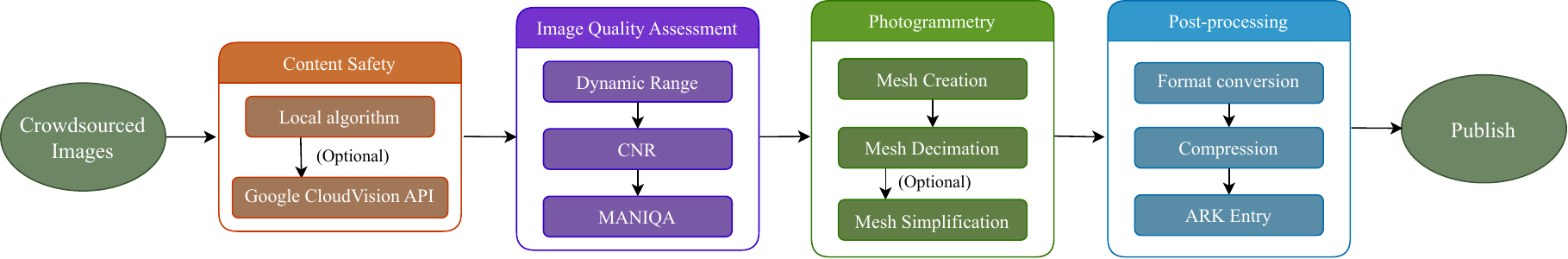}
  \caption{Overview of the application layer}
  \Description{Overview of the application layer, showing various components and subcomponents.}
  \label{fig:appli}
\end{figure*}

A notable aspect of Meshroom is that it exposes a vast number of parameters through its node graph interface (GUI), facilitating a high degree of customization. It also provides precompiled executable binaries for each of the steps, which can be used as standalone tools, and Python scripts to run the binaries as part of batch jobs. Many of the nodes support CUDA, which makes it possible to run them on a GPU, speeding up the reconstruction significantly. However, the built-in scripts impose presets with limited flexibility and are subject to the release cycle of the project for stable updates. So, for our use case, we have written a wrapper library in Python over the binaries compiled from the AliceVision Meshroom source code, which allows us to easily integrate their implementations into our pipeline, while also limiting the scope of the pipeline to the features we require. Creating our own wrapper library also lets us maintain a more stable pipeline, unaffected by the release cycle of AliceVision, and adapt the pipeline, as needed. These modifications may appear as individual nodes that implement newer techniques for instance, or to improve the performance of the pipeline by optimizing the node parameters and parallelizing the pipeline. The Meshroom nodes used in this project, have been listed in Table~\ref{tab:meshnodes}, along with brief descriptions and non-default parameters. We direct the interested reader to the Meshroom documentation\footnote{\url{https://meshroom-manual.readthedocs.io/en/latest/}} for more details on the individual nodes and their parameters. Note that, since our main focus is on web delivery of 3D models, primarily to smartphones with limited VRAM and compute, which is where we expect most people to access the website, we maintain a heavily decimated copy of the mesh with downsampled textures. This is the version that is served via the Tirtha website. Users can request high-quality textured meshes through a simple form on the website.
\begin{table*}
  \centering
  \caption{Meshroom nodes used in Tirtha's photogrammetry pipeline}
  \label{tab:meshnodes}
  \begin{tabular}{lllp{2.5in}}
    \toprule
    \textbf{Operation} & \textbf{Node} & \textbf{Non-default Parameters} & \textbf{Description} \\
    \midrule
    \multirow{5}{1.5cm}[-0.5em]{Feature Extraction \& Matching} & CameraInit & None & Initializes the cameras using the given preset \\
    & FeatureExtraction & None & Extracts features from the images \\
    & ImageMatching & None & Finds collocated images \\
    & FeatureMatching & None & Matches features between collocated images \\
    & StructureFromMotion & None & Computes camera poses \& the sparse point cloud \\
    & SfMTransform & Site-dependent, see code & Auto-orients the point cloud \\
    \midrule
    \multirow{5}{1.5cm}[0.5em]{Depth Map Estimation \& Meshing} & PrepareDenseScene & None & Undistorts images for dense reconstruction \\
    & DepthMapEstimation & None & Computes per-pixel depth maps \\
    & DepthMapFilter & None & Forces depth consistency for overlapping maps \\ 
    & Meshing & `estimateSpaceMinObservationAngle=30' & Computes a dense mesh from the depth maps \\
    \midrule
    \multirow{5}{1.5cm}[-0.2em]{Mesh Processing} & MeshFiltering & `keepLargestMeshOnly=1' & Removes disconnected components \\
    & MeshDecimate & `simplificationFactor=0.3' & Decimates raw mesh \\
    & MeshDenoising & `lmd=2,eta=1.5' & (Optionally) Denoises mesh \\
    & MeshResampling & `simplificationFactor=0.3' & (Optionally) Resamples mesh \\
    & Texturing & `textureSide=2048' & Computes mesh textures \\
    \bottomrule
  \end{tabular}
\end{table*}
The final stage of the pipeline entails post-processing the 3D models. It includes converting the mesh format to `.glb' or `.glTF', using the \chref{https://github.com/CesiumGS/obj2gltf}{obj2gltf} utility from CesiumJS, and compressing the mesh and textures with \chref{https://github.com/zeux/meshoptimizer/}{MeshOptimizer}. The format conversion is due to \texttt{<model-viewer>} not supporting the `.obj' format from Meshroom, and the ubiquity and ease-of-use of the `.glb' and `.glTF' formats. Notably, this would allow us to add 3D Tiles\footnote{\url{https://github.com/CesiumGS/3d-tiles}} streaming, which should greatly improve the performance of the website, by loading only the required portions of the 3D model, as a user navigates the model. A high level of mesh compression using `MeshOptimizer' speeds up loading and reduces bandwidth usage, up to $\approx80\%$ in some of our experiments. The final step is to create an entry in the Archival Resource Keys (ARK) database\footnote{\url{https://arks.org/}}, which attaches a globally unique and persistent identifier to each successful reconstruction of a CH site. The ARK entry is created using a customized version of the ARK generator and parser provided by the \chref{https://github.com/internetarchive/arklet}{arklet} utility, which is a Django application for creating and resolving ARK entries. The finalized 3D model is then published on the website for viewing and download. The major steps in the pipeline are shown in Fig.~\ref{fig:appli}.

While the Tirtha pipeline is completely automated, it does permit manual tweaks to certain aspects of the reconstruction. For instance, one can select a `center image' in the Django admin panel to align the camera poses, and can specify pose rotation if auto-orientation fails. Similarly, denoising and resampling are optional, while the simplification factor and a minimum observation angle can be prescribed. In our experiments, monuments with fine details, such as intricate carvings, are better represented by the raw mesh, while those with smooth surfaces benefit from denoised and resampled meshes. This is because denoising, resampling and large values for the minimum observation angle may lead to loss of detail and introduce artifacts, such as holes. Hence, we have made these processes configurable to present each monument in the best way possible.

The pipeline utilizes \chref{https://docs.celeryq.dev/en/stable/index.html}{Celery} and \chref{https://www.rabbitmq.com/}{RabbitMQ} for handling all the steps using a distributed task queue. Celery, which is well-integrated with Django, enables asynchronous task execution and queueing with RabbitMQ. Asynchronous processing enhances responsiveness and fault tolerance, while enabling easy scalability by adding more workers or clusters. Celery also supports autoscaling based on load and simplifies scheduling periodic tasks like database pruning, backups, and archiving old runs to a larger storage device, thereby adding to the automation aspect. Since Tirtha is an evolving codebase, unit testing is required and it is handled using \chref{https://docs.pytest.org/en/7.3.x/}{pytest} and automated using \chref{https://github.com/features/actions}{GitHub Actions}.

Tirtha is currently deployed on-premise in a Docker container on a Ubuntu 22.04 LTS server, with 24 vCPUs, 512 GB RAM, 4 TB storage (with larger archive storage), and 1 A100 GPU. However, due to its modular design, setting up Tirtha on cloud platforms like Amazon Web Services (AWS) should be straightforward. In particular, AWS' pay-as-you-go model makes it ideal for resource-limited heritage preservation and 3D documentation operations \cite{aws}.

\subsection{Data Layer}

We use a PostgreSQL database with Django ORM to store information about CH sites, 3D models, and contributed images. We chose PostgreSQL as the relational database for Tirtha due to its structured data capabilities, extensive features, seamless integration with Django, and future integration potential with OpenStreetMap and PostGIS. \chref{https://postgis.net/}{PostGIS} is a PostgreSQL extension that enables storage, querying, and spatial operations on geospatial data, such as CH site locations. This integration can enhance the website's search functionality by locating sites within a specific area. OpenStreetMap also utilizes PostGIS. Previous works have utilized PostGIS for storing and enhancing CH site data \cite{sochproto, soch}. For details about the database and a visual representation, refer to Appendix~\ref{app:data}.

\section{Early Results}
\label{sec:res}
\begin{table*}
  \centering
  \caption{Typical images from Crowdsourced and Expert-sourced data}
  \begin{tabular}{cc|cc}
    \hline
    \multicolumn{2}{c|}{\textbf{Somanatha Temple}} & \multicolumn{2}{|c}{\textbf{Gopinatha Temple}} \\
    \cline{1-2} \cline{3-4}
    \textbf{Crowdsourced} & \textbf{Expert-sourced} & \textbf{Crowdsourced} & \textbf{Expert-sourced} \\
    \hline
    \addlinespace[0.3em]
    \includegraphics[width=2.2cm]{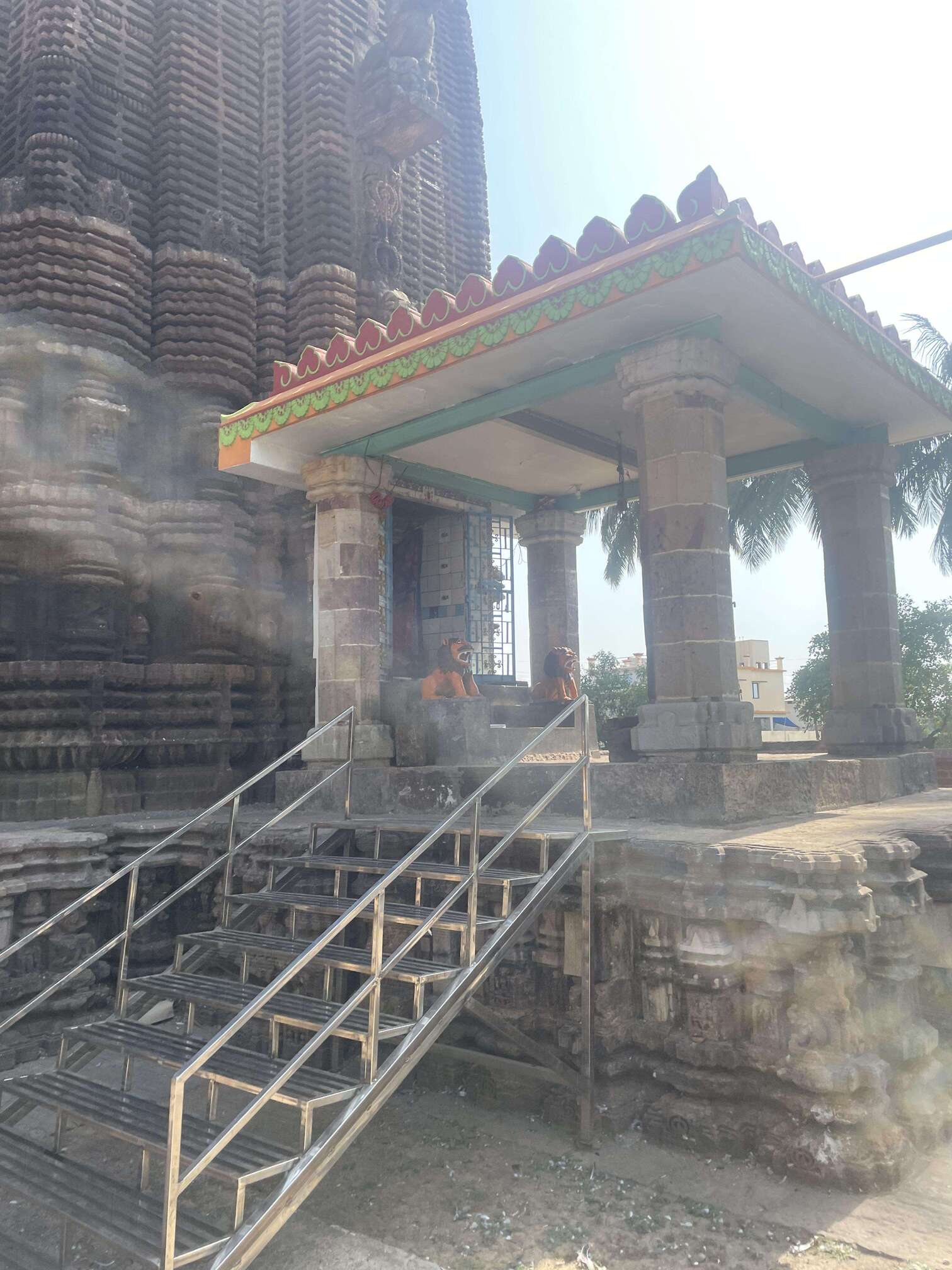} & \includegraphics[width=3.75cm]{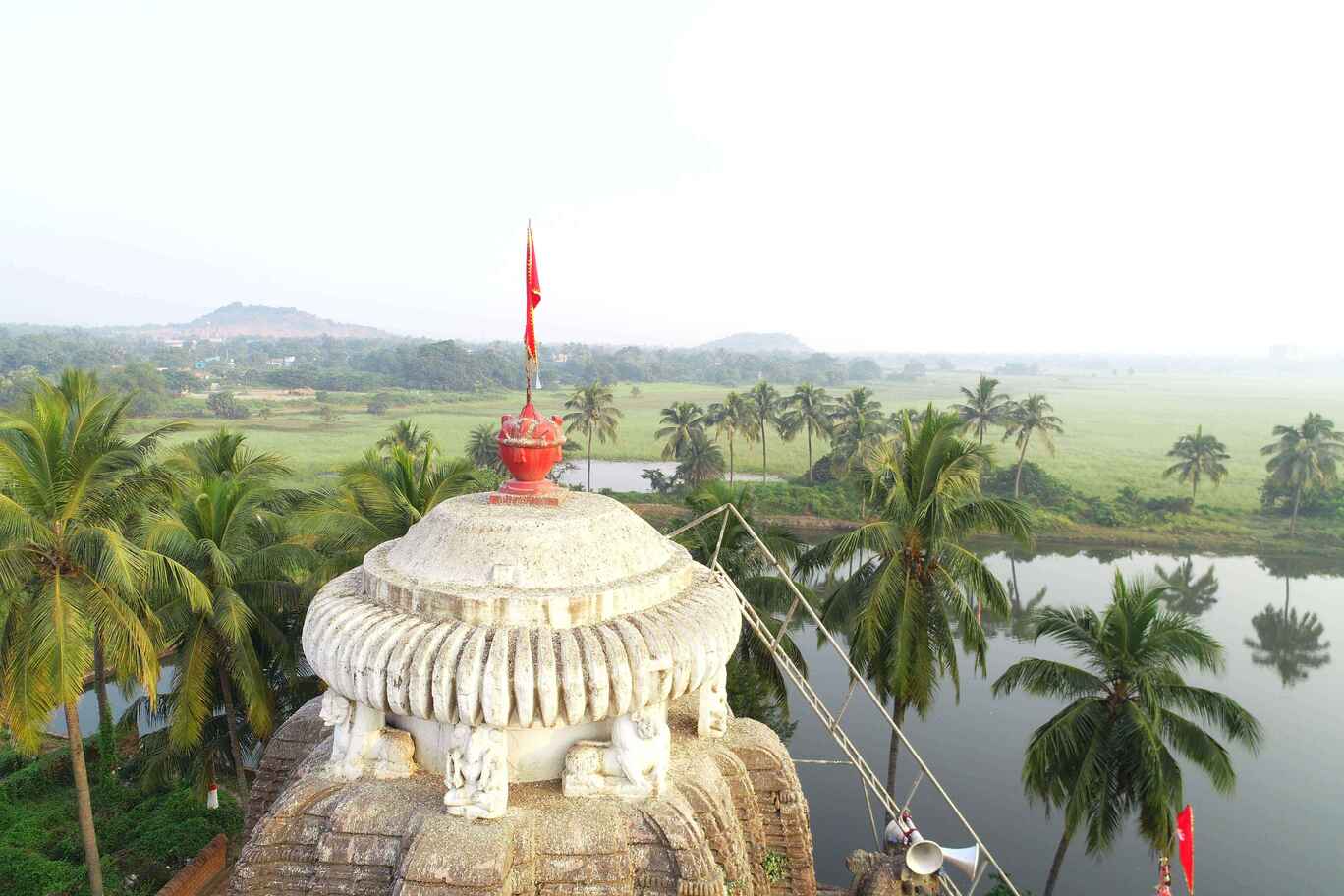} & \includegraphics[width=2.1cm]{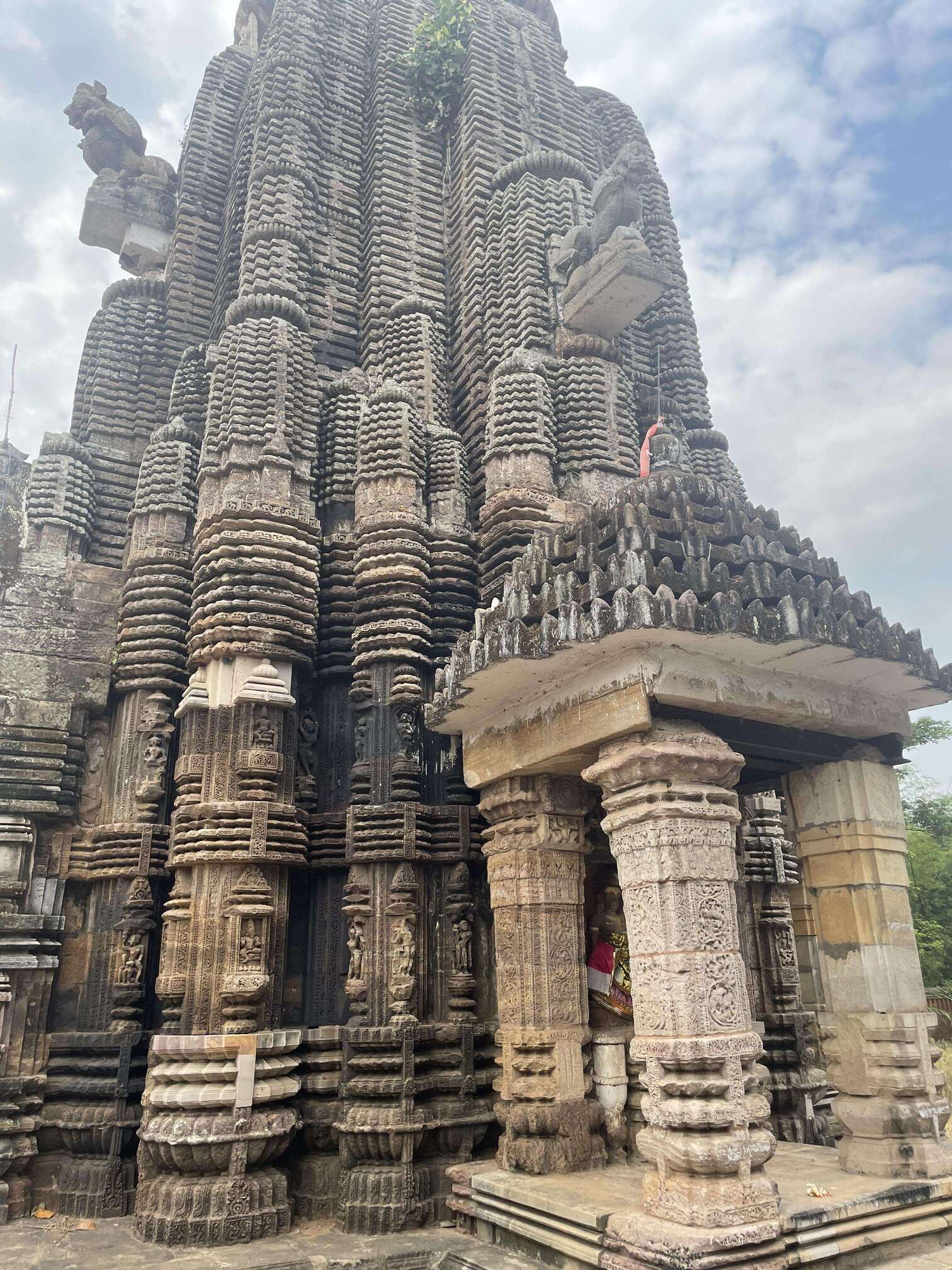} & \includegraphics[width=3.75cm]{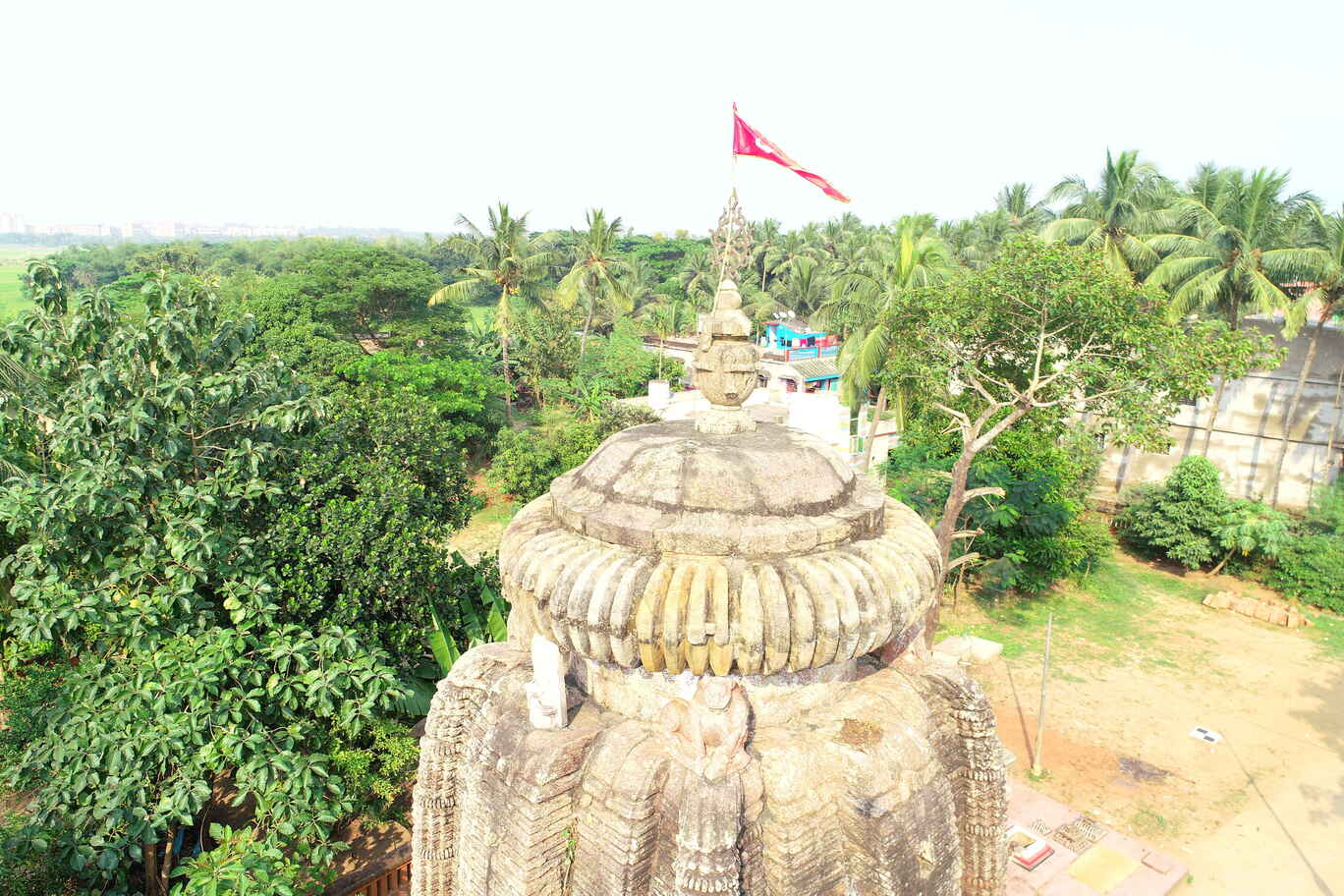} \\
    \includegraphics[width=2.2cm]{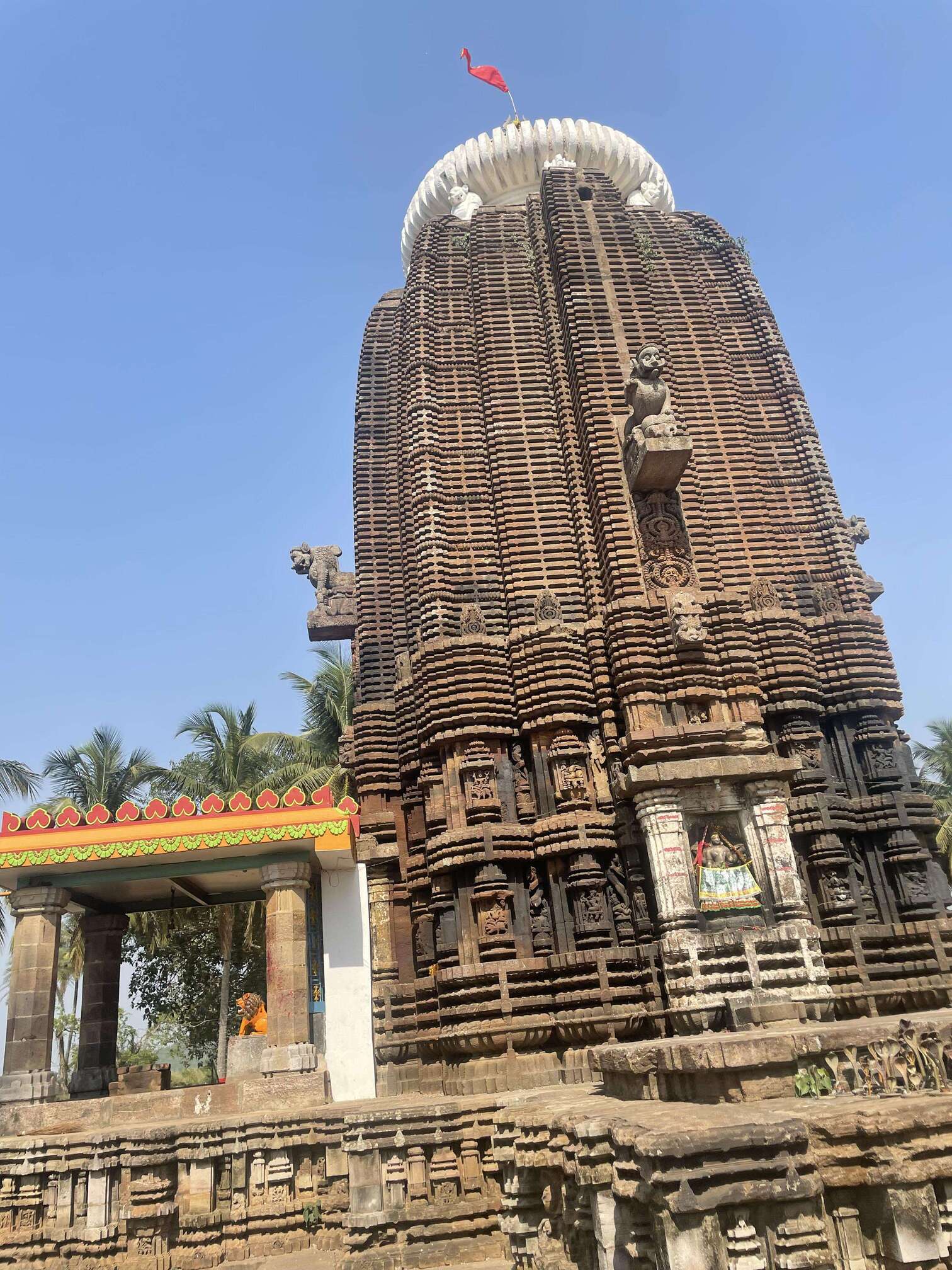} & \includegraphics[width=3.75cm]{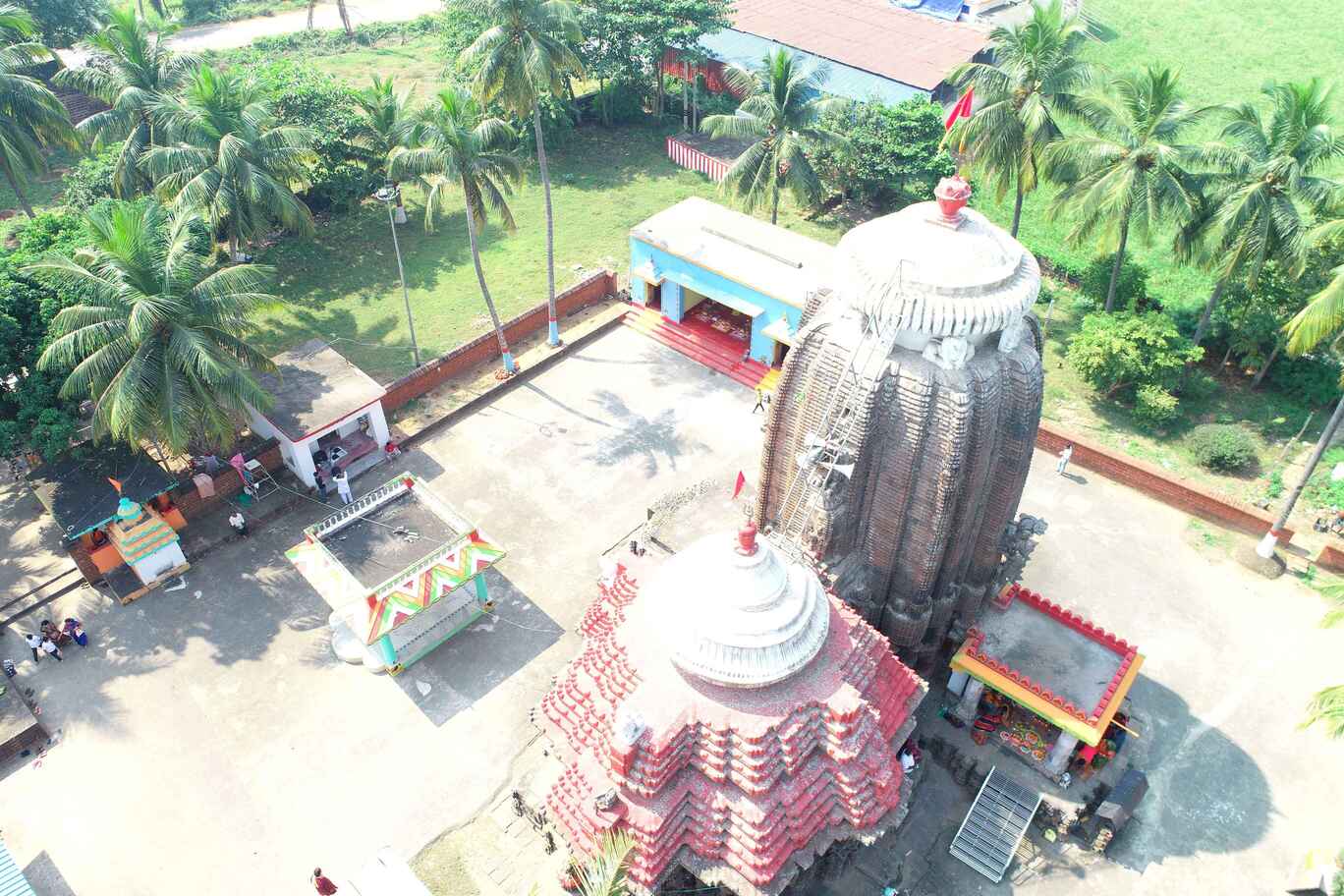} & \includegraphics[width=2.1cm]{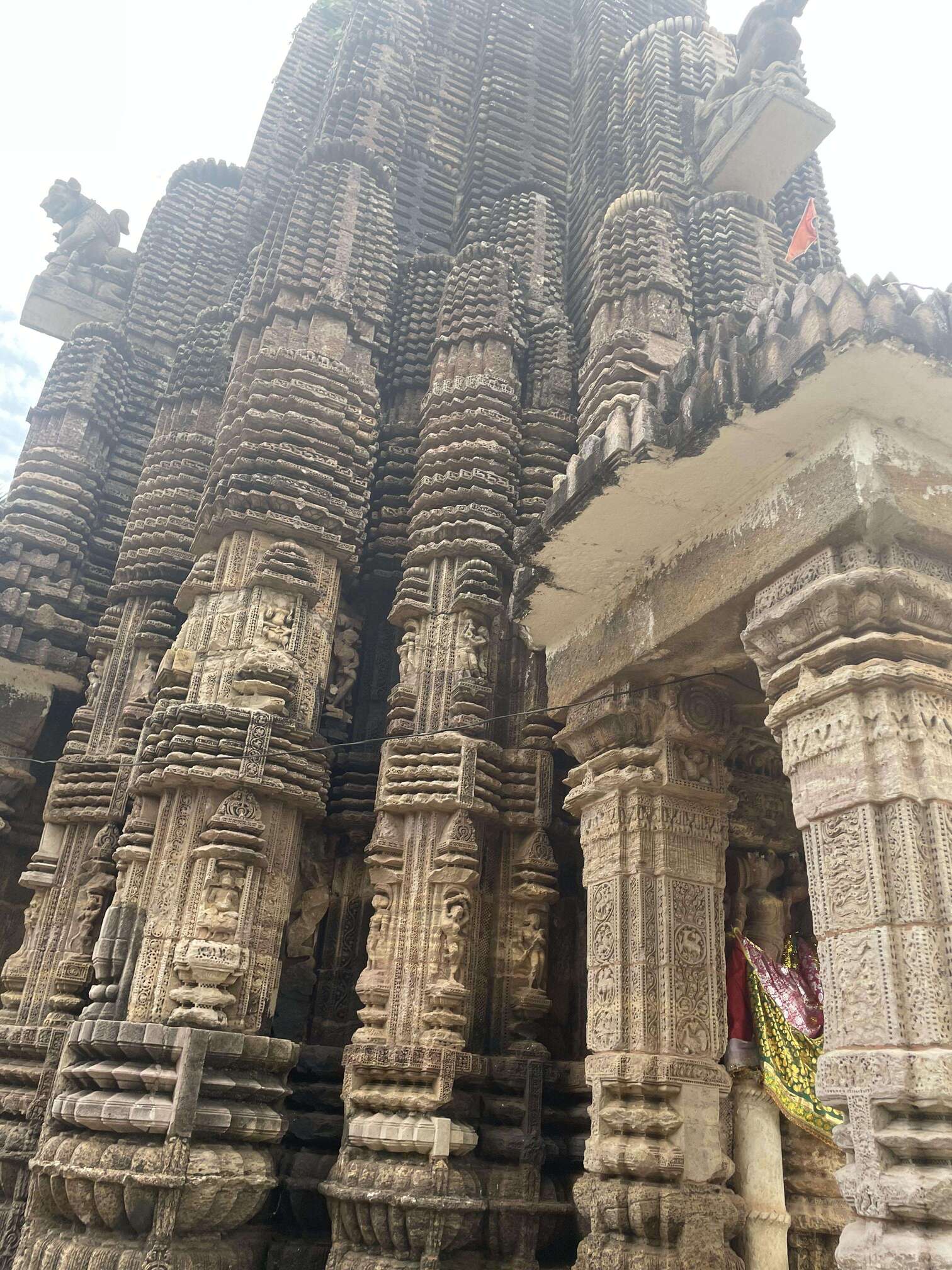} & \includegraphics[width=3.75cm]{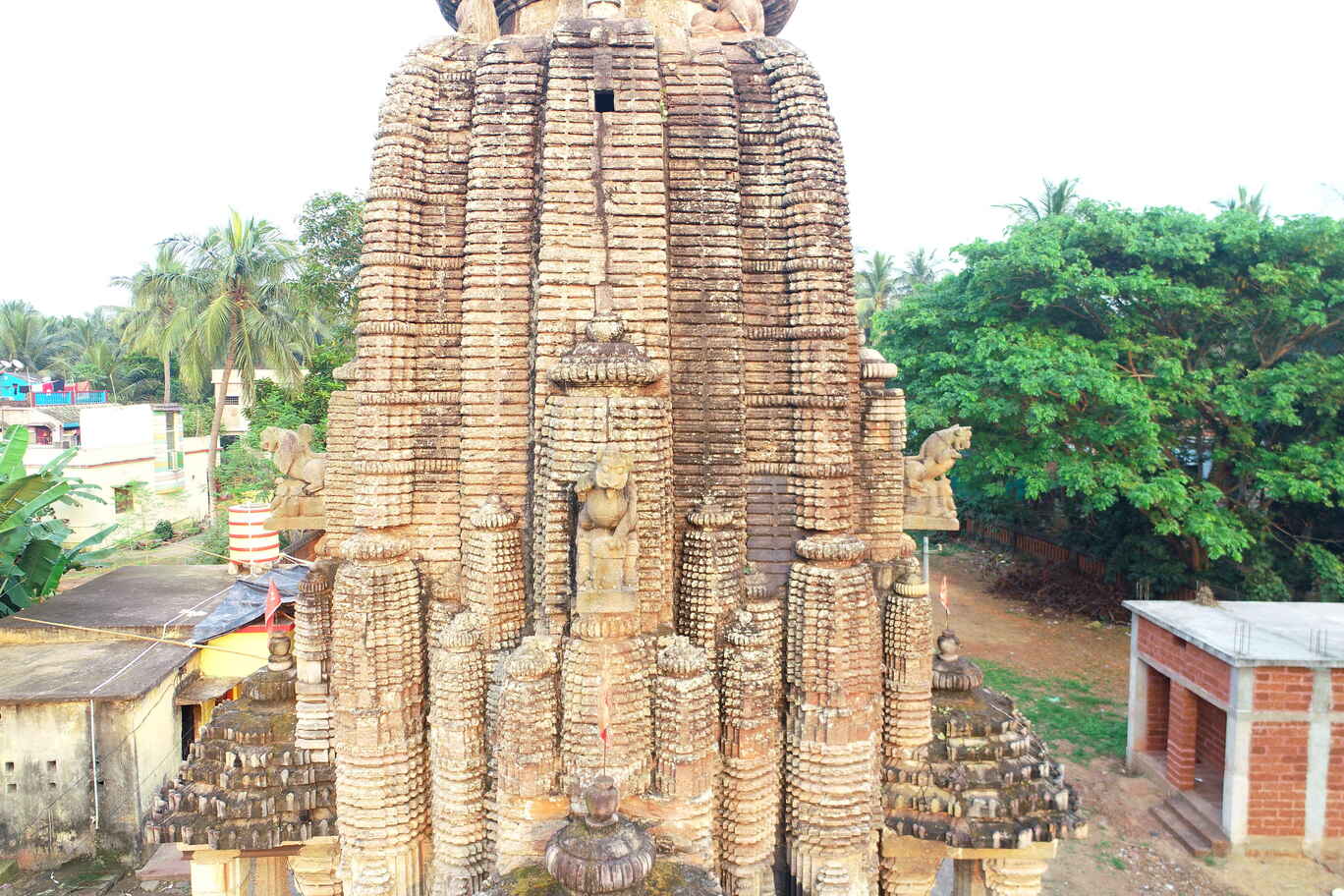} \\
    \hline
  \end{tabular}
  \label{tab:rep12}
\end{table*}

\begin{table*}
  \centering
  \caption{Reconstruction results for both sites}
  \begin{tabular}{c|cccc}
    \toprule
    \multirow{2}{*}{\textbf{CH Site}} & \multicolumn{2}{c}{\textbf{Registered Views}} & \multicolumn{2}{c}{\textbf{Mesh Compression}} \\
    \cline{2-5}
    & \textbf{Crowdsourced Data} & \textbf{Expert-sourced data} & \textbf{Crowdsourced Data} & \textbf{Expert-sourced data} \\
    \hline
    Somanatha Temple & 250 / 251 & 1513 / 1513 & 261 $\to$ 56 MB ($\approx$78.5\%) & 1667 $\to$ 326 MB ($\approx$80.4\%) \\
    Gopinatha Temple & 267 / 268 & 1956 / 1956 & 442 $\to$ 90 MB ($\approx$79.6\%) & 1331 $\to$ 197 MB ($\approx$85.2\%) \\
    \bottomrule
  \end{tabular}
  \label{tab:data}
\end{table*}

In this section, we discuss results from pilot studies conducted on two State Protected Monuments in Odisha, India. These monuments, namely the Somanatha and Gopinatha temples, located in the Khurdha district, have yet to be digitally documented. Built in 13th century AD, both temples are made of rare Baulamala sandstone and exemplify Kalinga architecture. Despite suffering damage and vandalism over time, the remaining carvings showcase intricate patterns and floral motifs. The Somanatha temple is dedicated to Lord Shiva, while the Gopinatha Temple is a Vaishnav monument. We have two sets of datasets for both sites: expert-sourced \& crowdsourced. The expert-sourced datasets were collected using drone mapping, resulting in high-resolution images free from negative attributes such as blur and overexposure. In contrast, the crowdsourced datasets consist of images captured by smartphone cameras without additional equipment, representing typical contributions from volunteers. Examples of these images can be seen in Table~\ref{tab:rep12}. 

\begin{figure*}
  \centering
  \begin{tabular}{ccc}
    \toprule
    Tirtha + Crowdsourced Data & Tirtha + Expert-sourced Data & Expert-sourced Mesh \\
    \midrule
    \begin{subfigure}[b]{0.17\textwidth}
      \centering
      \includegraphics[width=\linewidth]{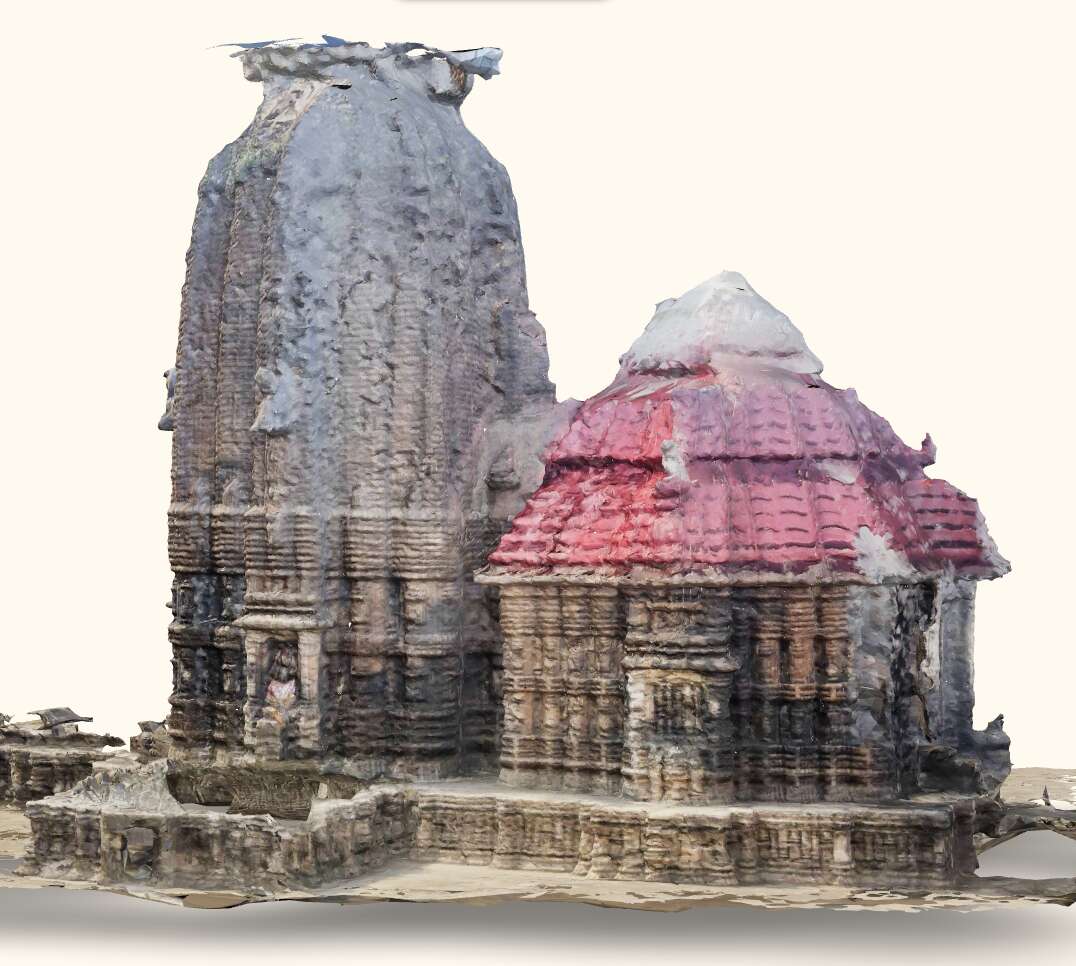}
    \end{subfigure} &
    \begin{subfigure}[b]{0.17\textwidth}
      \centering
      \includegraphics[width=\linewidth]{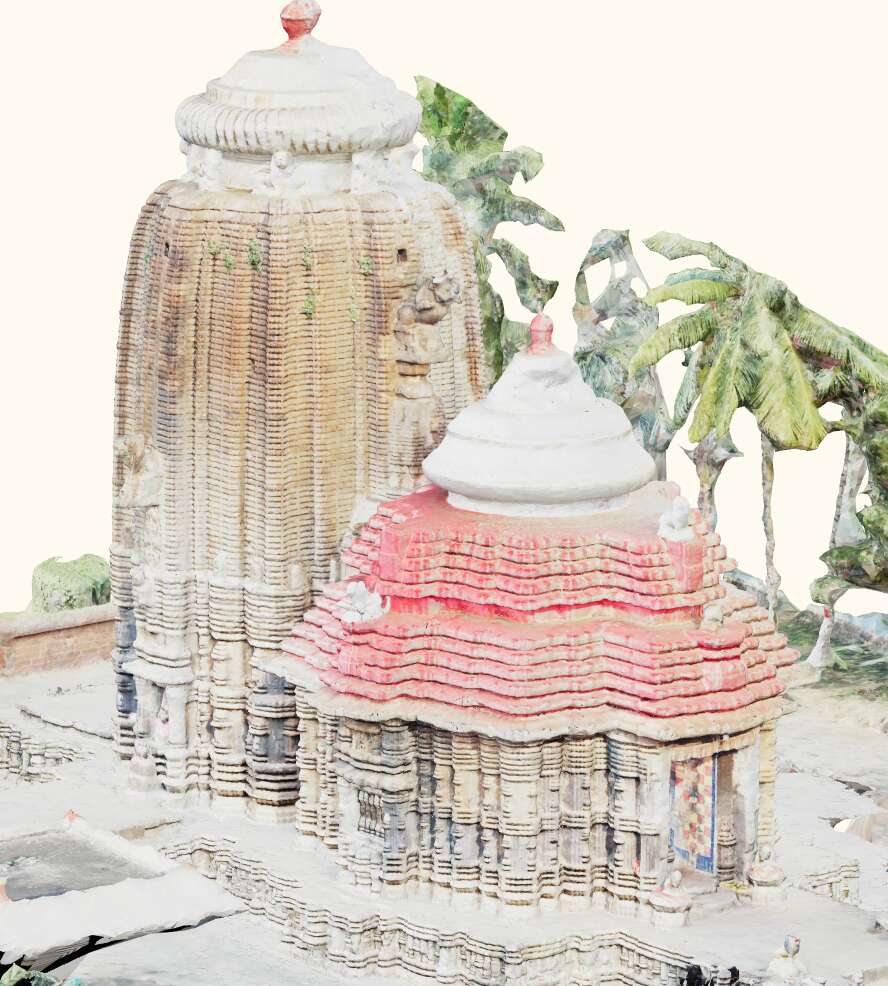}
    \end{subfigure} &
    \begin{subfigure}[b]{0.17\textwidth}
      \centering
      \includegraphics[width=\linewidth]{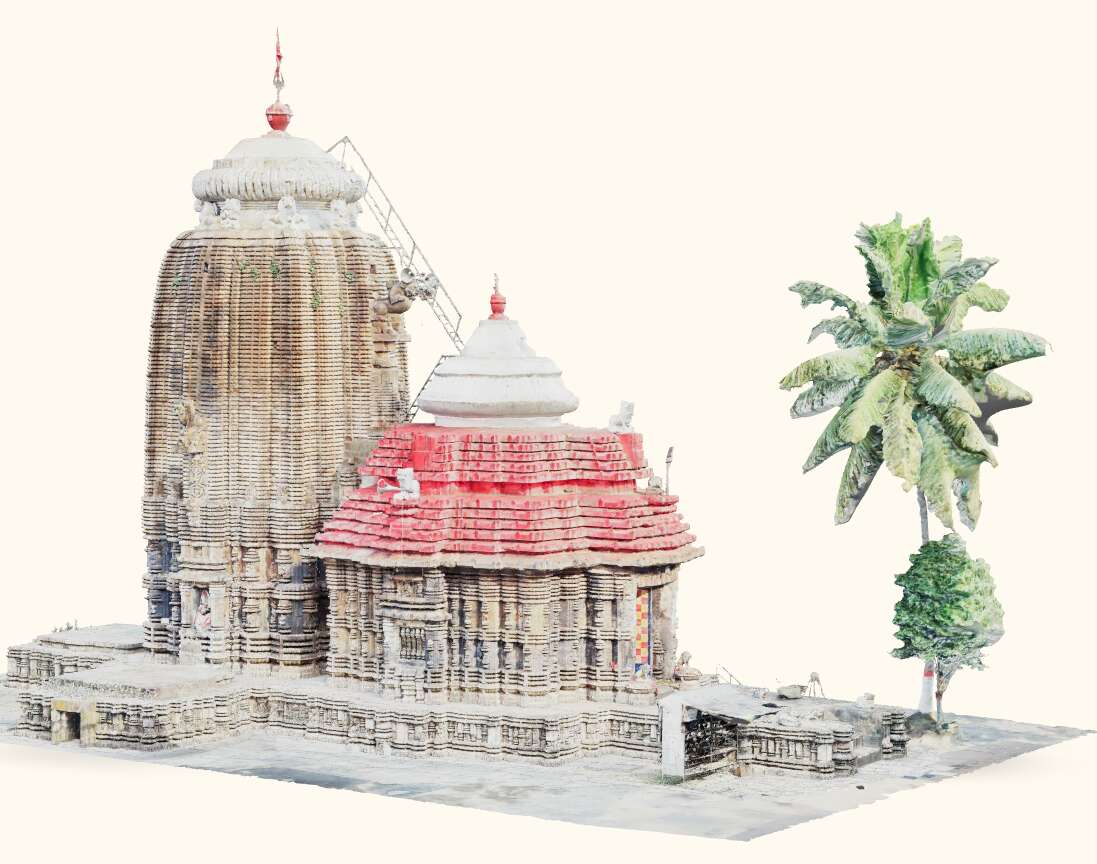}
    \end{subfigure} \\
    \begin{subfigure}[b]{0.17\textwidth}
      \centering
      \includegraphics[width=\linewidth]{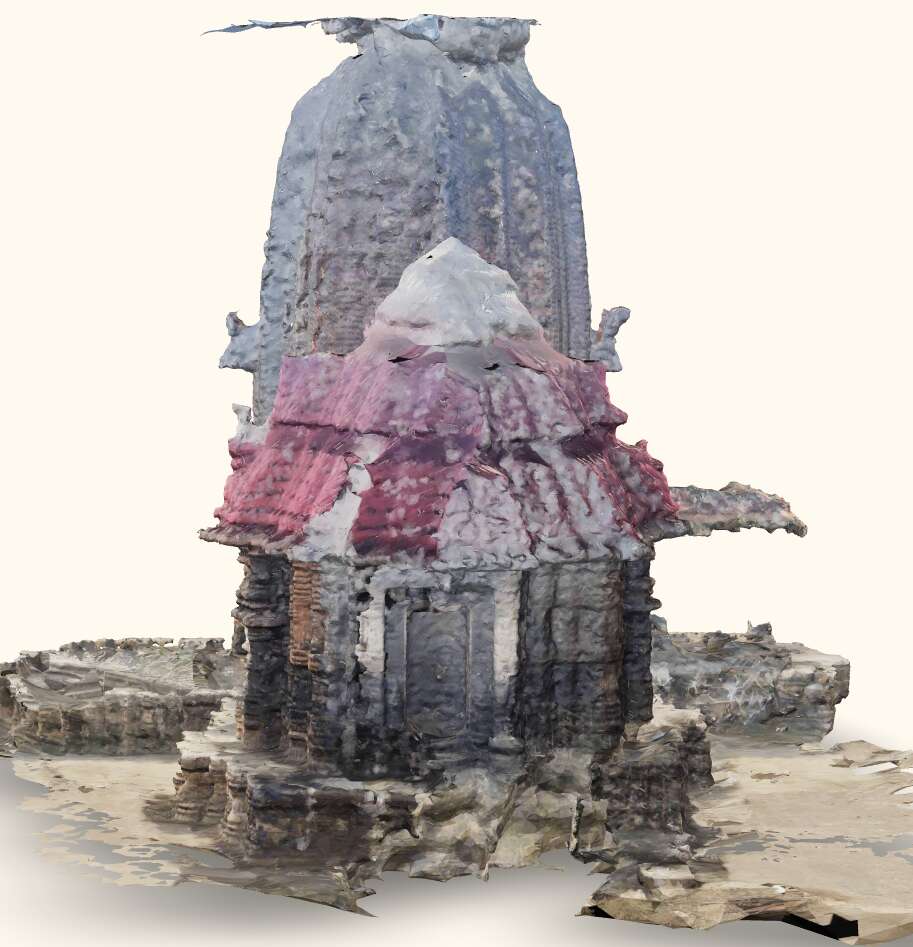}
    \end{subfigure} &
    \begin{subfigure}[b]{0.17\textwidth}
      \centering
      \includegraphics[width=\linewidth]{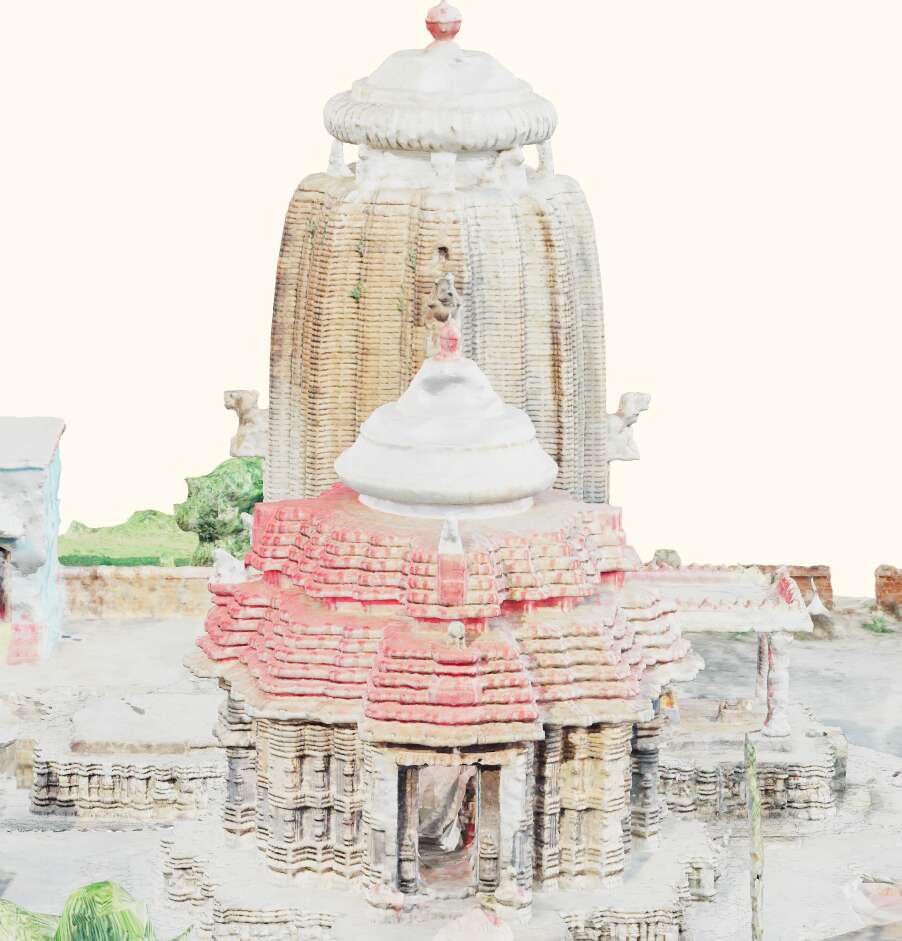}
    \end{subfigure} &
    \begin{subfigure}[b]{0.17\textwidth}
      \centering
      \includegraphics[width=\linewidth]{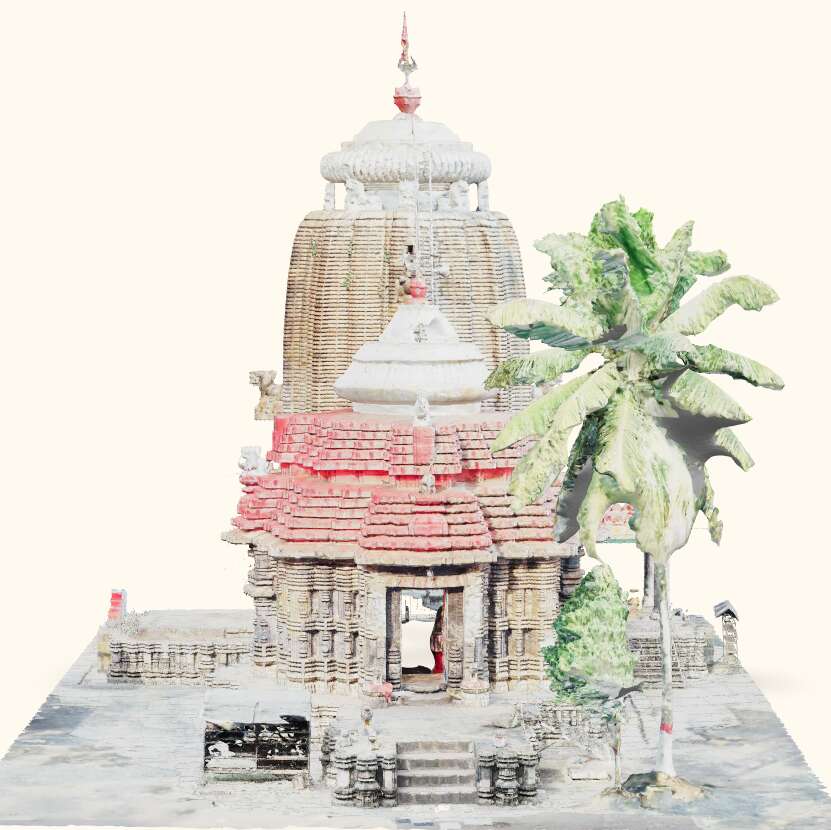}
    \end{subfigure} \\
    \begin{subfigure}[b]{0.17\textwidth}
      \centering
      \includegraphics[width=\linewidth]{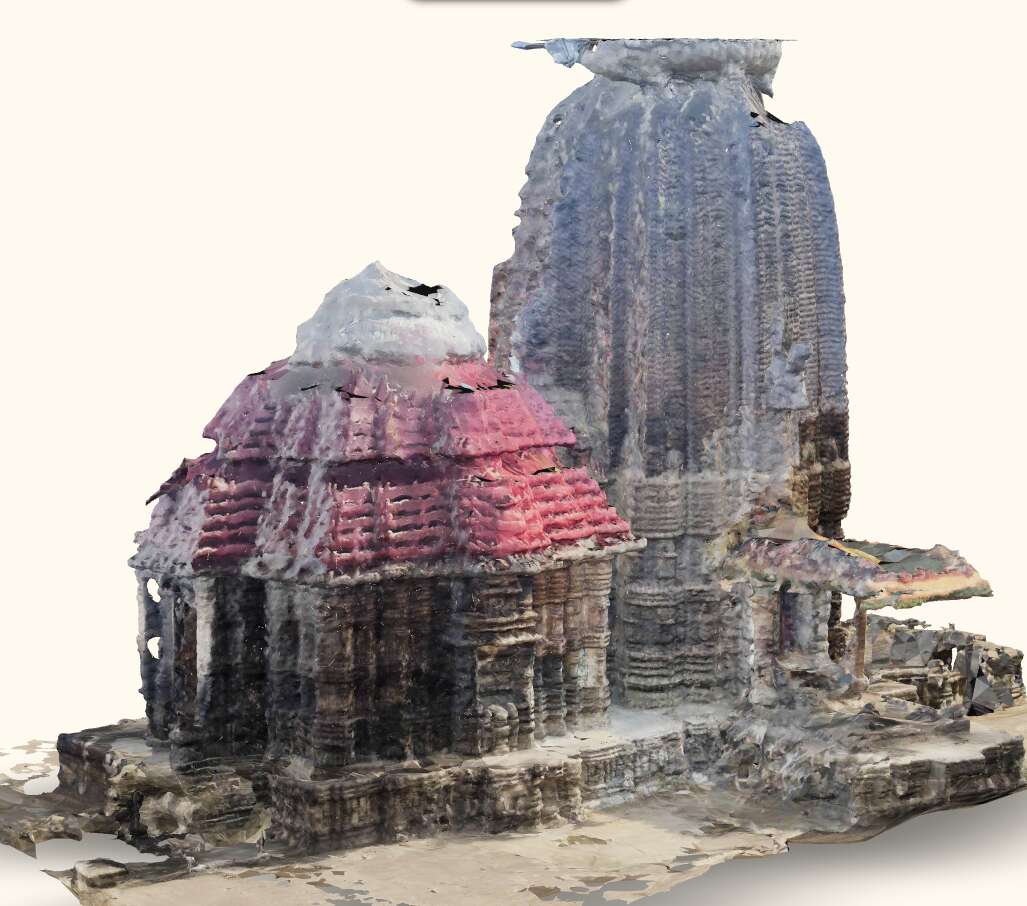}
    \end{subfigure} &
    \begin{subfigure}[b]{0.17\textwidth}
      \centering
      \includegraphics[width=\linewidth]{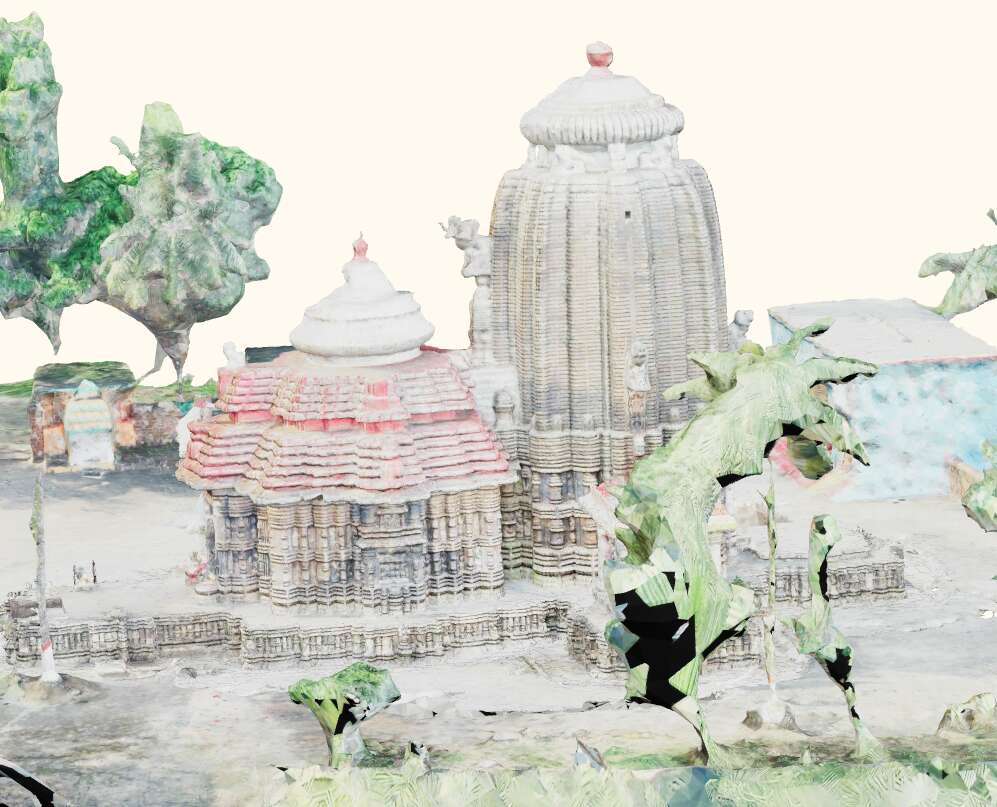}
    \end{subfigure} &
    \begin{subfigure}[b]{0.17\textwidth}
      \centering
      \includegraphics[width=\linewidth]{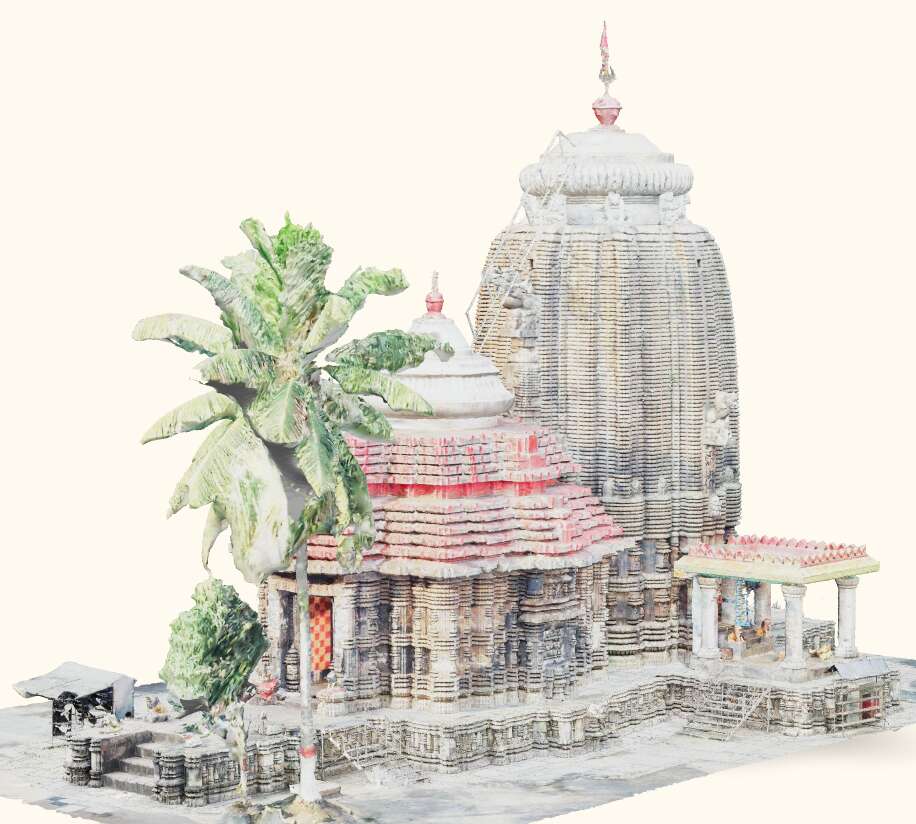}
    \end{subfigure} \\
    \begin{subfigure}[b]{0.17\textwidth}
      \centering
      \includegraphics[width=\linewidth]{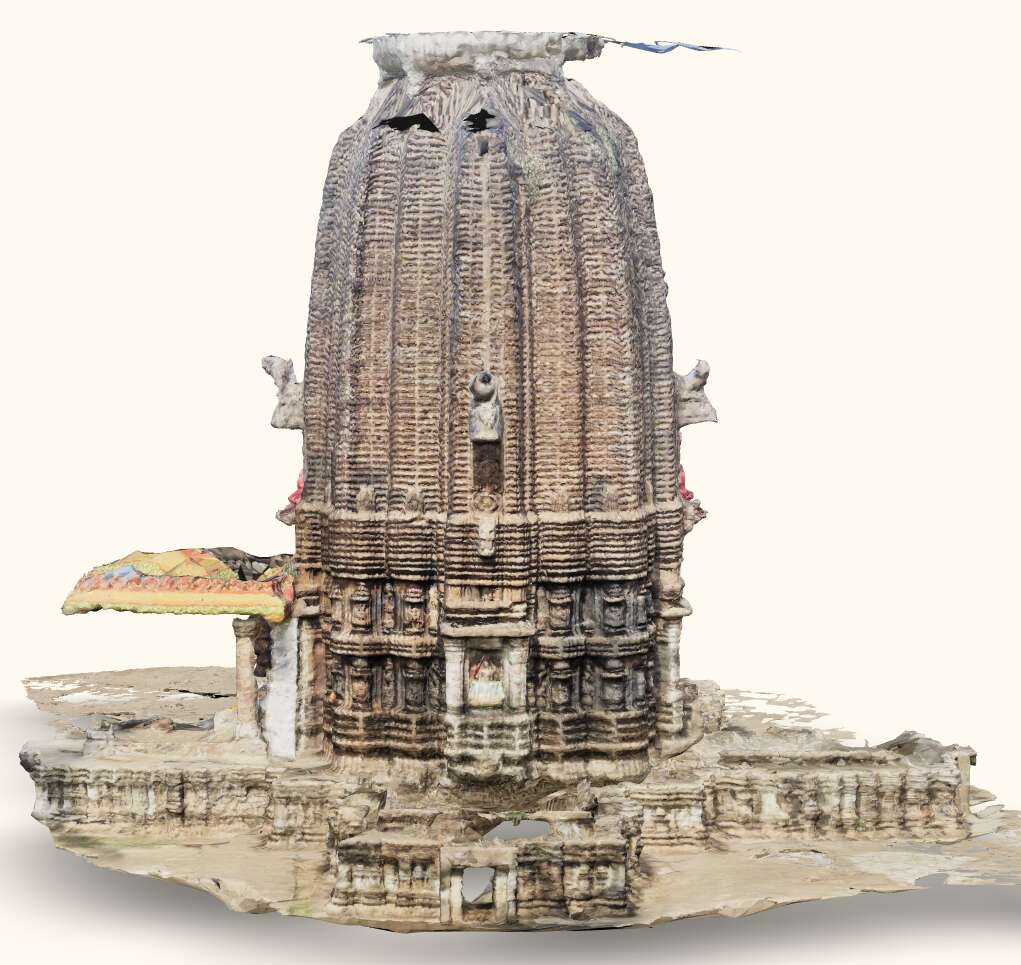}
    \end{subfigure} &
    \begin{subfigure}[b]{0.17\textwidth}
      \centering
      \includegraphics[width=\linewidth]{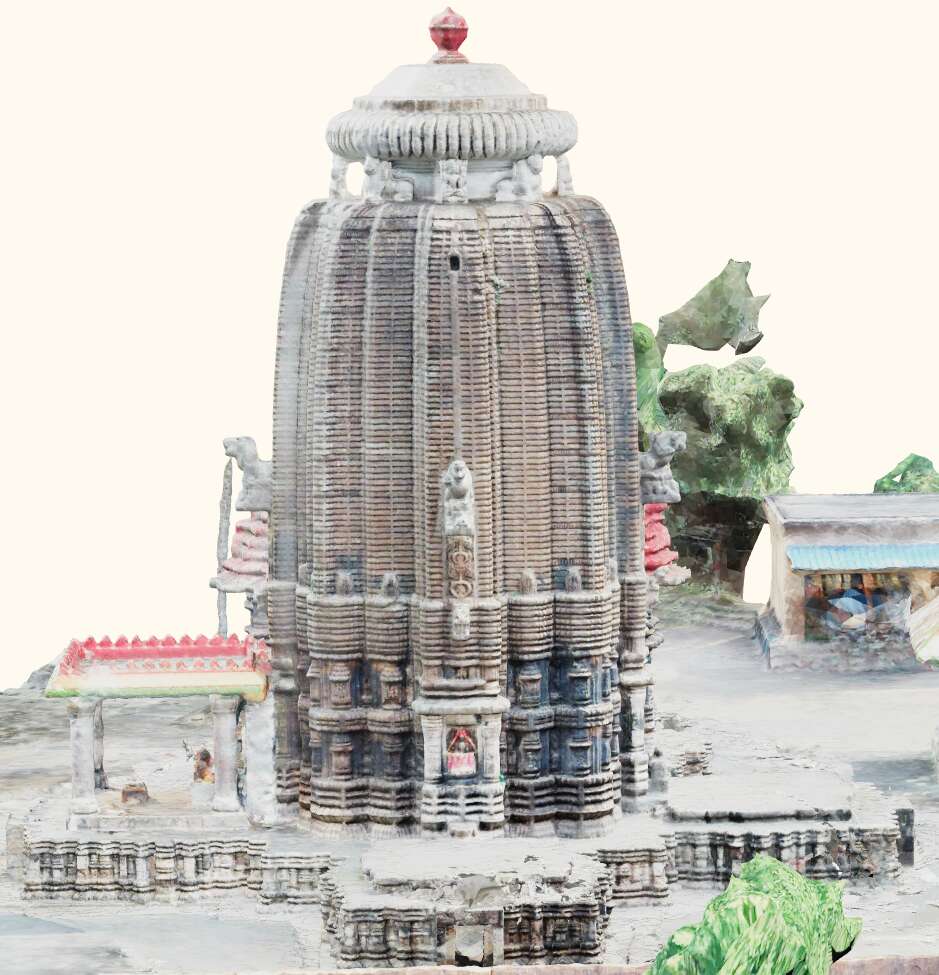}
    \end{subfigure} &
    \begin{subfigure}[b]{0.17\textwidth}
      \centering
      \includegraphics[width=\linewidth]{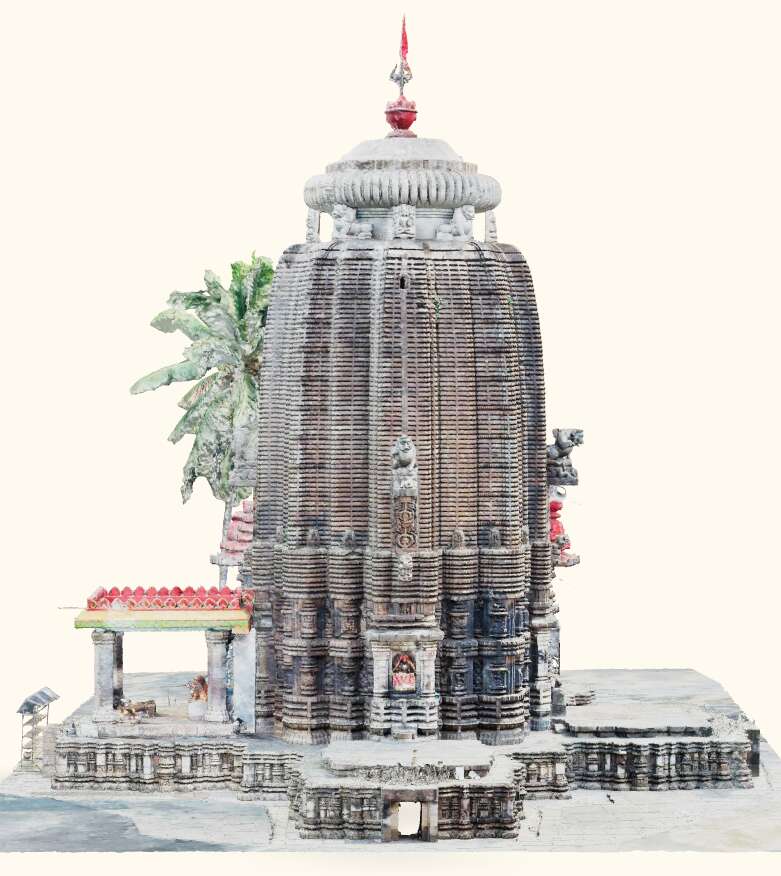}
    \end{subfigure} \\
    \bottomrule
  \end{tabular}
  \caption{Somanatha Temple: Comparing Tirtha's reconstructions on different datasets \& an expert-sourced mesh.}
  \Description{Comparison of 3D models of Somanatha Temple, Odisha, India, generated using Tirtha, with crowdsourced images, expert-sourced images, and an expert-sourced mesh.}
  \label{fig:somacomp}
\end{figure*}

In the context of crowdsourcing for digital heritage, data quality is widely discussed in the literature \cite{dqch}, as volunteers may lack technical expertise. Since the sites we are studying have not been previously documented and lack laser scan-generated ground truth meshes, we assess the quality of the generated mesh based on visual appearance and completeness. We also use the number of registered images in the SfM process as an indicator of the photogrammetry pipeline's robustness. Table~\ref{tab:data} summarizes these metrics for each case study. Despite some issues with the crowdsourced data, such as blur, lens glare, overexposure, and skewed angles, the SfM process successfully registered and posed almost all views correctly. In contrast, the expert-sourced data, which had superior quality, exhibited none of these issues, resulting in all poses being accurately calculated. This demonstrates the robustness of <Platform>'s photogrammetry pipeline. Additionally, we have benchmark 3D models for both temples generated using expert-sourced data via a proprietary software called \chref{https://www.agisoft.com/}{Agisoft Metashape}. Fig.~\ref{fig:somacomp} provides visual comparisons of the 3D models for Somanatha. The comparison for Gopinatha is available in Fig.~\ref{fig:gopicomp} in Appendix~\ref{app:appres}. The <Platform>-generated mesh on expert-sourced data is nearly indistinguishable from the expert-sourced mesh, indicating high quality. Fig.~\ref{fig:somazoomin} in Appendix~\ref{app:appres} compares the level of detail between these meshes, further confirming this observation. However, the quality of the <Platform>-generated mesh on crowdsourced data is not as good due to the aforementioned issues. Furthermore, the lack of comprehensive coverage from all angles hinders accurate reconstruction. The images in the crowdsourced dataset primarily capture the ground level, requiring the cameras to be tilted at various angles to capture the upper areas of the temples. This leads to unreliable depth maps, resulting in mismatched areas and skewed texture mapping in the mesh, distorting the intricate carvings and resulting in disconnected regions and floaters. Moreover, incomplete coverage of the top and inner areas of the temples in the crowdsourced data leads to incomplete meshes. Figure~\ref{fig:gopibadcomp} and Fig.~\ref{fig:somabadcomp} in Appendix~\ref{app:appres} showcase this. Despite these challenges, the photogrammetry pipeline produced reasonable approximations for both monuments using crowdsourced data, even with a far smaller number of images compared to expert-sourced data and suboptimal coverage of the sites. This suggests that with more data, the reconstruction would improve incrementally, making crowdsourcing a viable approach for this task. Additionally, with tighter thresholds for the IQA steps, the results improve, especially in terms of image registration and alignment in the photogrammetry stage, as detailed in Appendix~\ref{app:iops}.

Another important test for Tirtha is whether the simplified (decimated \& compressed) 3D models are suitable for web delivery in terms of mesh quality and file size. From the images presented here, we see that the meshes remain detailed even after decimation, while from Table~\ref{tab:data}, we note that `MeshOptimizer' compression leads to compact filesizes enabling efficient web delivery. Tirtha also keeps the raw, undecimated mesh for each run, available upon request. The following section highlights Tirtha's limitations and future plans.

\section{Discussion}
\label{sec:disc}

\subsection{Limitations}
\label{sec:lim}

It is clear that the quality of the final 3D model depends heavily on the input data. To improve the models, we can either guide volunteers to capture better images or enhance the image quality after they have been uploaded. Currently, we are exploring the latter using image preprocessing techniques. However, the compressed nature of the submitted images makes tasks like color correction without introducing artifacts challenging. We are also looking into ways to guide volunteers to capture better images, such as providing them with best practices or integrating relative mesh orientations with GIS data. Despite our efforts, some imperfections will remain, particularly for large structures with hard-to-reach areas, resulting in incomplete portions or ``holes'' in the models. Adding more pictures can improve the models but it also increases computational costs. Moreover, processing larger datasets takes a significant amount of time, depending on the number of images and scene complexity. We have parallelized Tirtha and optimized AliceVision to mitigate this, although the chosen parameters may not be optimal for all cases. Furthermore, as one of Tirtha's goals is to create a 3D model dataset for research purposes, the main challenge lies in quantifying the quality of the crowdsourced data and corresponding 3D reconstructions. For sites where higher resolution scans are not feasible or available, and for sites lost due to natural disasters or human activities, there is no ground truth to assess accuracy. Instead, we can only compare the visual quality of the reconstructions to the input images and evaluate their completeness and fidelity. While this remains a major limitation, it should be noted that even low-quality reconstructions serve the purpose of conserving the memory of the site and facilitating its study, even if they lack metric accuracy or completeness. Another challenge is compressing and delivering the meshes over the web. While MeshOptimizer can achieve substantial compression ratios, decompression requires significant VRAM, which may exceed the limits of mobile devices like smartphones, leading to out-of-memory errors during mesh loading. Lastly, crowdsourcing remains an ongoing challenge for Tirtha, particularly in terms of volunteer incentives and continued engagement. So far, we have successfully crowdsourced images for only a few sites due to factors like limited awareness about the platform, lack of interest in listed sites, or restricted access. We are actively working on addressing many of these issues, as discussed in the next subsection.

\subsection{Future Work}
\label{sec:fw}
In previous sections, we discussed the benefits of incorporating GIS and 3D Tiles streaming into the platform for improved immersion and responsiveness. To achieve this, we need to finalize the Tirtha API to support the integration of GIS data and to further support the research community. Native mobile apps are also being considered to serve as guides for contributors during image capture, offering more flexibility compared to web apps. On-the-edge processing within the apps could filter out poor-quality images and provide immediate feedback, leading to better model reconstruction. Leveraging LIDAR technology \cite{lidar, smartphones} and ARKit/ARCore could further enhance functionality. For now, the project can be deployed as a Progressive Web App (PWA) with support for image acquisition and on-the-edge processing. Additionally, investigating advancements in photogrammetry pipelines, such as Neural Radiance Fields (NeRFs) \cite{nerf} and Neural Surface Reconstruction \cite{neuralangelo}, could offer novel solutions, although computational requirements and web rendering challenges remain \cite{mobilenerf, nerfw}. It would also be interesting to explore image inpainting and texture generation techniques \cite{bhid, bhid2, hampibook1} to generate partial textures or fill gaps in the reconstructed meshes. We are optimistic about the project's ability to leverage the power of FOSS, leading to potential extensions to the pipeline. Besides, FOSS inherently promotes community engagement and collaboration, which is crucial for the project's success. To further incentivize contributors, we will explore methods like providing rewards, such as badges, and introducing gamification elements \cite{heritoge1, ch}, wherein a scoring system, based on image quality, could encourage better image capture and improve reconstructions. Signage at cultural heritage sites \cite{remotemon2}, with QR codes linking to the Tirtha page, can engage tourists and locals in contributing, particularly for neglected sites. Platform localization via community effort will be explored, making Tirtha accessible to a wider audience, which is necessary for places such as India. Crowdsourcing can also be leveraged for annotating and validating models \cite{csframeindia}. Implementing a discussion forum, such as using GitHub Discussions, can facilitate collaboration and feedback. Outside of making the platform more intuitive and accessible, we are looking into partnering with local archaeological or heritage conservation organizations to increase awareness and community engagement.

\section{Conclusion}
\label{sec:conc}
Advancements in web technologies, smartphone cameras, and tourism have made crowdsourced image collection a cost-effective way to document cultural heritage sites. In this work, we have detailed Tirtha, an end-to-end fully automated web-based crowdsourcing platform, that enables easy and inexpensive participation in digital heritage documentation, which is particularly essential in resource-limited countries, where many CH sites may be neglected or under-documented. It allows anyone with a smartphone and internet access, including non-experts, to contribute images. These images are used in a photogrammetry pipeline to create 3D models of the sites. The resulting models and images are stored as datasets under a non-commercial Creative Commons license, facilitating further use in research and creative disciplines. Each reconstruction is assigned a persistent and unique ARK identifier for tracking in research. This also lets us document a CH site through time, as its 3D model is updated with newer images, as a kind of ``Digital Twin''. Furthermore, Tirtha is open-source and built on open-source tools, making it customizable and extensible. CH preservation organizations can deploy it on their own infrastructure, adapting it to related heritage domains, while contributing improvements to the community. This approach can lead to the creation of decentralized repositories for crowdsourced images and 3D models, benefiting scholars and heritage conservation. The collected image sets would be long-term resources for constructing better models as algorithms improve. We expect Tirtha to foster a community that keeps the pipeline up-to-date, by contributing advancements in MVS. This has been a problem with existing solutions. We also envision the data products from Tirtha being used in creative projects, such as extended reality experiences, like in virtual CH site tours, and heritage education \cite{edu}.

We have highlighted limitations of this work, such as the need for reliable incentives for contributors and improving data quality. Moreover, we discussed some technical constraints, such as assessing data and reconstruction quality without ground truth and disseminating models on smartphones. In Future Work (Subsection~\ref{sec:fw}), we outlined potential ways to address many of these limitations, while also deliberating on the future improvements to Tirtha, such as newer ways to leverage the community or to incorporate bleeding-edge reconstruction techniques. Regardless of the current limitations, Tirtha offers robustness and flexibility, while cultivating community engagement, which is essential for heritage conservation. Lastly, cultural heritage sites are not just important for their historical and societal significance, but also for promoting sustainable development, as emphasized by UNESCO. For instance, conserving cultural heritage sites can provide opportunities for local communities to engage in sustainable tourism and support their livelihoods. So, it is imperative to preserve these sites. We anticipate that Tirtha will play an impactful role in this regard.

\begin{acks}
  This project is funded by La fondation Dassault Systèmes (DSF Project ID: IN- 2021-3-02), with support from Odisha State Archaeology. We thank the priests of the Somanatha \& Gopinatha temples, and Annada Prasad Behera for their valuable feedback. Special thanks to the FOSS developers behind Django, Celery, jQuery, Gunicorn, Nginx, RabbitMQ, PostgreSQL, MeshOptimizer, obj2gltf, OpenCV, NSFWJS, MANIQA, and \texttt{<model-viewer>}. Special recognition goes to the members of the AliceVision project for their work on Meshroom, which made this project possible.
\end{acks}

\bibliographystyle{ACM-Reference-Format}
\bibliography{./refs.bib}

\appendix

\section{Data Layer}
\label{app:data}
\begin{figure*}
  \centering
  \includegraphics[width=\textwidth]{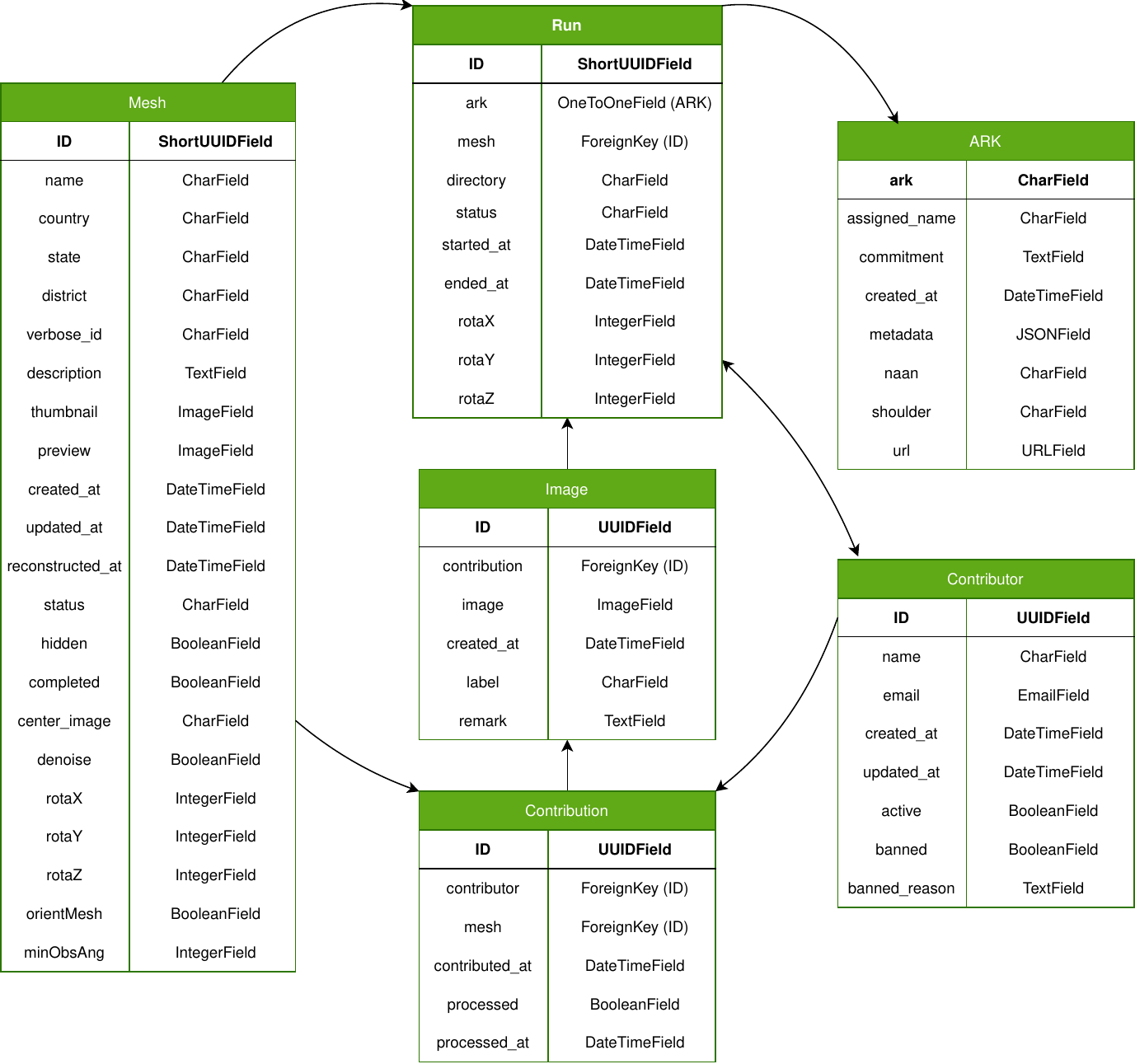}
  \caption{Entity relationship diagram of the database. Single-headed arrows represent one-to-many relationships, while double-headed arrows represent many-to-many relationships.}
  \Description{Entity relationship diagram of the database, showing the various tables in the PostgreSQL database and their inter-relationships. Single-headed arrows represent one-to-many relationships, while double-headed arrows represent many-to-many relationships.}
  \label{fig:erd}
\end{figure*}
The central table in Tirtha's database is the `Mesh' model, containing site metadata like name, description and location. The location fields are used to create unique plaintext Verbose\_IDs that power the website's search functionality and make it simple for contributors to identify CH sites. Optional reconstruction properties, such as `center\_image', orientation overrides, whether to denoise and minimum observation angle are also stored in the `Mesh' model, while the `status' field tracks the reconstruction progress, and the `completed' field marks sites with sufficiently thorough 3D models as concluded. Post-completion, a site does not accept more contributions. The `Mesh' model relates to the `Image' model, which stores image paths and IQA results. Images in each upload are grouped with the `Contribution' model, while the `Contributor' model stores names and email addresses, as supplied by the identity provider, among other moderation-related fields. The final components are the `ARK' and the `Run' models. The former stores the ARK persistent identifier for each reconstruction, along with metadata for the site and the run in JSON format, while the latter stores information about each reconstruction run, including status, timestamps, images \& contributions used, and relates to a unique ARK entry, if the run succeeds. The entity relationship diagram of the database is shown in Fig.~\ref{fig:erd}.

\section{Image Quality Assessment Results}
\label{app:iops}
\begin{figure*}
  \centering
  \begin{tabular}{cccc}
    \toprule
    \multirow{1}{2.5cm}[6.6em]{\textbf{``Good'' Images}} &
    \begin{subfigure}[t]{0.25\textwidth}
      \centering
      \includegraphics[width=\linewidth,angle=90]{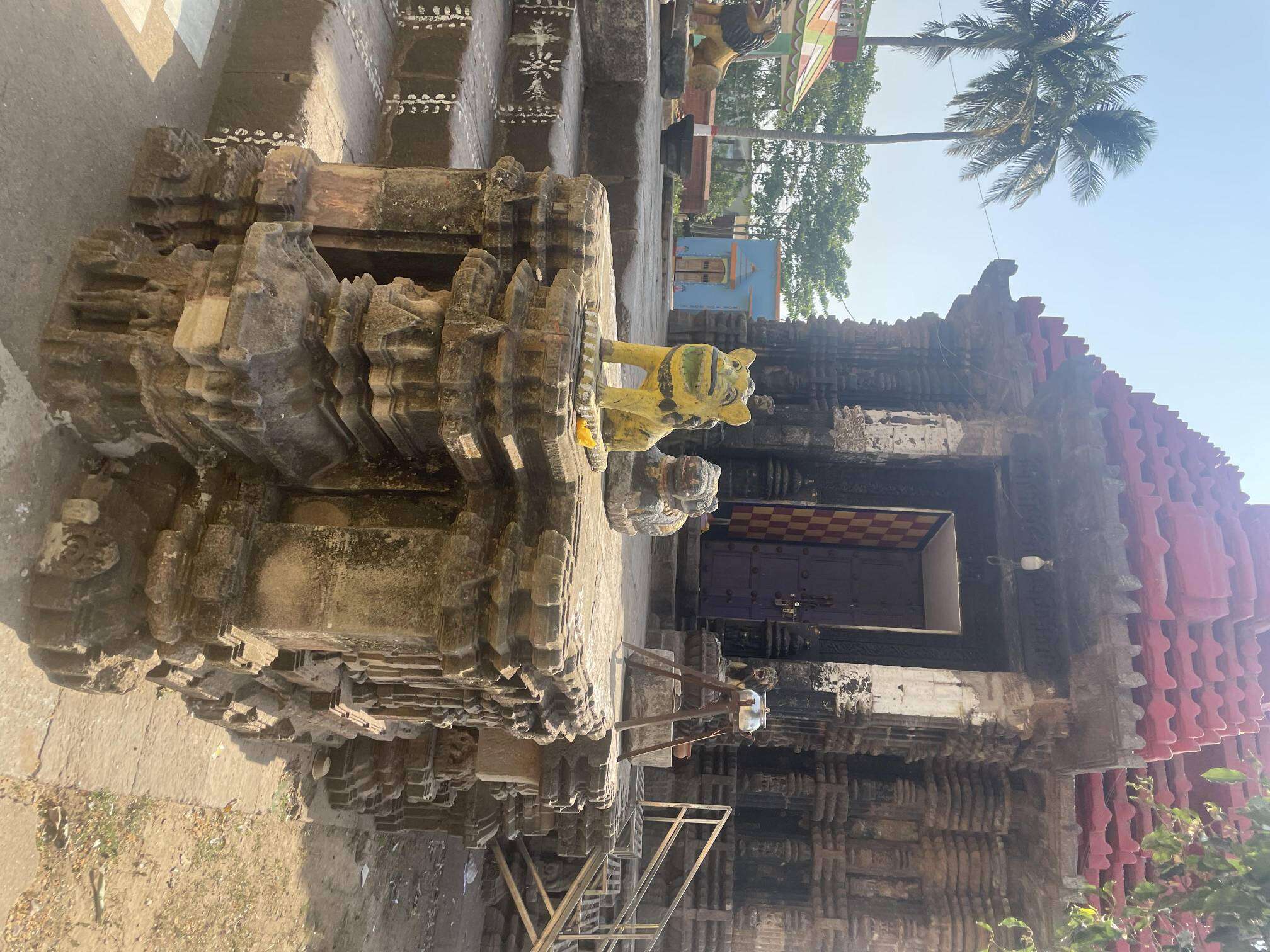}
    \end{subfigure} &
    \begin{subfigure}[t]{0.25\textwidth}
      \centering
      \includegraphics[width=\linewidth,angle=90]{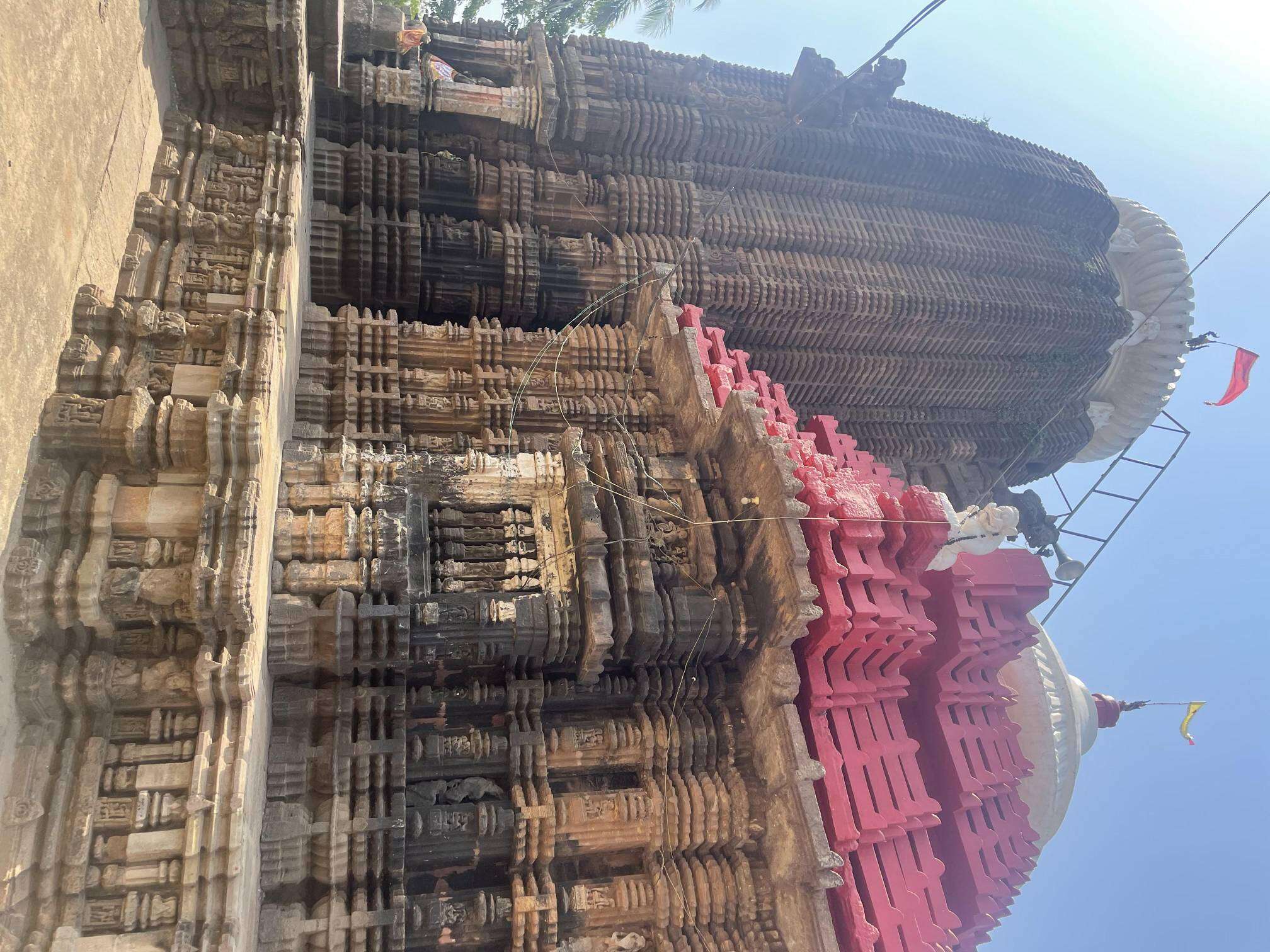}
    \end{subfigure} &
    \begin{subfigure}[t]{0.25\textwidth}
      \centering
      \includegraphics[width=\linewidth,angle=90]{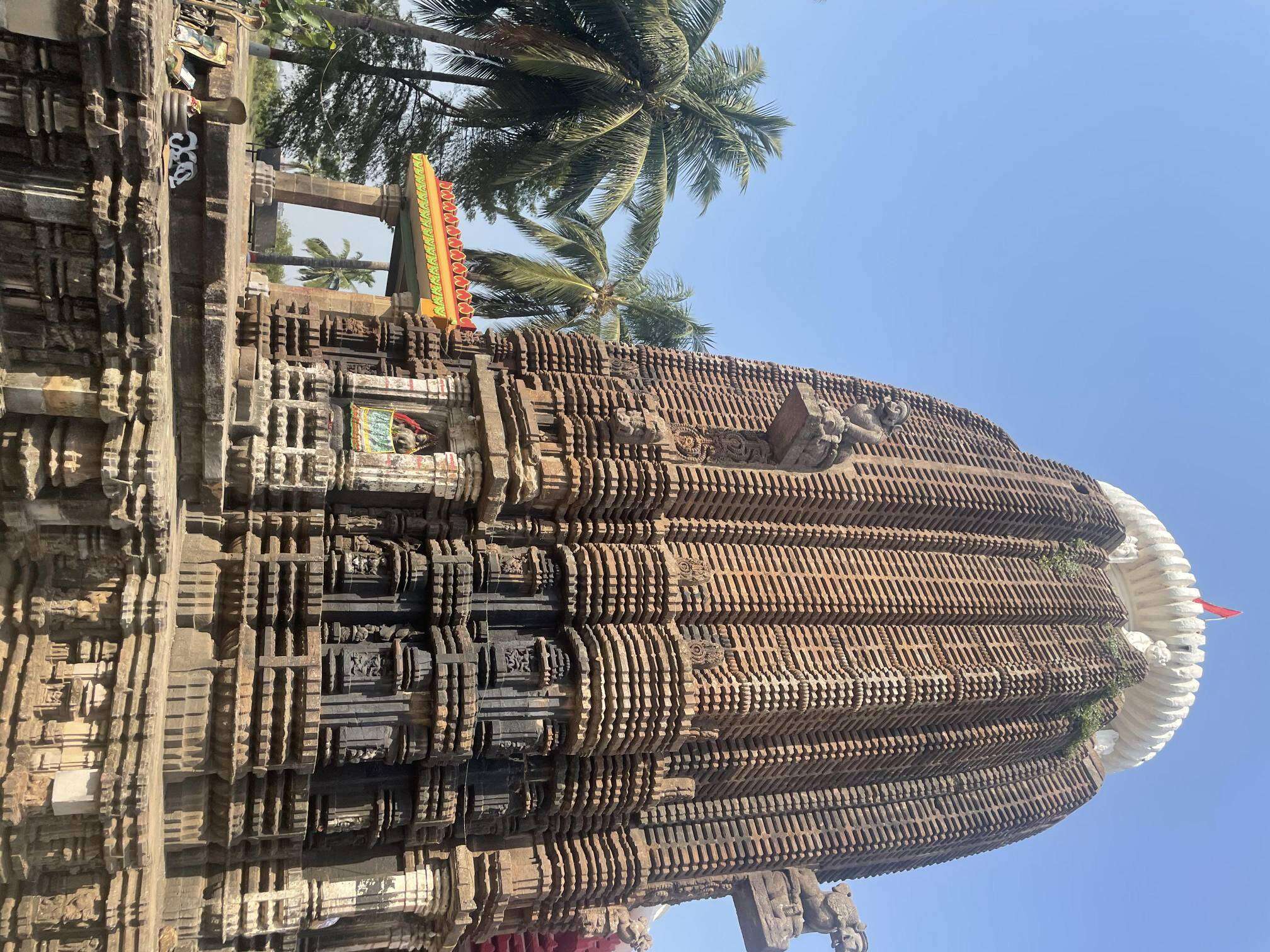}
    \end{subfigure} \\
    \midrule
    \multirow{1}{2.5cm}[6.6em]{\textbf{``Bad'' Images}} &
    \begin{subfigure}[t]{0.25\textwidth}
      \centering
      \includegraphics[width=\linewidth,angle=90]{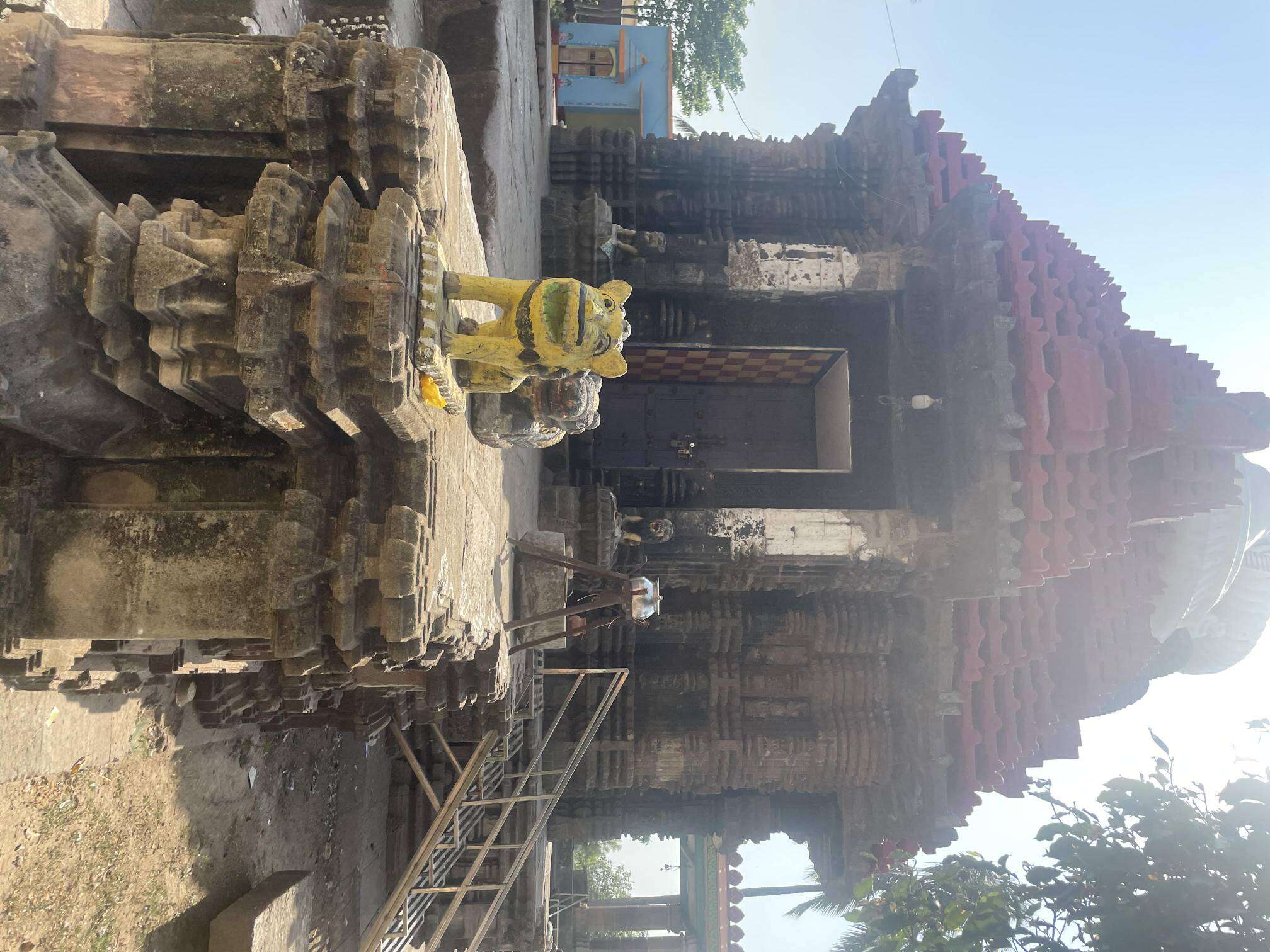}
    \end{subfigure} &
    \begin{subfigure}[t]{0.25\textwidth}
      \centering
      \includegraphics[width=\linewidth,angle=90]{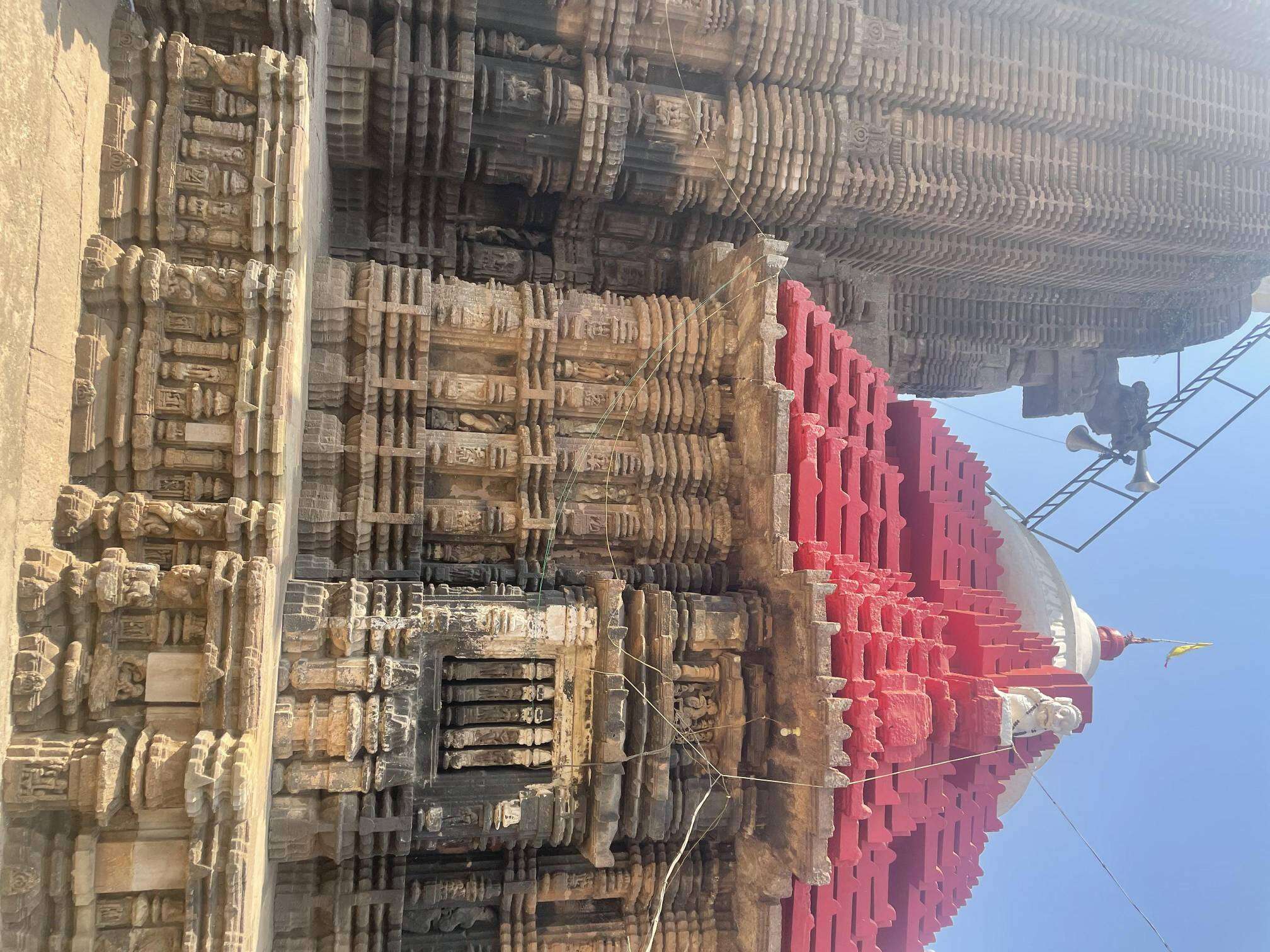}
    \end{subfigure} &
    \begin{subfigure}[t]{0.25\textwidth}
      \centering
      \includegraphics[width=\linewidth,angle=90]{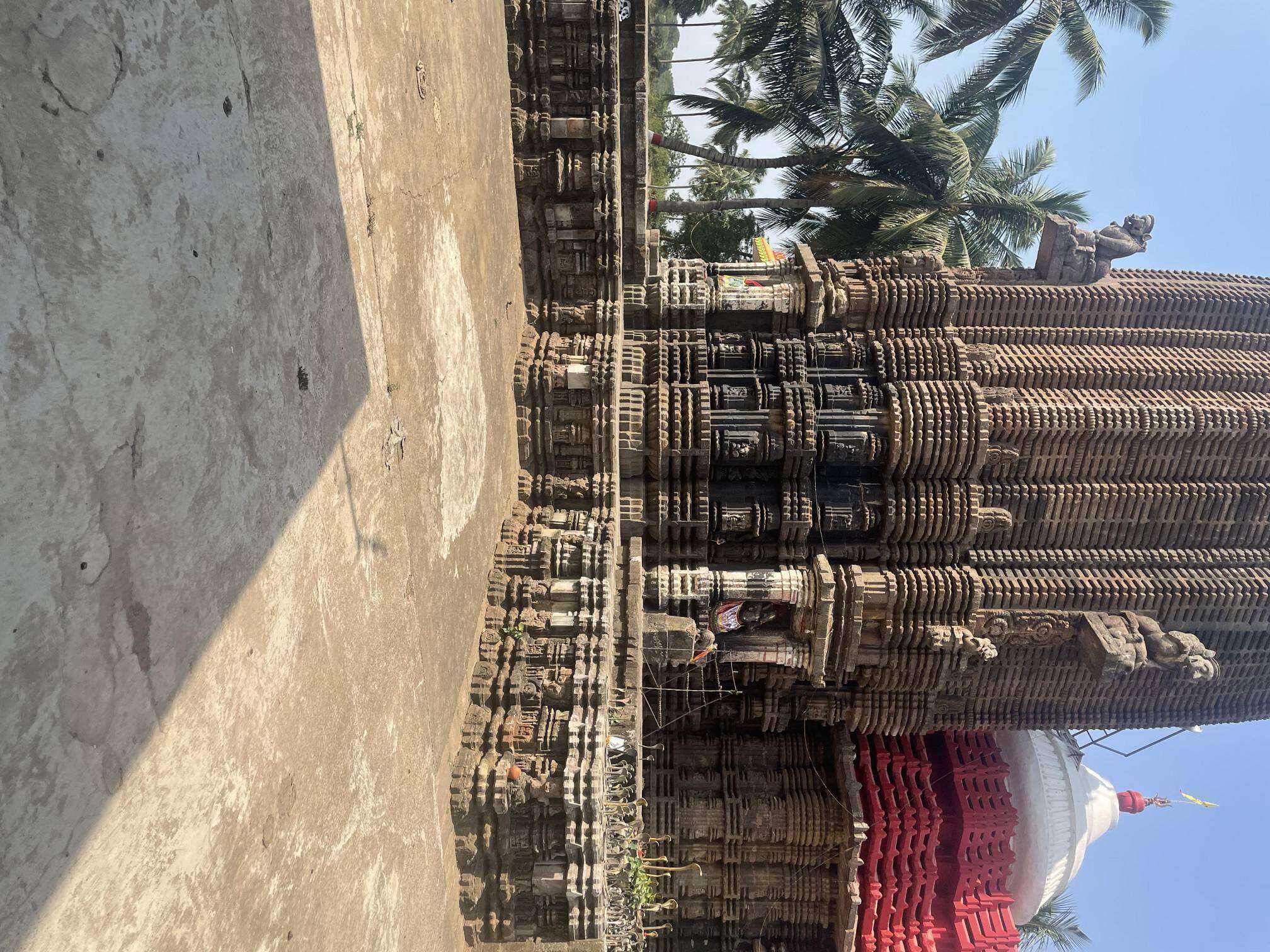}
    \end{subfigure} \\
    \bottomrule
  \end{tabular}
  \caption{``Good'' (positive) and ``Bad'' (negative) filtering results for the crowdsourced Somanatha Temple dataset. The ``Bad'' row includes images filtered out by the DR, CNR and MANIQA thresholds, respectively. The ``Good'' row includes images that passed all of the thresholds (Chosen thresholds - DR: 100; CNR: 17.5; MANIQA: 0.6).}
  \Description{Comparing the ``Good'' (positive) and ``Bad'' negative filtering results for the crowdsourced Somanatha Temple dataset. The ``Bad'' row includes images that were filtered out by the Dynamic Range (CR) threshold, the Contrast-to-Noise Ratio (CNR) threshold and No Reference Image Quality Assessment (using MANIQA) thresholds, respectively. On the other hand, the ``Good'' row includes images that passed not filtered out by any of the thresholds.}
  \label{fig:iops1}
\end{figure*}

\begin{figure*}
  \centering
  \begin{tabular}{cccc}
    \toprule
    \multirow{1}{1.5cm}[6.3em]{\textbf{Somanatha}} &
    \begin{subfigure}[t]{0.25\textwidth}
      \centering
      \includegraphics[width=\linewidth]{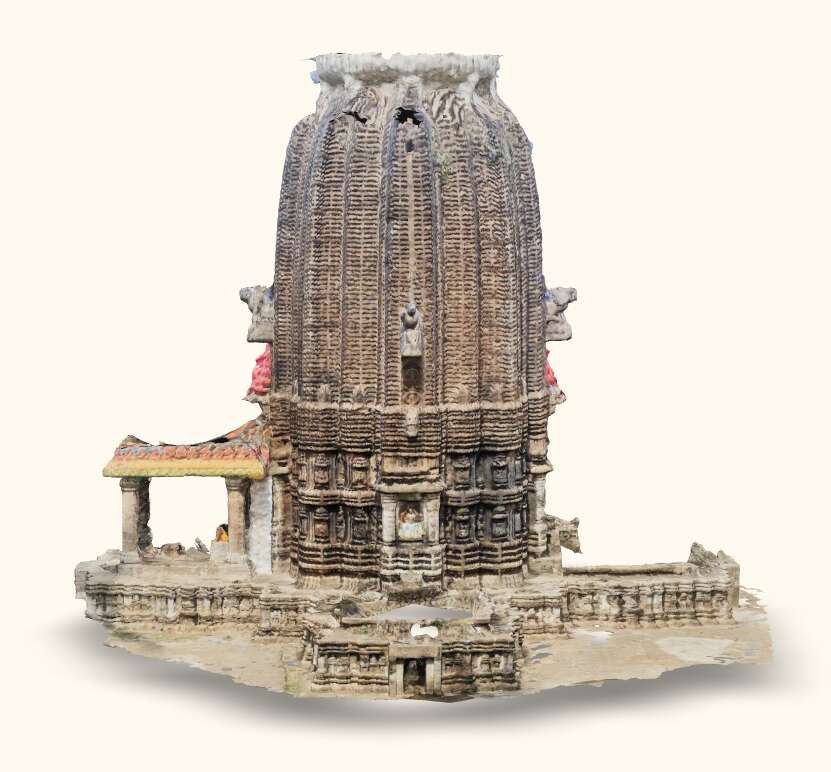}
    \end{subfigure} &
    \begin{subfigure}[t]{0.25\textwidth}
      \centering
      \includegraphics[width=\linewidth]{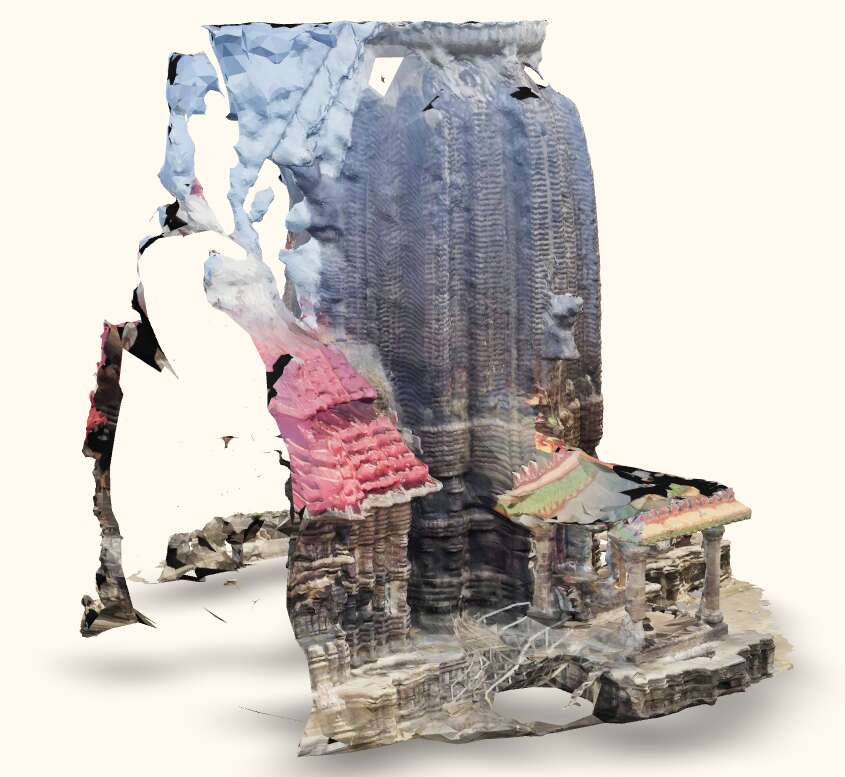}
    \end{subfigure} &
    \begin{subfigure}[t]{0.24\textwidth}
      \centering
      \includegraphics[width=\linewidth]{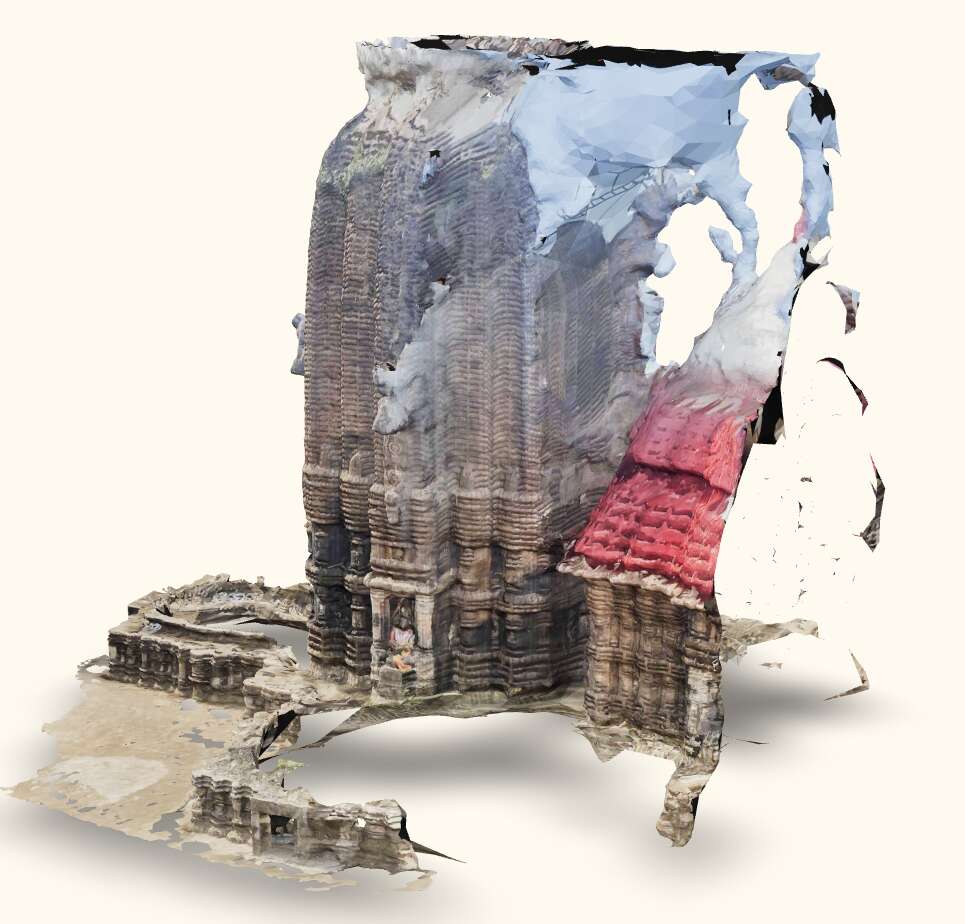}
    \end{subfigure} \\
    \midrule
    \multirow{1}{1.5cm}[6.6em]{\textbf{Gopinatha}} &
    \begin{subfigure}[t]{0.25\textwidth}
      \centering
      \includegraphics[width=\linewidth]{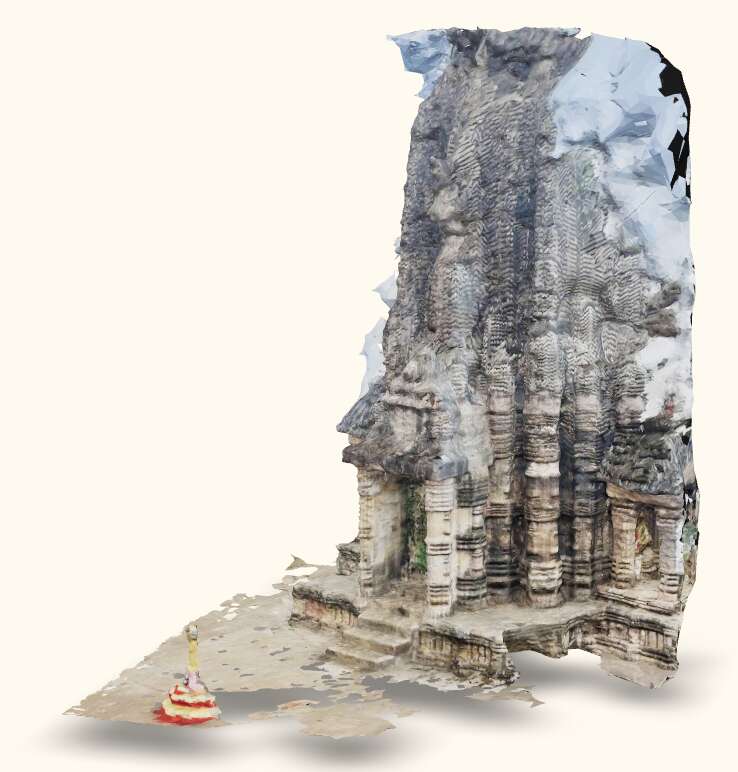}
    \end{subfigure} &
    \begin{subfigure}[t]{0.25\textwidth}
      \centering
      \includegraphics[width=\linewidth]{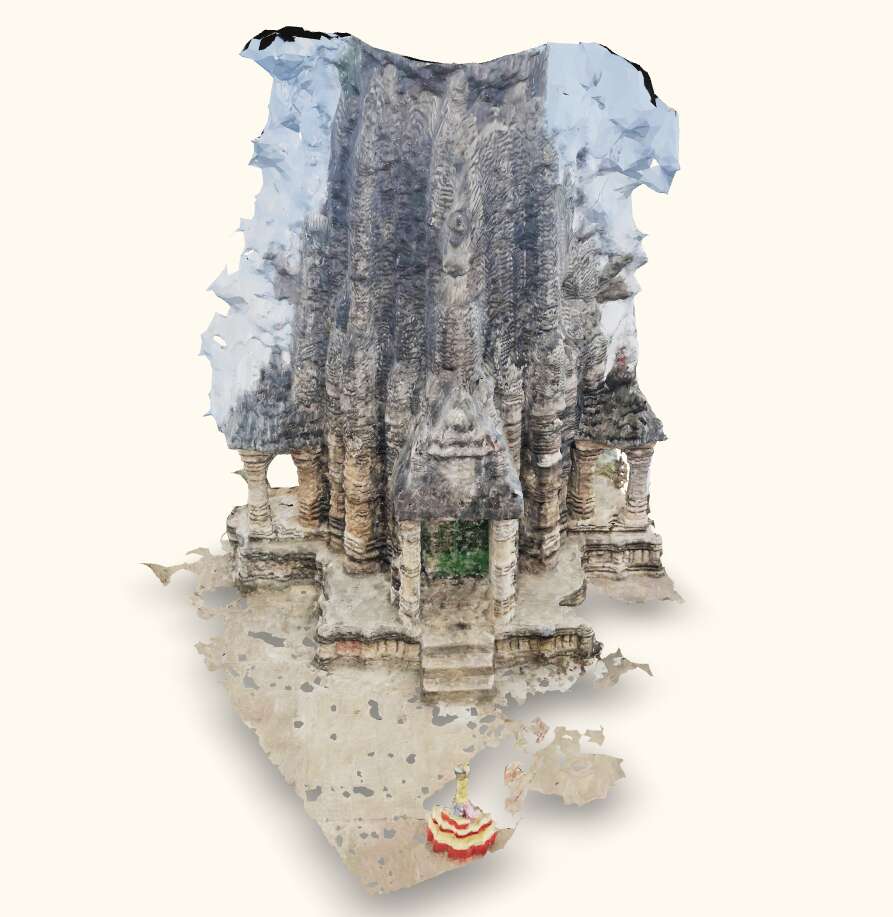}
    \end{subfigure} &
    \begin{subfigure}[t]{0.26\textwidth}
      \centering
      \includegraphics[width=\linewidth]{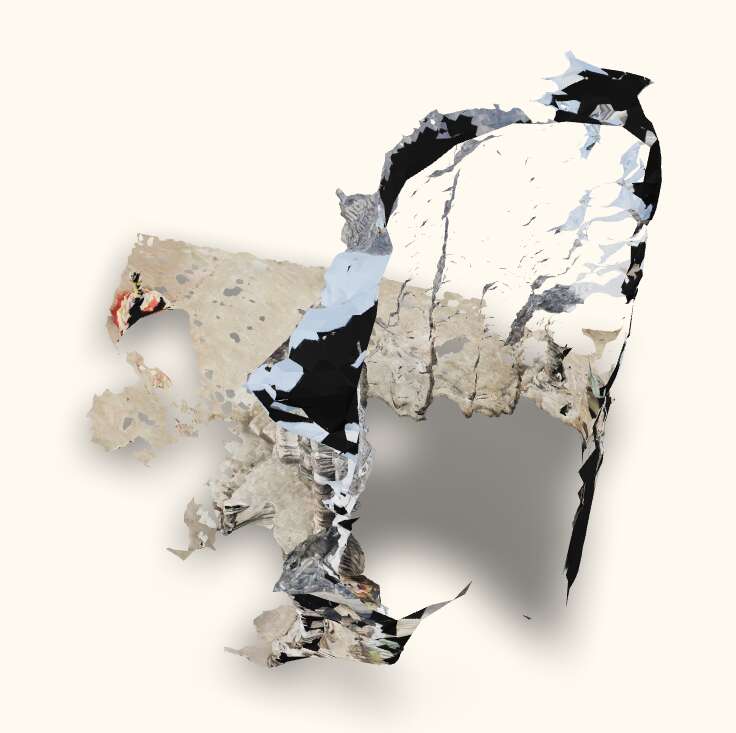}
    \end{subfigure} \\
    \bottomrule
  \end{tabular}
  \caption{Reconstruction results for the two crowdsourced datasets with the aforementioned thresholds for IQA steps. Compare with Fig.~\ref{fig:gopibadcomp} and Fig.~\ref{fig:somabadcomp}.}
  \Description{Comparing the reconstruction results for the two crowdsourced datasets after applying the aforementioned strict thresholds for IQA steps.}
  \label{fig:iops2}
\end{figure*}

\begin{figure*}
  \centering
  \begin{subfigure}[b]{0.52\textwidth}
    \centering
    \includegraphics[width=\textwidth]{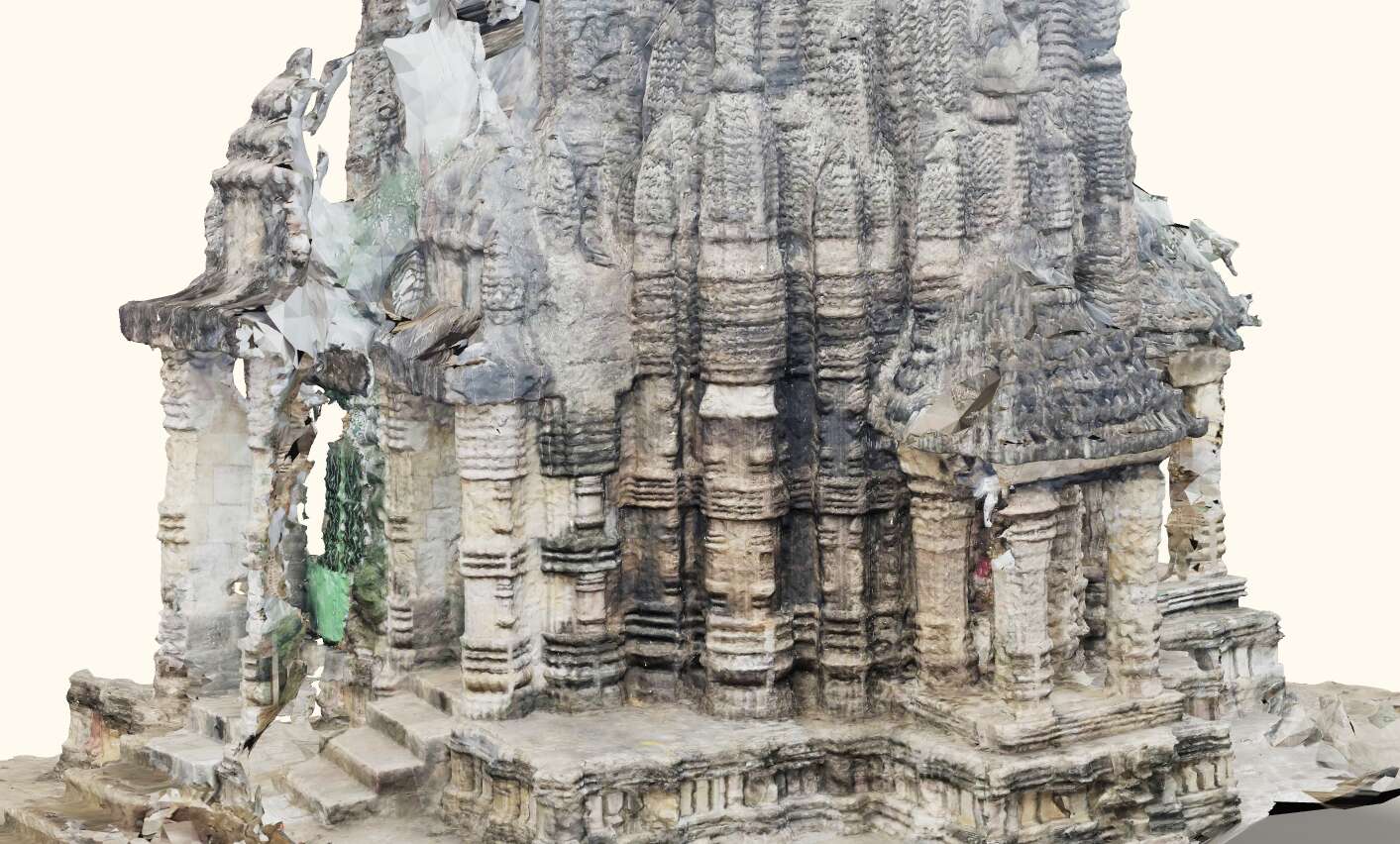}
    \caption{Mesh reconstructed using crowdsourced data}
  \end{subfigure}%
  \hspace{0.25cm}
  \begin{subfigure}[b]{0.41\textwidth}
    \centering
    \includegraphics[width=\textwidth]{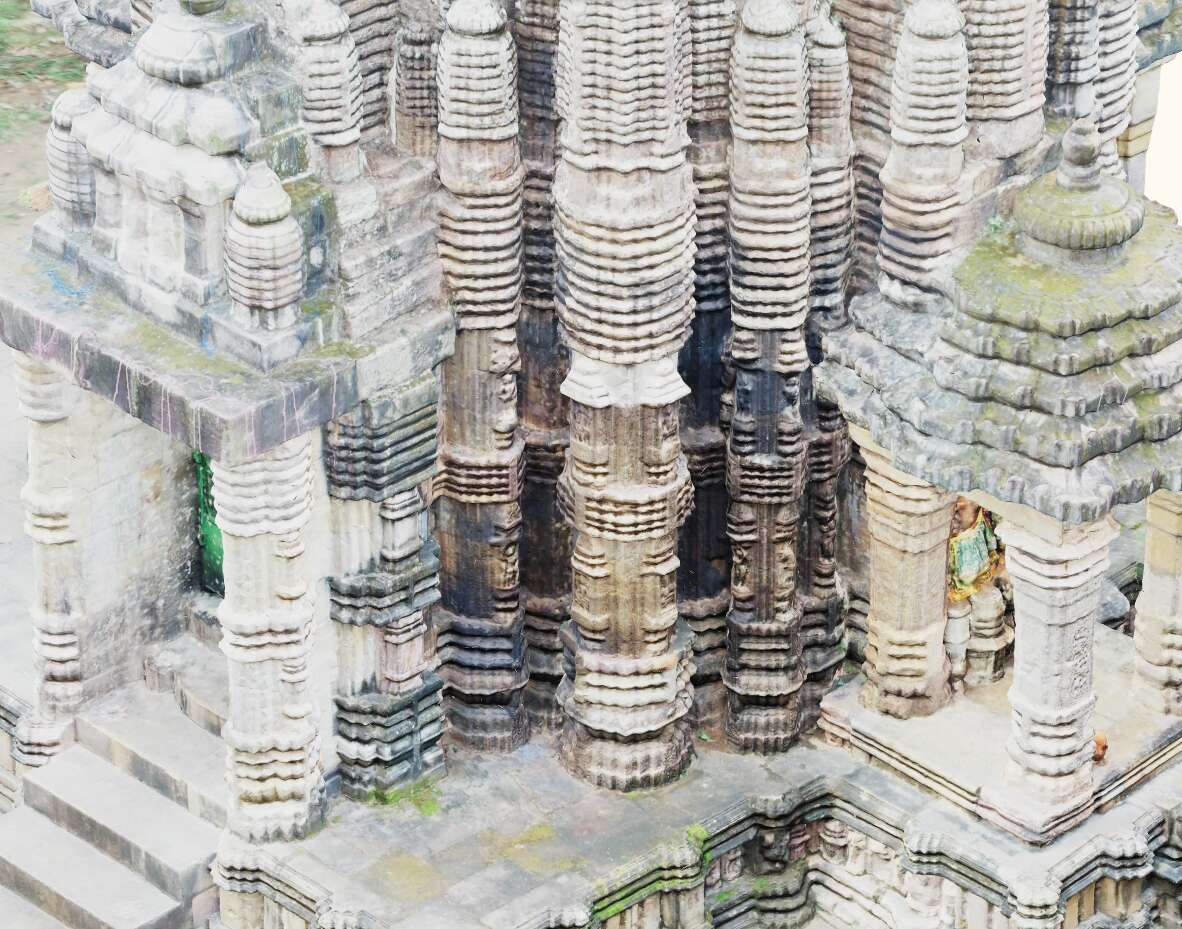}
    \caption{Mesh reconstructed using expert-sourced data}
  \end{subfigure}
  \caption{Gopinatha Temple: Contrast between similar areas of the reconstructions}
  \Description{Highlighting a problematic area in the reconstruction of Gopinatha Temple, generated using crowdsourced images, as compared to the same area in the reconstruction using expert-sourced data.}
  \label{fig:gopibadcomp}
\end{figure*}

\begin{figure*}
  \centering
  \begin{subfigure}[b]{0.44\textwidth}
    \centering
    \includegraphics[width=0.9\textwidth]{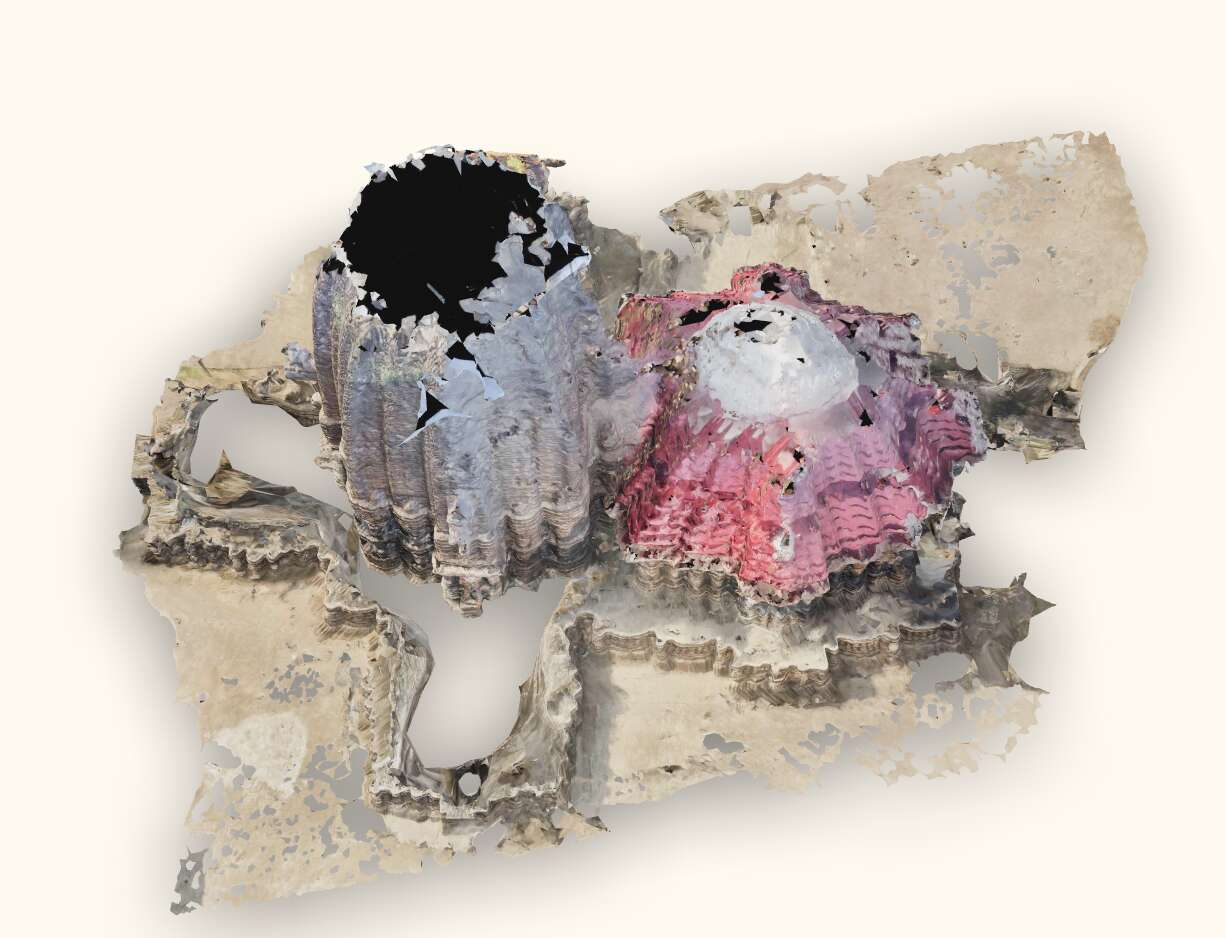}
    \caption{Mesh reconstructed using crowdsourced data}
  \end{subfigure}%
  \hspace{0.25cm}
  \begin{subfigure}[b]{0.41\textwidth}
    \centering
    \includegraphics[width=0.9\textwidth]{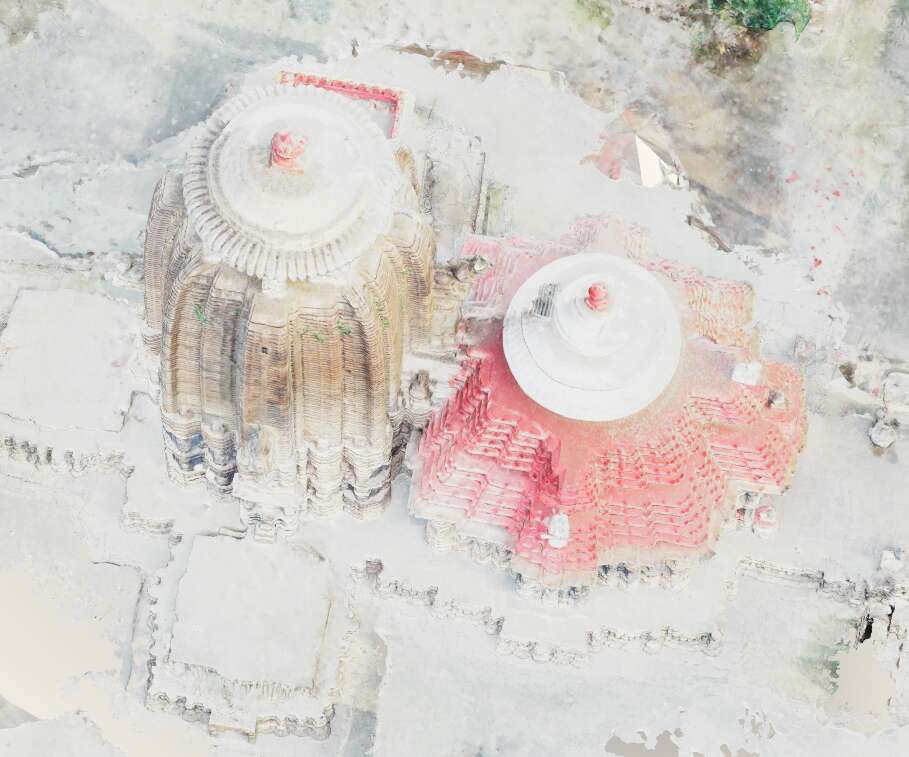}
    \caption{Mesh reconstructed using expert-sourced data}
  \end{subfigure}
  \caption{Somanatha Temple: Contrast between similar areas of the reconstructions}
  \Description{Highlighting a problematic area in the reconstruction of Somanatha Temple, generated using crowdsourced images, as compared to the same area in the reconstruction using expert-sourced data.}
  \label{fig:somabadcomp}
\end{figure*}

Here, we highlight some results from the IQA steps. Figure~\ref{fig:iops1} shows the IQA results for the crowdsourced Somanatha Temple dataset. The ``Bad images'' row includes images that were filtered out by the Dynamic Range (DR), Contrast-to-Noise Ratio (CNR) and the No Reference Image Quality Assessment (using MANIQA) thresholds, respectively, while the ``Good images'' row includes images that passed all of the IQA thresholds. The threshold values are mentioned in the image caption. Correspondingly, Fig.~\ref{fig:iops2} shows how these strict thresholds for the IQA steps impact the reconstruction. For reference, one may compare the meshes in Fig.~\ref{fig:iops2} with Fig.~\ref{fig:gopibadcomp} \& Fig.~\ref{fig:somabadcomp}. These outcomes indicate that strict thresholds improve image registration \& alignment and hence the overall reconstruction quality, but sacrifice completeness due to more filtered images, reflecting a quality-completeness tradeoff. To address this, we are exploring making the thresholds configurable on a case-by-case basis. For instance, the Somanatha Temple dataset has a large number of images with lens flares, which are primarily filtered out by the DR threshold. Hence, the DR threshold can be relaxed for this dataset, while the CNR and MANIQA thresholds can be made stricter. On the other hand, the Gopinatha Temple dataset has a large number of images with overexposure, which are mainly discarded by the CNR or MANIQA threshold. Hence, the CNR threshold can be relaxed for this dataset, while the DR and MANIQA thresholds can be made stricter. Further improvements to the IQA steps, such as using learning-based algorithms, are also being investigated.

\section{Extended Results}
\label{app:appres}
\begin{figure*}
  \centering
  \begin{tabular}{ccc}
    \toprule
    Tirtha + Crowdsourced Data & Tirtha + Expert-sourced Data & Expert-sourced Mesh \\
    \midrule
    \begin{subfigure}[t]{0.2975\textwidth}
      \centering
      \includegraphics[width=\linewidth]{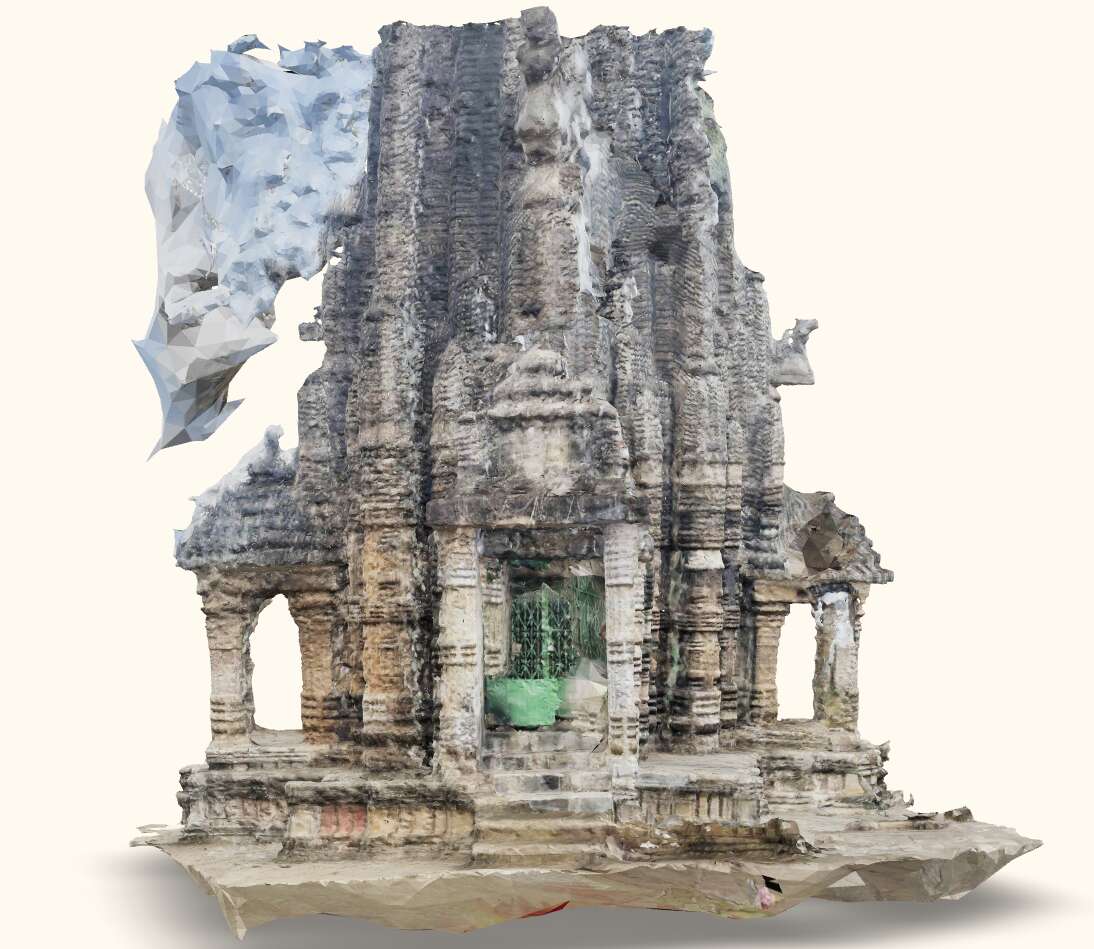}
    \end{subfigure} &
    \begin{subfigure}[t]{0.24\textwidth}
      \centering
      \includegraphics[width=\linewidth]{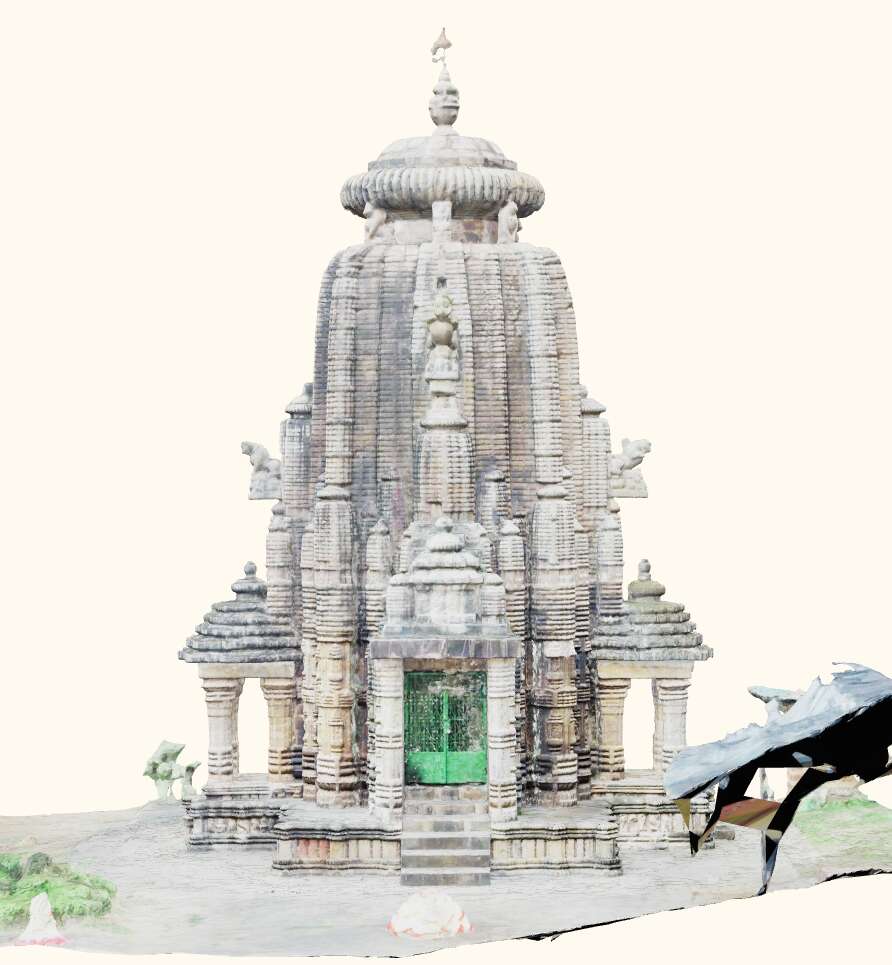}
    \end{subfigure} &
    \begin{subfigure}[t]{0.255\textwidth}
      \centering
      \includegraphics[width=\linewidth]{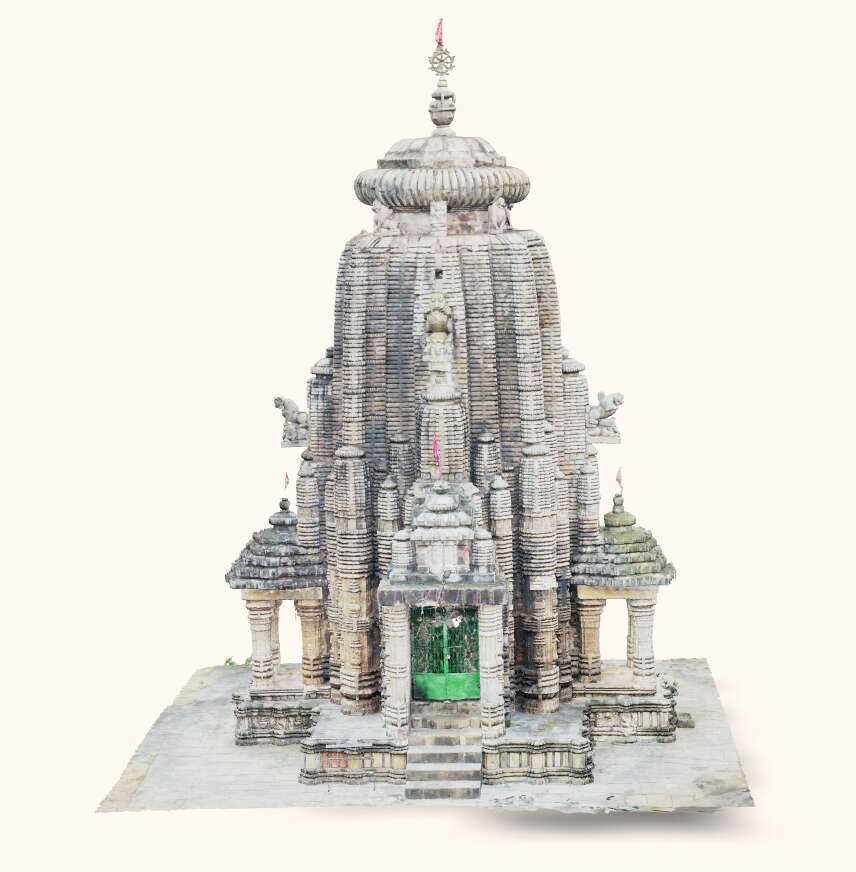}
    \end{subfigure} \\
    \begin{subfigure}[t]{0.29\textwidth}
      \centering
      \includegraphics[width=\linewidth]{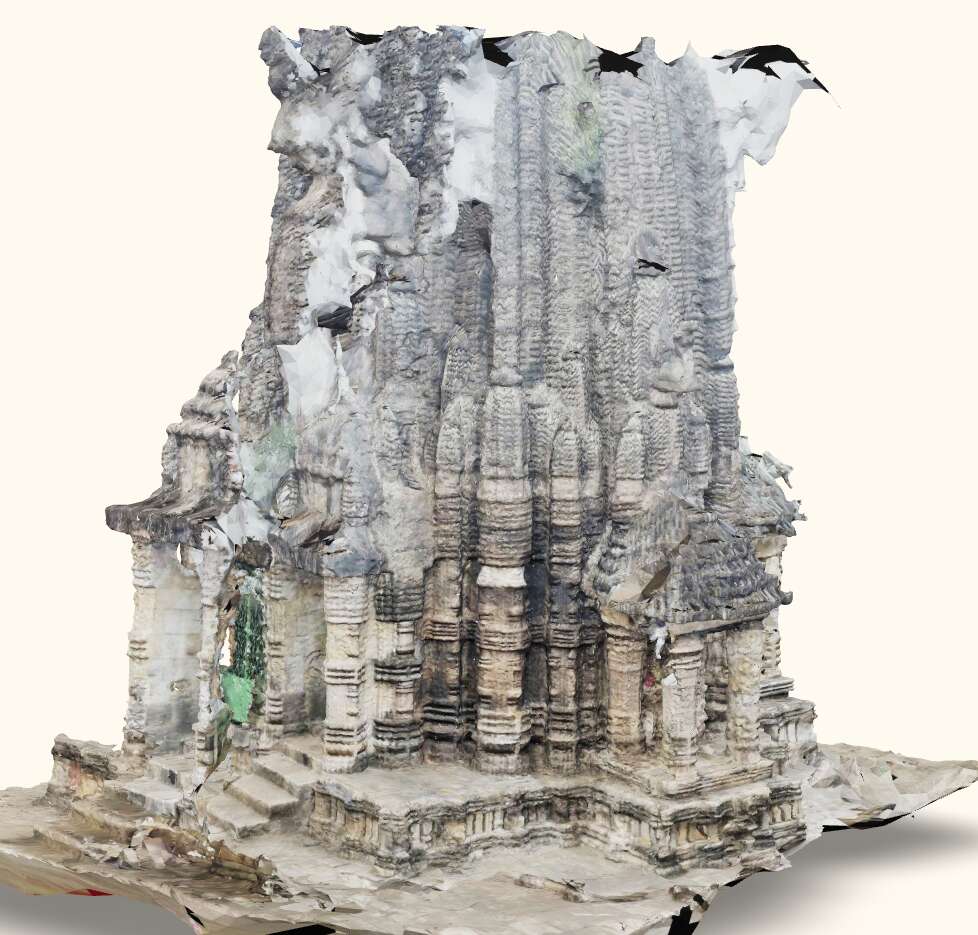}
    \end{subfigure} &
    \begin{subfigure}[t]{0.24\textwidth}
      \centering
      \includegraphics[width=\linewidth]{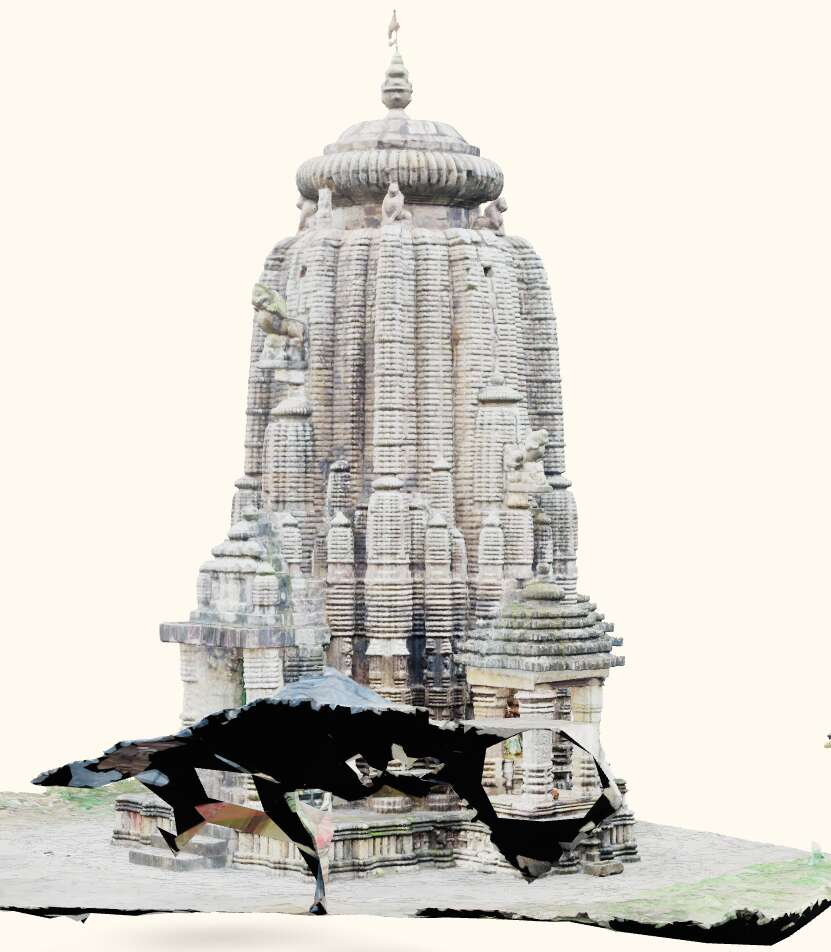}
    \end{subfigure} &
    \begin{subfigure}[t]{0.28\textwidth}
      \centering
      \includegraphics[width=\linewidth]{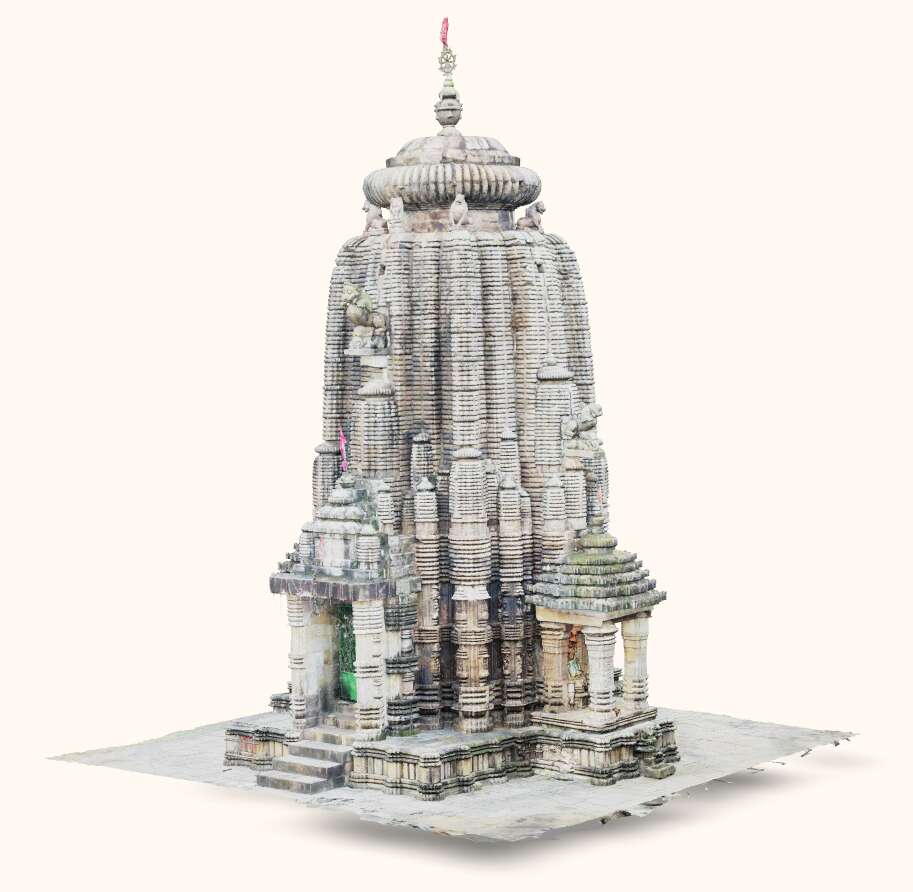}
    \end{subfigure} \\
    \begin{subfigure}[t]{0.29\textwidth}
      \centering
      \includegraphics[width=\linewidth]{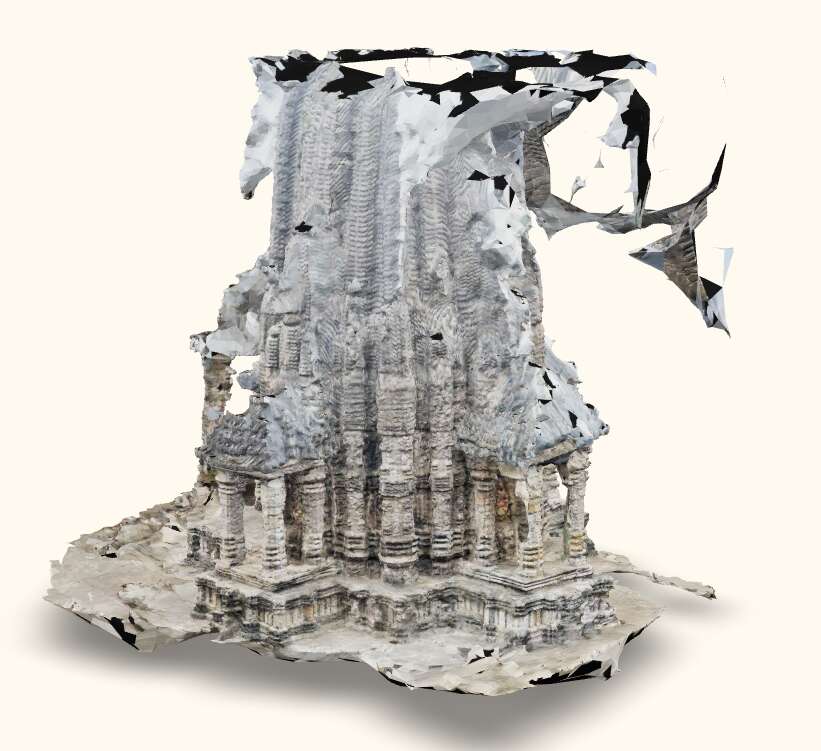}
    \end{subfigure} &
    \begin{subfigure}[t]{0.24\textwidth}
      \centering
      \includegraphics[width=\linewidth]{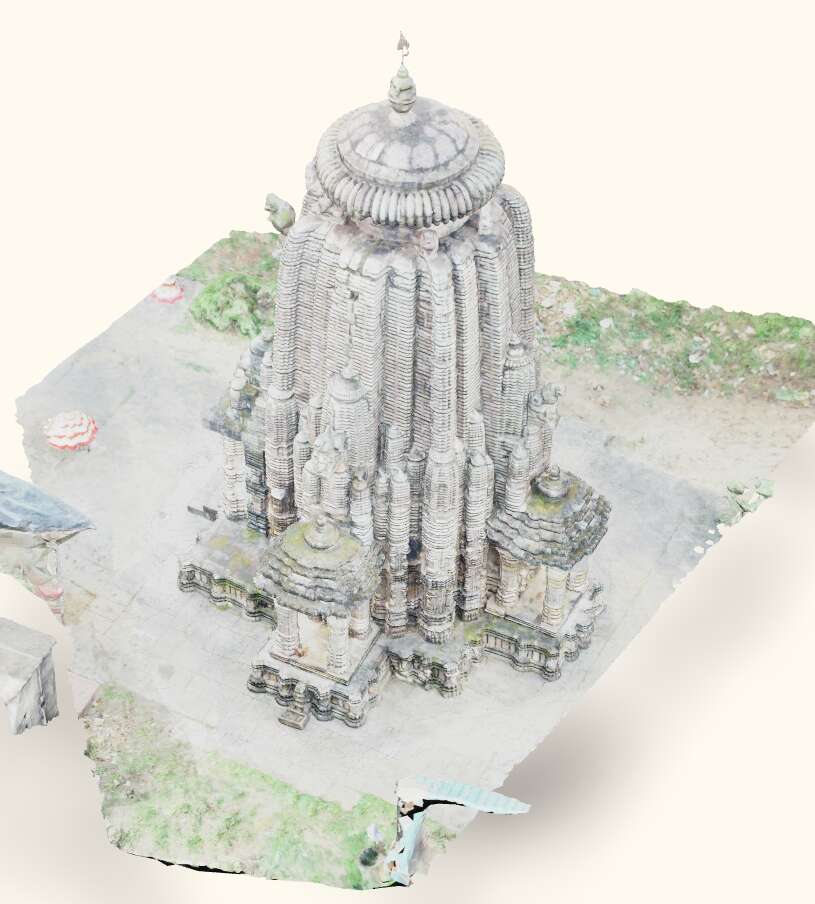}
    \end{subfigure} &
    \begin{subfigure}[t]{0.265\textwidth}
      \centering
      \includegraphics[width=\linewidth]{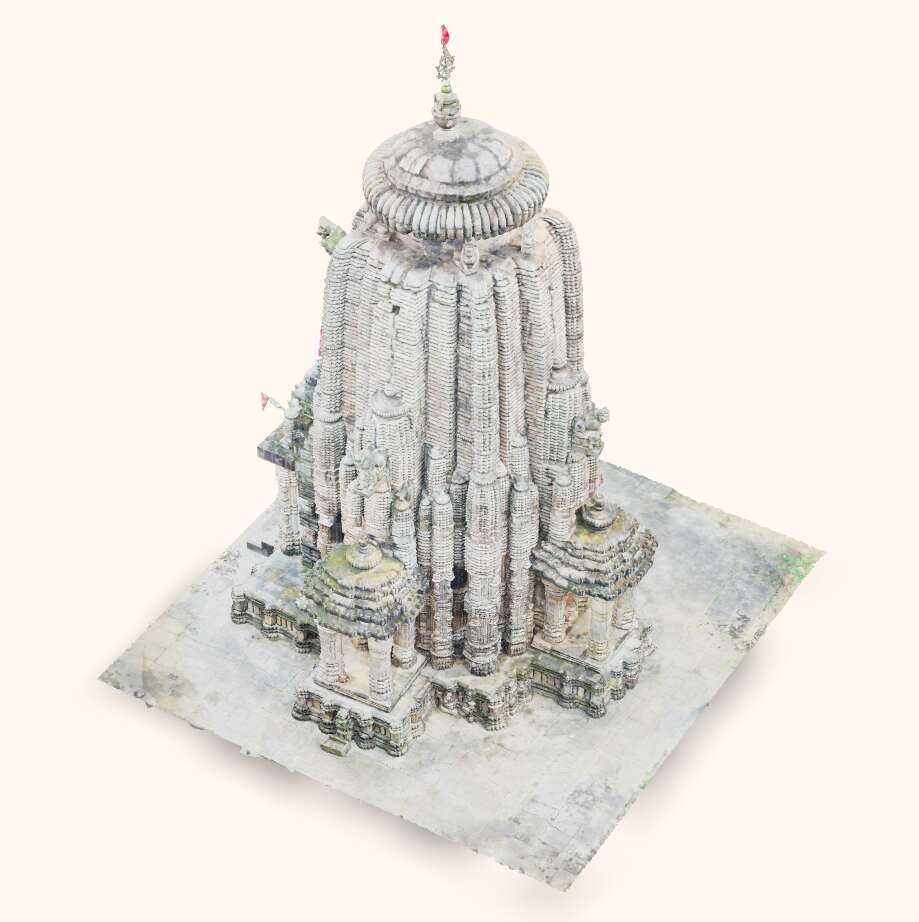}
    \end{subfigure} \\
    \begin{subfigure}[t]{0.29\textwidth}
      \centering
      \includegraphics[width=\linewidth]{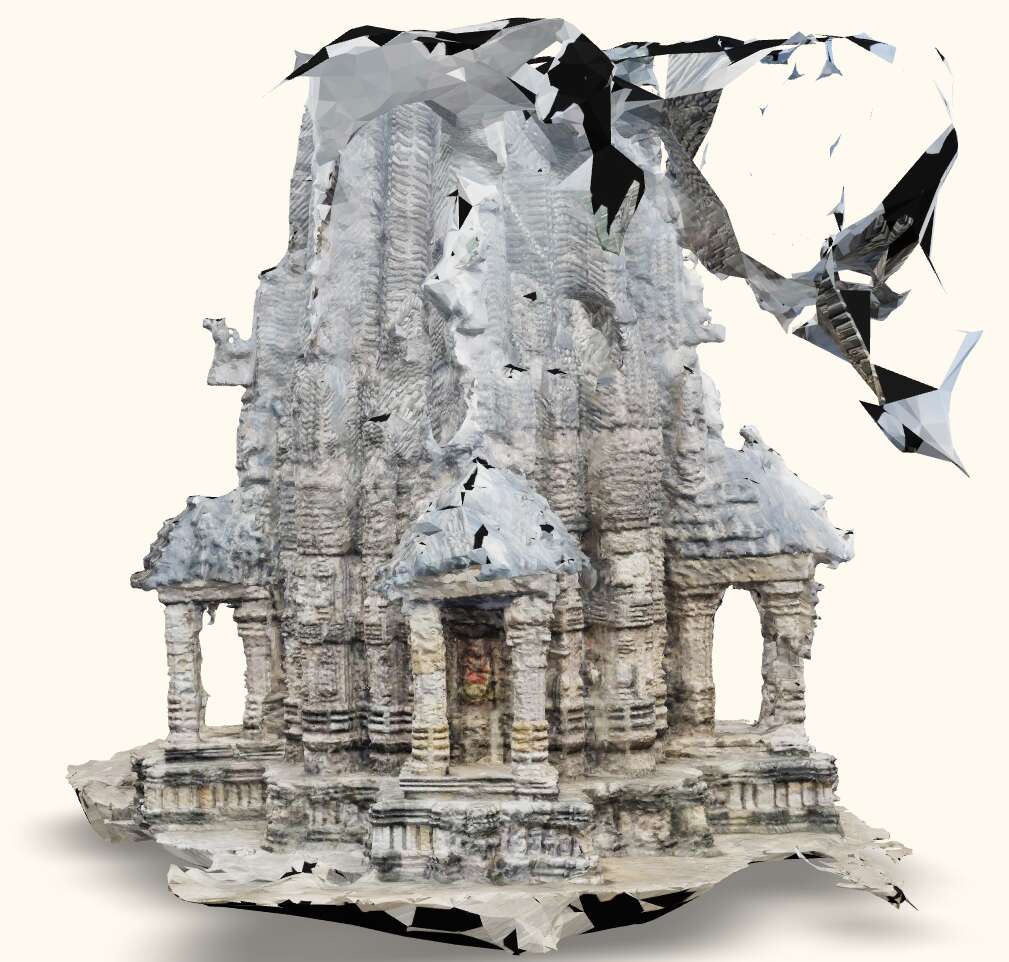}
    \end{subfigure} &
    \begin{subfigure}[t]{0.22\textwidth}
      \centering
      \includegraphics[width=\linewidth]{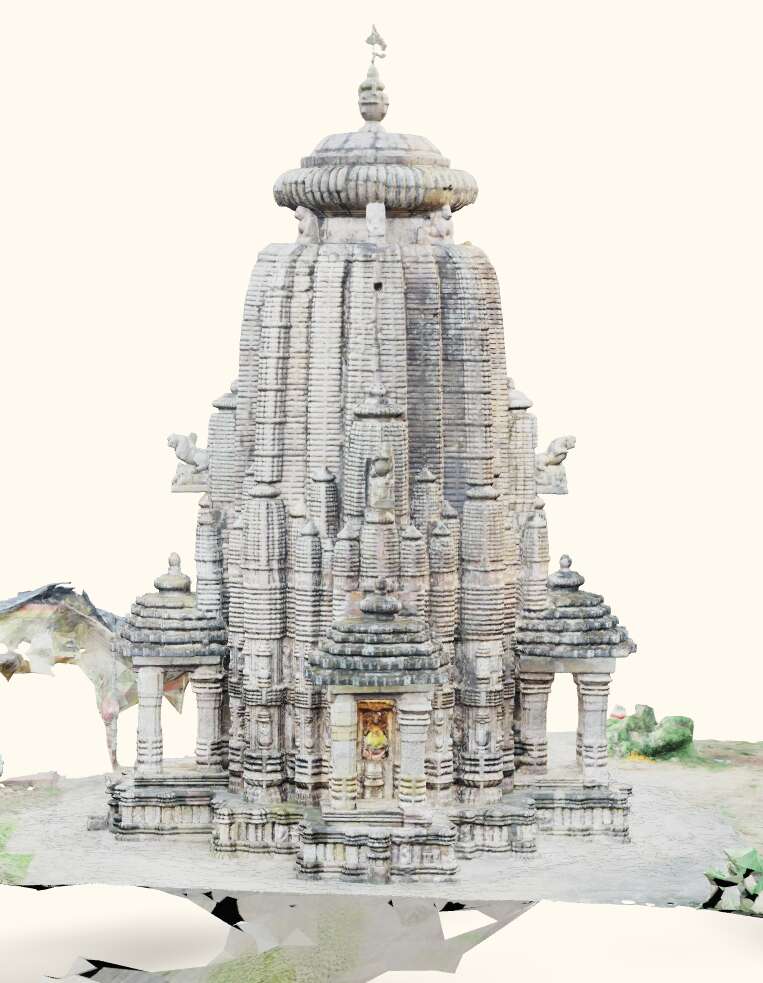}
    \end{subfigure} &
    \begin{subfigure}[t]{0.27\textwidth}
      \centering
      \includegraphics[width=\linewidth]{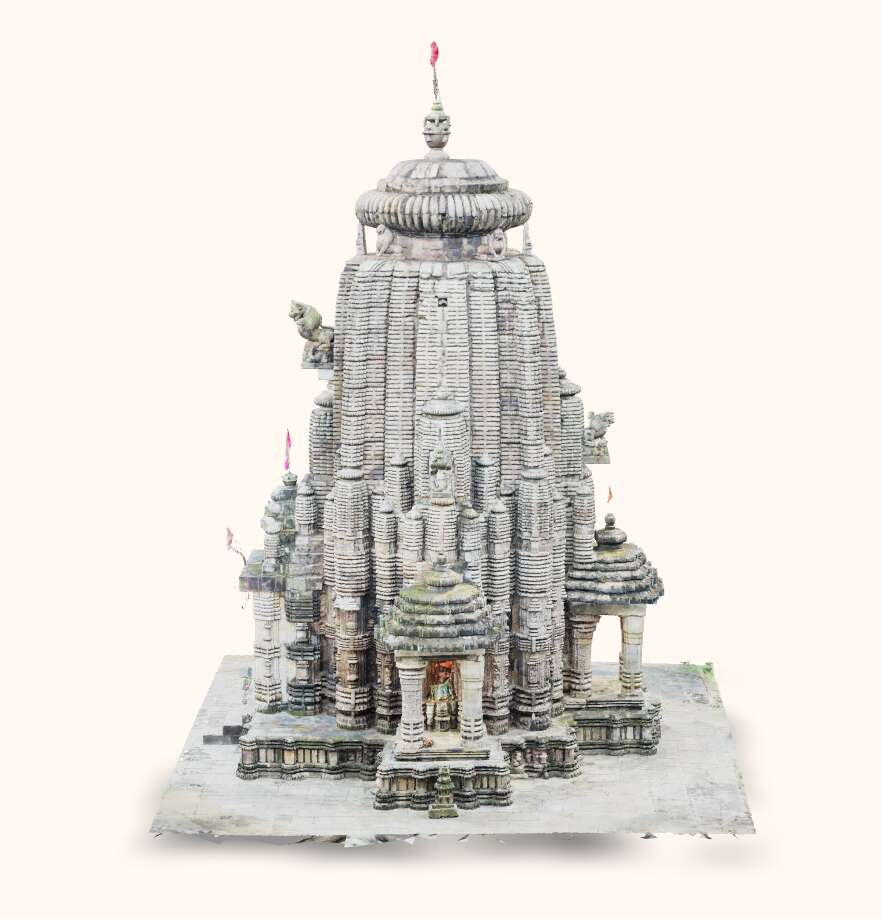}
    \end{subfigure} \\
    \bottomrule
  \end{tabular}
  \caption{Gopinatha Temple: Comparing the level of detail between Tirtha's and Metashape's reconstructions}
  \Description{Highlighting the difference in the level of detail between Tirtha's reconstruction and Metashape's reconstruction on expert-sourced data.}
  \label{fig:gopicomp}
\end{figure*}

\begin{figure*}
  \centering
  \begin{subfigure}[t]{0.4\textwidth}
    \centering
    \includegraphics[width=\textwidth]{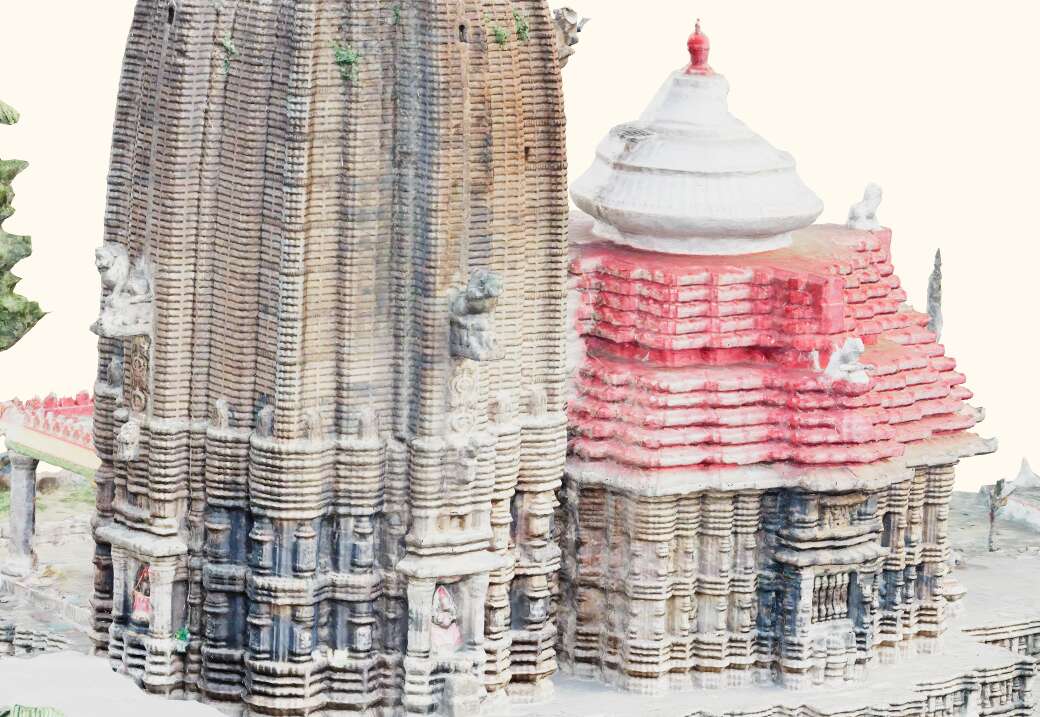}
    \caption{Tirtha-reconstructed Mesh}
  \end{subfigure}%
  \hspace{0.25cm}
  \begin{subfigure}[t]{0.43\textwidth}
    \centering
    \includegraphics[width=\textwidth]{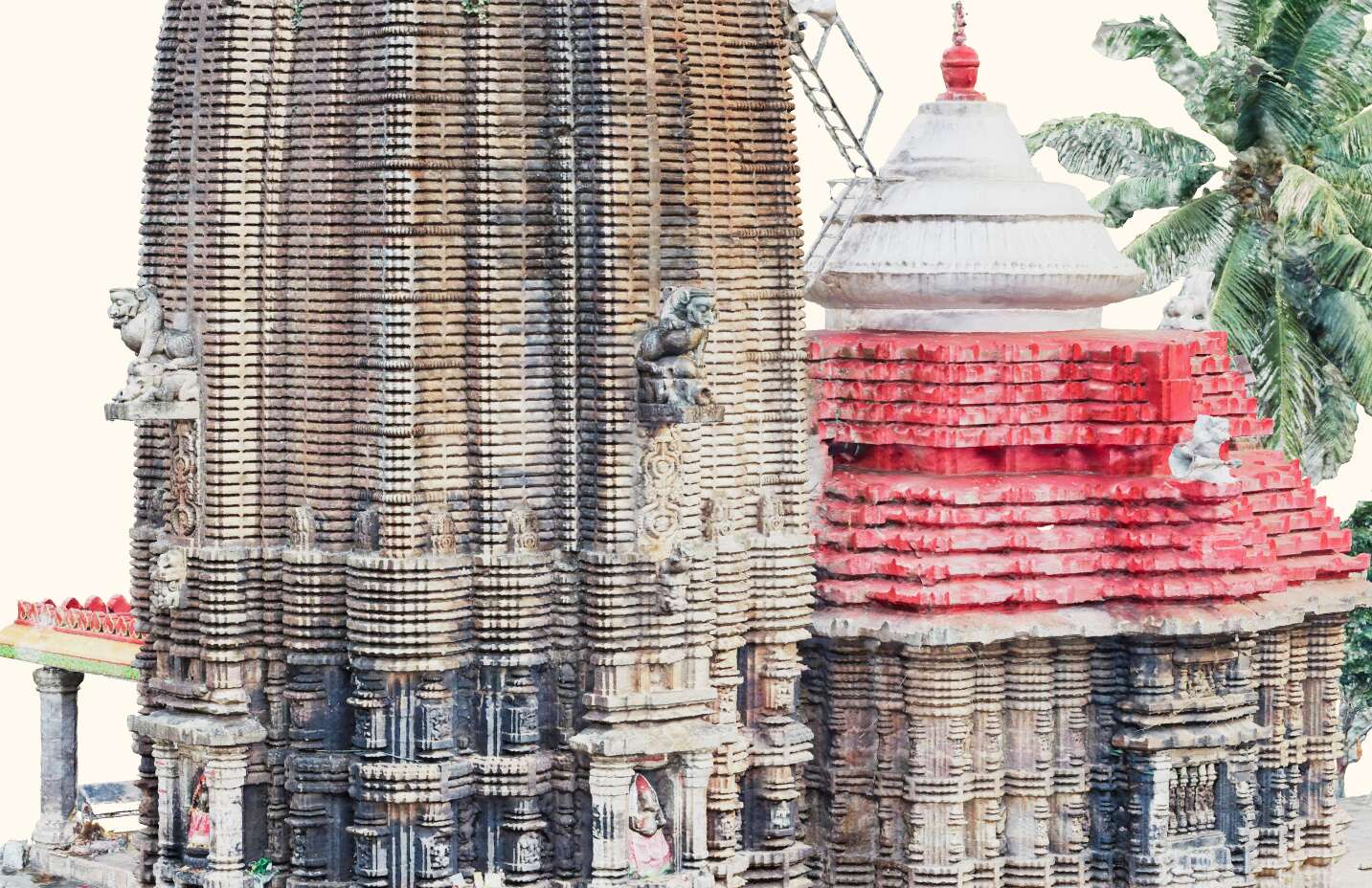}
    \caption{Expert-sourced Mesh}
  \end{subfigure}
  \caption{Somanatha Temple: Comparing the level of detail between Tirtha's and Metashape's reconstructions}
  \Description{Highlighting the difference in the level of detail between Tirtha's reconstruction and Metashape's reconstruction on expert-sourced data.}
  \label{fig:somazoomin}
\end{figure*}

\end{document}